\documentclass[10pt,journal,compsoc]{IEEEtran}

\usepackage{tabularx,array}

\newcolumntype{L}[1]{>{\raggedright\arraybackslash}p{#1}}
\newcolumntype{Y}{>{\raggedright\arraybackslash}X}

\usepackage[table]{xcolor}
\usepackage{xltabular}
\usepackage{booktabs}
\usepackage{multicol}

\definecolor{taskClassif}{RGB}{253, 237, 236} 
\definecolor{taskSeg}{RGB}{232, 248, 245}     
\definecolor{taskAnomaly}{RGB}{240, 240, 255}
\definecolor{taskCrossModal}{RGB}{255, 245, 230}
\definecolor{taskReportGen}{RGB}{230, 245, 255}
\definecolor{taskEmerging}{RGB}{250, 240, 245}
\definecolor{taskSurvival}{RGB}{238, 246, 232} 

\newcolumntype{L}[1]{>{\raggedright\arraybackslash}p{#1}}
\newcolumntype{Y}{>{\raggedright\arraybackslash}X}

\usepackage{amsmath,amssymb,amsfonts,mathtools}

\usepackage{graphicx}
\usepackage{grffile}
\usepackage[caption=false,font=footnotesize]{subfig}
\usepackage{float}

\usepackage{booktabs}
\usepackage{multirow}
\usepackage{array}
\usepackage{tabularx}
\usepackage{longtable}
\usepackage{ltablex}
\usepackage{makecell}
\keepXColumns

\usepackage{algorithm}
\usepackage{algorithmic}

\usepackage{xcolor}
\usepackage{enumitem}

\usepackage{cite}
\usepackage{url}

\usepackage{textcomp}
\usepackage{stfloats}
\usepackage{verbatim}

\begin{document}
\bstctlcite{IEEEexample:BSTcontrol}

\title{Deep Prototype Learning: A Unified Taxonomy and Survey in Medical Imaging}

\author{
Luu Le*,
Thanh-Huy Nguyen*,
Zhixiang Wang,
Tien-Dat Chung,
Zhen Zhang,
Md Mostafijur Rahman, 
Mohsen Ali,
Waqas Sultani,
Aggelos Katsaggelos,
Syed M. Anwar,
Shantanu Ghosh,
Kayhan Batmanghelich,
Pedro R. A. S. Bassi,
Bo Zhou,
Zongwei Zhou,
and Ulas Bagci,~\IEEEmembership{Senior Member,~IEEE}

\thanks{This work has been submitted to the IEEE for possible publication.}

\thanks{
Luu Le, Thanh-Huy Nguyen, Zhixiang Wang, Bo Zhou, Aggelos Katsaggelos, and Ulas Bagci
are with the Department of Radiology,
Feinberg School of Medicine,
Northwestern University,
Chicago, IL 60611, USA.
}
\thanks{
Pedro R. A. S. Bassi and Zongwei Zhou
are with the Department of Computer Science,
Johns Hopkins University,
Baltimore, MD 21218, USA.
}
\thanks{
Tien-Dat Chung is with the Knight Foundation School of Computing and Information Sciences,
Florida International University,
Miami, FL 33174, USA.
}
\thanks{
Md Mostafijur Rahman
is with the Department of Electrical and Computer Engineering,
Texas Tech University,
Lubbock, TX 79409, USA.
}
\thanks{
Shantanu Ghosh and Kayhan Batmanghelich
are with the Department of Electrical and Computer Engineering,
Boston University,
Boston, MA, USA.
}
\thanks{
Syed M. Anwar
is with the Children's National Research Institute, Washington DC, USA.
}
\thanks{
Zhen Zhang is with Department of Radiation Oncology, Zhejiang Cancer Hospital,  Hangzhou, China and Maastro Clinic in Maastricht, Netherlands. He is affiliated with Northwestern's BagciLab too.
}

\thanks{
Mohsen Ali and Waqas Sultani are 
with Information Technology University, Lahore, Pakistan.
}

\thanks{
Corresponding author: Ulas Bagci
(e-mail: ulas.bagci@northwestern.edu).
}
}

\markboth{}%
{Bagci: Deep Prototype Learning: A Unified Taxonomy and Survey in Medical Imaging}


\maketitle


\begin{abstract}
    
Deep learning has fundamentally transformed medical image analysis, yet its deployment in high-stakes clinical workflows remains severely hindered by black-box opacity, massive annotation requirements, and vulnerability to domain shifts. Prototype learning has emerged as a compelling paradigm to overcome these bottlenecks by introducing explicit reference structures, ranging from interpretable visual parts that mirror clinical case-based reasoning to data-efficient semantic anchors. Despite its rapid adoption, the literature remains highly fragmented across different computational formulations and clinical domains. In this paper, we provide a comprehensive survey of prototype learning in medical imaging. We present a general organizing formulation and propose a Representation, Learning Strategy, and Utilization (R-L-U) taxonomy to deconstruct the architectural evolution of prototype-based models. Utilizing this taxonomy, we critically and quantitatively analyze 101 core papers using prototypes in medical imaging, selected from 486 studies included following the PRISMA 2020 guidelines, across diverse clinical tasks, including disease classification and diagnosis, medical image segmentation, survival and prognosis prediction, anomaly detection, cross-modal radiology report generation, and other clinical and anatomical tasks. Notably, only 26 of the 101 core papers provide explicit part- or case-based explanations, and only four of them test whether these explanations are faithful. Finally, we highlight some fundamental challenges and discuss further research directions.

\end{abstract}



\section{Introduction}
\label{sec:introduction}

Deep learning has fundamentally transformed medical image analysis, achieving state-of-the-art performance across a broad spectrum of clinical tasks, including disease screening, lesion detection, anatomical segmentation, image registration, image retrieval, and image generation and enhancement \cite{litjens2017survey, shen2017deep}. In particular, deep learning models have achieved high diagnostic accuracy across specialized domains, such as ophthalmology \cite{ting2019deep}, pulmonology \cite{ma2020survey}, cardiology \cite{chen2020deep}, dermatology \cite{jeong2023deep}, and histopathology \cite{van2021deep}.

\begin{figure}[!t]
    \centering
    \includegraphics[width=\linewidth]{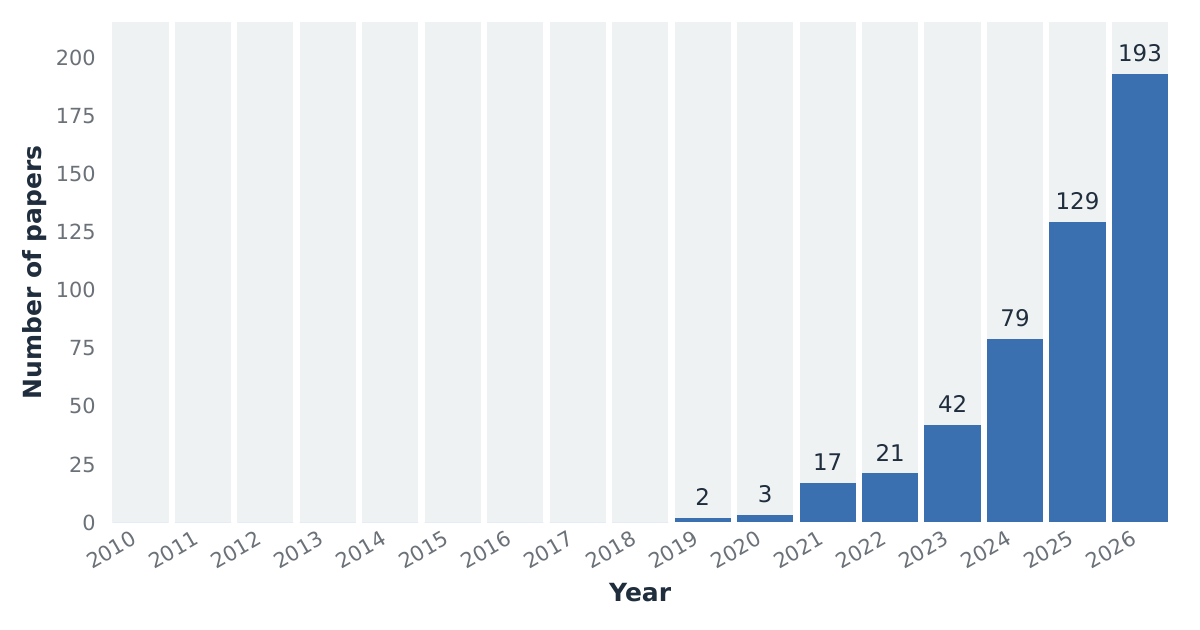}
    \caption{\textbf{Number of papers using prototypes in medical imaging per year} ($N=486$; 2026 covers studies up to November 7, 2026).}
    \label{fig:papers_per_year}
\end{figure}
Despite these successes, the deployment of deep learning systems in real-world clinical settings remains limited by three critical challenges: \emph{black-box opacity, severe annotation scarcity, and vulnerability to domain shifts} \cite{litjens2017survey,salahuddin2022transparency,wang2025deep}. Most high-performing models operate as black boxes, producing predictions without explicit reasoning pathways that clinicians can inspect. In safety-critical domains, this opacity undermines clinical trust and complicates regulatory approval \cite{salahuddin2022transparency,wang2025deep}. Furthermore, training these conventional architectures requires massive cohorts of densely annotated data, which is prohibitively expensive and time-consuming for specialized clinicians to produce \cite{litjens2017survey,wang2025deep}. Finally, even when trained on large datasets, these models frequently exhibit severe performance drops during external validation. These failures stem not only from distribution shifts across patient demographics and hardware vendors \cite{varoquaux2022machine,guan2021domain}, but also from shortcut learning \cite{geirhos2020shortcut}, where models rely on spurious, non-pathological features rather than true disease signs.

\begin{figure*}[!t]
    \centering
    \includegraphics[width=0.95\textwidth]{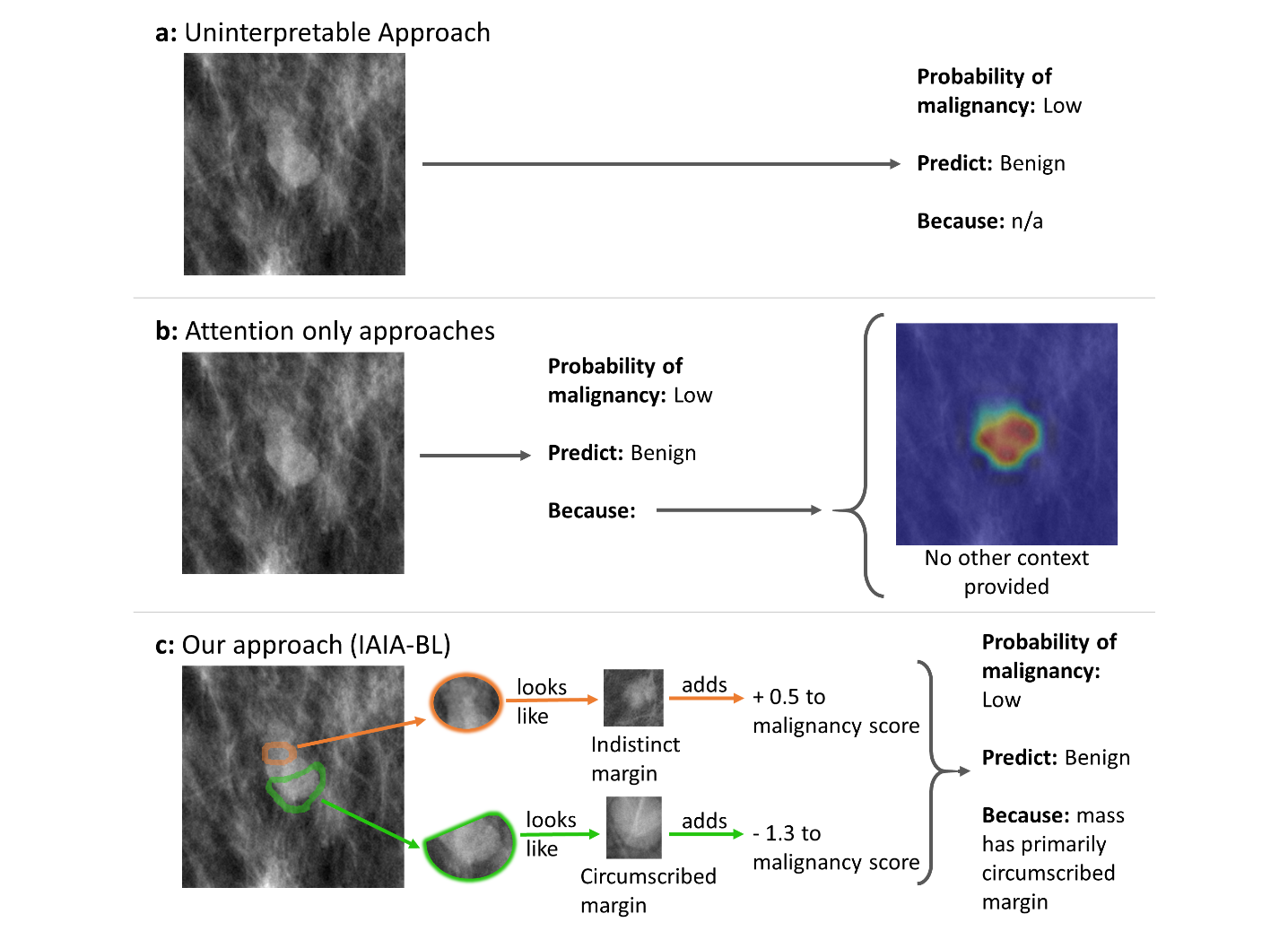}
    \caption{
    \textbf{Illustration of different levels of interpretability in medical image classification.} 
    (a) Conventional black-box models provide predictions without explicit evidence. 
    (b) Attention-based approaches can localize image regions associated with a prediction but may not indicate which clinically meaningful features support the decision. 
    (c) Prototype-based reasoning, illustrated by IAIA-BL \cite{barnett2021iaia}, links localized image regions to representative patterns and uses their evidence to support the final prediction (image from \cite{barnett2021iaia}).
    }
    \label{fig:intro_motivation}
\end{figure*}

Prototype learning has emerged as a compelling paradigm to address these three challenges while preserving the representational power and accuracy of deep neural networks \cite{snell2017prototypical, chen2019this}. Rather than relying solely on abstract latent embeddings, prototype-based models explicitly represent knowledge through a set of \emph{prototypes}, feature vectors, or localized parts 
\cite{li2018deep,chen2019this,donnelly2021deformable}, which can be utilized to visualize disease categories or anatomical structures \cite{barnett2021iaia, wei2024mprotonet}. First, to address opacity, prototype learning transitions beyond uninterpretable predictions and context-free saliency maps by grounding decisions in part-based reasoning (Fig.~\ref{fig:intro_motivation}). As illustrated in the figure, while conventional black boxes offer no rationale and attention heatmaps merely highlight regions without diagnostic meaning, part-based prototype models can decompose the finding into localized parts, match them to reference patterns, and accumulate explicit additive evidence that supports the prediction; whether this evidence is faithful and clinically meaningful must be evaluated separately (Section~\ref{sec:explainability_of_prototypes}) \cite{barnett2021iaia, kim2021xprotonet, huy2025interactive}. Second, to overcome annotation scarcity and long-tailed class distribution, prototypes naturally support few-shot learning and rare disease classification by anchoring predictions to a small set of labeled exemplars rather than requiring dense pixel-wise supervision, even extending to completely label-free anomaly detection \cite{zhu2023few, xiang2024exploiting}. Third, to cope with domain shift, prototypes act as compact anchors that can be re-estimated from a few target-domain samples or aligned across sites and clients without sharing raw images \cite{yu2023source,wang2024pfmnet,zhao2025medfedproto}, and part-based explanations can expose the shortcut features behind poor generalization \cite{gautam2022demonstrating,nauta2023interpreting}. 

Crucially, across these clinical applications, prototype learning divides into two distinct operational regimes that govern the narrative of this survey: (i) \textit{explicitly interpretable architectures}, predominantly deployed in disease classification and grading, where localized part- or case-based prototypes directly expose visual evidence for clinical auditing; and (ii) \textit{representation- and data-efficient mechanisms}, heavily utilized in dense segmentation, anomaly detection, and federated learning, where implicit prototypes act as non-parametric support centroids, momentum-smoothed regularizers, or cross-modal memory banks to combat annotation scarcity and institutional domain shifts.

Despite the rapid growth of prototype learning methods (Fig.~\ref{fig:papers_per_year}), the literature has become highly fragmented. This fragmentation stems from the fact that prototype learning has evolved independently across multiple parallel research communities, such as few-shot learning \cite{snell2017prototypical, wang2019panet}, interpretable deep learning \cite{chen2019this, nauta2021neural}, federated learning \cite{tan2022fedproto,zhao2025medfedproto}, and cross-modal representation learning \cite{chen2021cross, cheng2023prior,zhang2024disease}. Although these methods share the common principle of representing knowledge through prototypes, they employ inconsistent terminology, disjointed mathematical formulations, and diverse architectural designs, making it difficult for practitioners to select or adapt these methodologies for specific clinical workflows.

To resolve the current methodological fragmentation, we propose a unified perspective that interprets prototype learning through three complementary design dimensions:
\begin{itemize}
    \item \textbf{Prototype Representation:} \emph{What information does a prototype encode?} Existing methods employ diverse mathematical forms, spatial granularities, and cardinalities to encode clinical knowledge.
    \item \textbf{Prototype Learning Strategy:} \emph{How are prototypes constructed and optimized?} Representative approaches span discriminative, contrastive, and generative paradigms using various optimization mechanisms.
    \item \textbf{Prototype Utilization:} \emph{How are learned prototypes incorporated into downstream tasks?} Current mechanisms range from fixed distance-based inference to learnable metrics that yield implicit, part-based, or case-based explanations.
\end{itemize}

In summary, the primary contributions of this survey are fourfold:
\begin{itemize}
    \item We present a comprehensive overview of deep prototype learning that systematically traces the technical evolution of its core components from foundational computer vision frameworks to specialized architectures.
    \item We introduce a novel multi-dimensional taxonomy structured around Representation, Learning Strategy, and Utilization (R-L-U) to formalize the design space and categorize existing methodological variations.
    \item We apply our proposed R-L-U taxonomy to map, categorize, and quantitatively analyze 101 core studies selected from 486 medical imaging studies included following the Preferred Reporting Items for Systematic Reviews and Meta-Analyses (PRISMA) 2020 guidelines, offering a structured alignment between algorithmic designs and clinical applications.
    \item We turn this analysis into practical guidelines for choosing a prototype design and for evaluating prototype models and their explanations.

\end{itemize}

\begin{figure*}[!t]
    \centering
    \includegraphics[width=\linewidth]{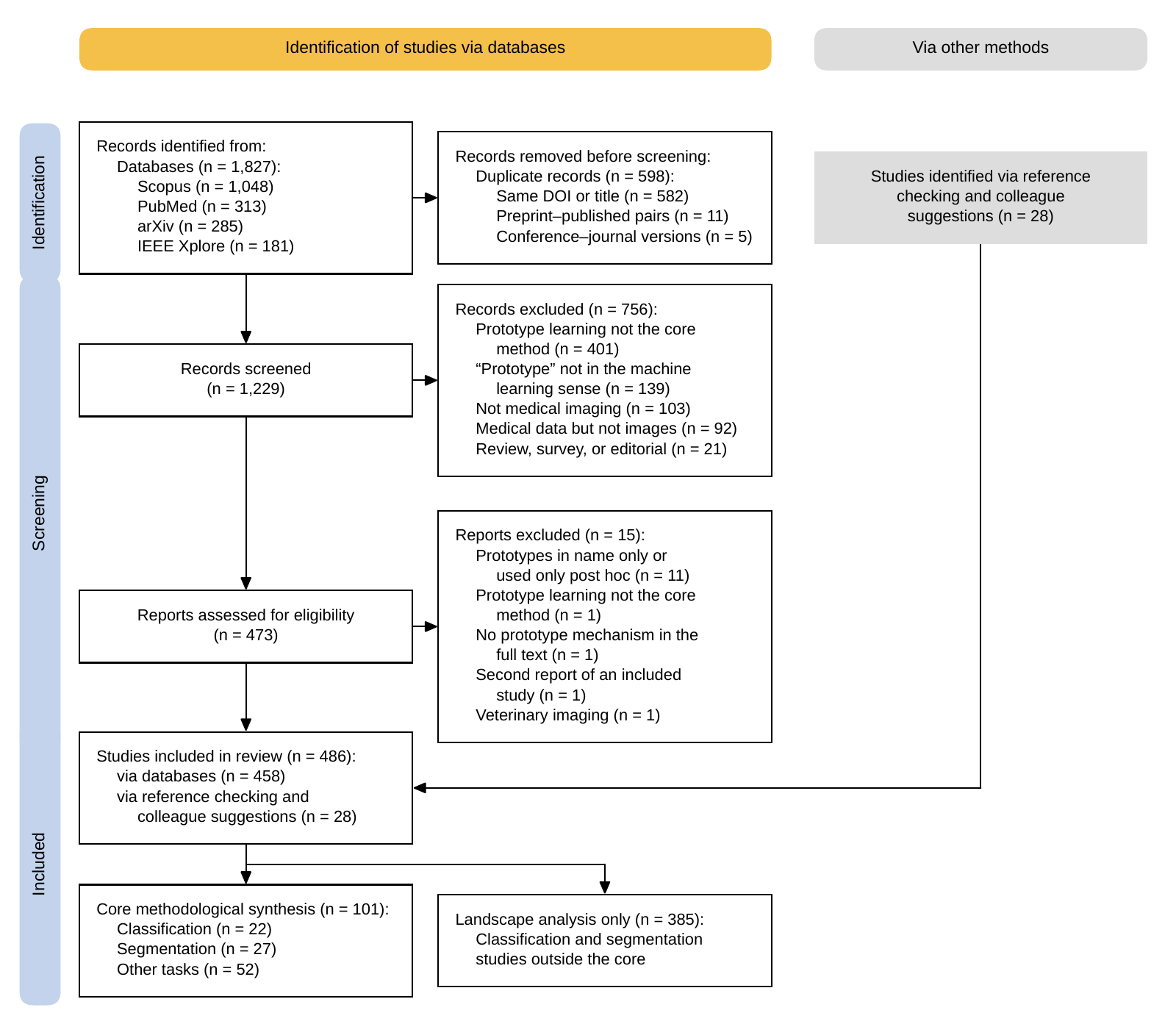}
    \caption{\textbf{PRISMA 2020 flow diagram of the literature selection} \cite{page2021prisma}.}
    \label{fig:prisma_flow}
\end{figure*}
The remainder of this paper is organized as follows. Section~\ref{sec:survey_methodology} defines the survey scope and literature selection. Section~\ref{sec:background} defines the fundamental concepts of prototype learning, outlines the unified mathematical formulation, and introduces the R-L-U taxonomy. Section~\ref{sec:evolution_of_prototype_learning} traces the architectural evolution of prototype methodologies across both general computer vision and medical domains. Section~\ref{sec:medical_image_applications} comprehensively reviews the application of prototype networks categorized by clinical task. Section~\ref{sec:practical_guidelines} turns these findings into practical guidelines for choosing and evaluating prototype designs. Finally, Section~\ref{sec:challenges_future_directions} discusses open challenges and future directions, followed by the conclusion in Section~\ref{sec:conclusion}.

\section{Survey Methodology}
\label{sec:survey_methodology}

\subsection{Survey Scope}
The scope of this survey is organized around two primary goals: (1) to construct a simple yet comprehensive taxonomy (Section~\ref{sec:core_components}) that gives researchers a clear big picture of the prototype learning framework; and (2) to examine how current prototype learning methods are used across clinical settings (Section~\ref{sec:medical_image_applications}). We focus strictly on 2D, 3D, video, and whole-slide medical imaging modalities and only include a multi-modal framework if it applies prototype learning across paired medical images and clinical text, genomics, or tabular electronic health record (EHR) data. Studies restricted solely to tabular EHR data or 1D physiological signals are excluded.

Across all tasks, we define eligibility by mechanism rather than terminology. Specifically, a study is included if it constructs, maintains, or optimizes explicit reference structures (prototypes) to guide training or inference, even when termed exemplars, codebooks, or memory banks. To distinguish these from classical methods, we apply a strict operational rule: a method is included if the construction and comparison of these reference anchors are explicitly integrated into the network's training objective (e.g., episodic few-shot learning) or parameter optimization. Conversely, we exclude ordinary $k$-nearest neighbor ($k$-NN) and content-based image retrieval (CBIR) methods that apply post hoc distance matching using a frozen gallery of embeddings, where the reference structure plays no part in training. Standard linear classifiers and purely post hoc saliency methods are also excluded. The last update to the included papers was on November 7, 2026. 

\subsection{Literature Selection}
 To translate this scope into a concrete corpus, we searched Scopus, PubMed, IEEE Xplore, and arXiv using a unified core search logic adapted to the specific syntax of each database. The comprehensive search string combined prototype-centric terminology with medical imaging modalities in titles and abstracts:
\begin{quote}
\small\raggedright
(``prototype-based'' OR ``ProtoPNet'' OR ``part-prototype'' OR ``prototype-guided'' OR ``prototype learning'' OR ``prototypical network'' OR ``class prototype'' OR ``prototype alignment'' OR ``memory bank'' OR ``codebook'' OR ``case-based reasoning'') AND (``medical image'' OR ``medical imaging'' OR ``MRI'' OR ``magnetic resonance'' OR ``CT'' OR ``computed tomography'' OR ``ultrasound'' OR ``echocardiography'' OR ``x-ray'' OR ``radiograph'' OR ``mammography'' OR ``fundus'' OR ``OCT'' OR ``dermoscopy'' OR ``pathology'' OR ``histopathology'' OR ``whole-slide'' OR ``WSI'')
\end{quote}
All searches covered 2010 to 2026. In Scopus, results were limited to English-language articles, conference papers, and book chapters. We intentionally excluded generic nearest-neighbor or metric-learning terms without explicit prototype keywords to avoid an enormous number of false-positive hits from classical pattern recognition.
 
The selection follows the PRISMA 2020 guidelines \cite{page2021prisma} (Fig.~\ref{fig:prisma_flow}). Our searches retrieved 1,827 records (Scopus 1,048, IEEE Xplore 181, PubMed 313, arXiv 285). After removing 598 duplicates, 1,229 records were screened by title and abstract, and 756 were excluded: ``prototype'' not used in the machine learning sense (139), prototype learning not the core method (401), no medical imaging (103), medical data that are not images (92), and reviews or editorials (21). This yielded 473 studies for full-text assessment, of which 15 were excluded, 11 of them because their prototypes were used in name only or only post hoc; the other four were a study in which prototype learning was not the core method, a study with no prototype mechanism in the full text, a second report of an included study, and a study on veterinary images, leaving 458 eligible studies. Additionally, we iteratively checked references across all selected studies and consulted colleagues to identify underlying general vision frameworks (forming our foundational taxonomy in Section~\ref{sec:core_components}) as well as clinical studies using alternative terminology missed by keyword search. Through this, we manually added 28 highly relevant medical prototype studies. In total, 486 studies were included. Two authors screened and coded the studies, and any case they could not resolve, including ambiguous taxonomic codes, was settled through discussion and consensus with a senior author. When largely overlapping work appeared across multiple venues (e.g., as a preprint, a conference proceeding, and a journal extension), only the most complete and recent journal publication was retained to prevent duplicate counting. To ensure a transparent selection process, a machine-readable extraction sheet is provided in the Supplementary Material.
 
Following eligibility screening, we observed strong task saturation, with classification (234 studies) and segmentation (200 studies) comprising about nine-tenths of the eligible literature. To prevent routine clinical adaptations in these two dominant categories from overshadowing architectural taxonomy and emerging clinical applications, we implemented a two-tier synthesis protocol. All eligible studies were retained for macro-level bibliometric and clinical task mapping, after which a methodological core filter was applied. Specifically, within classification and segmentation, we included studies if they met all five criteria. (i) Prototype contribution: the prototype mechanism is the main contribution, or one of the main named contributions with its own ablation, rather than off-the-shelf class means or centroids plugged into a standard pipeline (e.g., routine prototype alignment or prototype pseudo-labeling with no new prototype rule). (ii) Methodological innovation: the study changes the prototype design on at least one axis of our taxonomy (Section~\ref{sec:core_components}), namely its representation, learning rule, or use at inference. We define an incremental application as a study that applies an existing named design (e.g., ProtoNet, ProtoPNet, or PIP-Net) to a new organ, modality, or dataset while changing only the backbone, preprocessing, or hyperparameters, or that repeats a prototype idea already present in the core without a substantive change. (iii) Archival evidence: the study is neither an arXiv-only preprint nor a workshop paper, and it is not evaluated on private data alone. (iv) Venue: a Q1 journal (SJR), an A*/A conference (CORE), or a core medical imaging conference (MICCAI, IPMI, MIDL, ISBI, EMBC). Lastly, (v) Evidence: the study compares against prior methods and reports an ablation that removes or alters the prototype component. Conversely, to capture the full breadth of emerging clinical directions, all eligible studies across the remaining specialized tasks, including preprints, were retained for in-depth qualitative synthesis.

The resulting 101 core studies (22 classification, 27 segmentation, and 52 on other tasks) are discussed in Section~\ref{sec:medical_image_applications}, listed in Table~\ref{tab:medical_applications}, and coded in Appendix Table~\ref{tab:appendix_rlu_coding}. Because the core is designed to capture methodological contributions, its counts describe the range of prototype designs that have been proposed, not how often each design is used in routine applications.
 
\subsection{Quantitative Analysis}
\label{sec:quantitative_analysis}

\begin{figure}[!t]
    \centering
    \includegraphics[width=\linewidth]{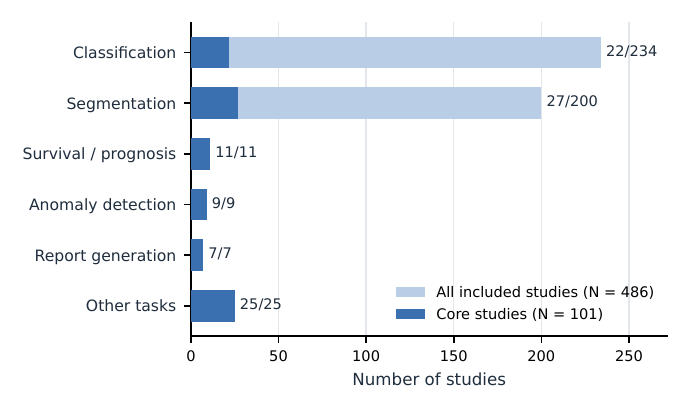}
    \caption{\textbf{Number of studies per clinical task}. Each bar shows all included studies in each task, and the dark segment shows the core studies; labels give core/total. ``Other tasks'' covers regression, image synthesis, representation learning and pre-training, open-set recognition and out-of-distribution (OOD) detection, lesion and cell detection, landmark localization, retrieval, registration, and MR reconstruction.}
    \label{fig:studies_per_task}
\end{figure}

Before reviewing individual methods, we summarize the 486 included studies by publication year, clinical task, and literature discovery source.

\textbf{Publication trend.} Prototype learning in medical imaging is a recent field. The first included studies appeared in 2019, and the number of studies has grown rapidly since 2023, from 42 in 2023 to 79 in 2024 and 129 in 2025 (Fig.~\ref{fig:papers_per_year}). Overall, 401 of the 486 studies (82.5\%) were published from 2024 onward.

\textbf{Clinical tasks.} The literature is concentrated in two tasks: classification (234 studies) and segmentation (200 studies) together account for 89.3\% of all studies (Fig.~\ref{fig:studies_per_task}). Only 49 of these 434 studies (11.3\%) meet the core criteria, as many apply established prototype architectures to new datasets without changing the prototype mechanism. The remaining 52 studies cover more specialized tasks, led by survival and prognosis prediction (11) and anomaly detection (9), followed by report generation (7), regression (5), detection and landmark localization (4), representation learning (4), image synthesis, open-set recognition, and retrieval (3 each), registration (2), and MR reconstruction (1); all of them are included in the core.

\textbf{Literature Discovery sources.} Each study is credited to the first database that returned it, in the order Scopus, IEEE Xplore, PubMed, and arXiv. Most studies came from Scopus (391), followed by arXiv (45), IEEE Xplore (12), and PubMed (10); a further 28 studies were found only through reference checking and colleague suggestions.

These observations shape the rest of this survey: Section~\ref{sec:medical_image_applications} traces the main methodological lines in classification and segmentation through representative core studies and covers the less-studied tasks in full.
 
\subsection{Related Surveys}
 
Several recent surveys have examined topics adjacent to our scope, though each focuses on a single paradigm rather than prototype learning as a unified field. In general machine learning, Ji et al.~\cite{ji2025comprehensive} survey self-interpretable neural networks, while Nauta et al.~\cite{nauta2023anecdotal} review quantitative evaluation properties for explainable AI. Narrowing to the medical explainable AI domain, Patr\'{i}cio et al.~\cite{patricio2023explainable} and Hou et al.~\cite{hou2026self} review explainable and self-explainable methods for medical image analysis, treating part-prototype networks as one branch alongside saliency and concept-based models. Conversely, in the few-shot learning regime, Pachetti and Colantonio~\cite{pachetti2024systematic} review few-shot learning in medical imaging, while Dissanayake et al.~\cite{dissanayake2025few} focus specifically on few-shot medical image segmentation.
 
\begin{table*}[!t]
\centering
\footnotesize
\renewcommand{\arraystretch}{1.2}
\caption{\textbf{Comparison between our survey and related surveys on explainable AI, few-shot learning, and part-prototype models}. To the best of our knowledge, our survey is the first to treat prototype learning as a unified field in medical imaging and to organize it through a dedicated design taxonomy. \textit{Search transparency}: the search strategy and selection counts are reported; $\circ$ = partially.}
\label{tab:related_surveys}
\setlength{\tabcolsep}{3pt}
\begin{tabular*}{\textwidth}{@{\extracolsep{\fill}}lccccccc}
\toprule
\textbf{Survey} & \textbf{Year} & \textbf{Medical} & \textbf{\makecell{Interpret-\\ability}} & \textbf{Few-shot} & \textbf{\makecell{Prototype-\\centered}} & \textbf{\makecell{Prototype\\taxonomy}} & \textbf{\makecell{Search\\transparency}} \\
\midrule
Nauta et al.~\cite{nauta2023anecdotal} & 2023 & $\times$ & \checkmark & $\times$ & $\times$ & $\times$ & \checkmark \\
Patr\'{i}cio et al.~\cite{patricio2023explainable} & 2023 & \checkmark & \checkmark & $\times$ & $\times$ & $\times$ & $\times$ \\
Pachetti and Colantonio~\cite{pachetti2024systematic} & 2024 & \checkmark & $\times$ & \checkmark & $\times$ & $\times$ & \checkmark \\
Ji et al.~\cite{ji2025comprehensive} & 2025 & $\times$ & \checkmark & $\times$ & $\times$ & $\times$ & $\times$ \\
Elhadri et al.~\cite{elhadri2025looks} & 2025 & $\times$ & \checkmark & $\times$ & \checkmark & $\times$ & \checkmark \\
Dissanayake et al.~\cite{dissanayake2025few} & 2025 & \checkmark & $\times$ & \checkmark & $\times$ & $\times$ & $\circ$ \\
Hou et al.~\cite{hou2026self} & 2026 & \checkmark & \checkmark & $\times$ & $\times$ & $\times$ & $\circ$ \\
\midrule
\textbf{Ours} & 2026 & \checkmark & \checkmark & \checkmark & \checkmark & \checkmark & \checkmark \\
\bottomrule
\end{tabular*}
\end{table*}
 
While these surveys provide valuable perspectives, they do not center on prototype learning as a unified field in medical imaging as summarized in Table~\ref{tab:related_surveys}. Instead, they either overlook the clinical domain or confine prototype learning to narrow, isolated niches rather than a unified field.

Closest to our topic, Elhadri et al.~\cite{elhadri2025looks} review part-prototype models in general computer vision and organize their open challenges, including the quality and number of prototypes, generalization, and non-standardized evaluation. Our survey differs in three ways: it covers medical imaging across tasks beyond classification, it includes implicit prototypes (support-derived, momentum-updated, and memory-based) in addition to explicitly interpretable part-prototype models, and it codes each core study with a reproducible R-L-U design taxonomy rather than organizing open challenges. Our survey offers an up-to-date, prototype-centered synthesis that connects disparate directions under our taxonomy, providing researchers with a complete picture of how prototype learning is designed, evaluated, and deployed across medical imaging.

\section{Background and Unified Mathematical Framework}
\label{sec:background}

Prototype learning has a long history in pattern recognition and nearest-neighbor classification \cite{liu2001evaluation}. Early prototype-based methods represented classes using representative vectors or codebook prototypes, as exemplified by Learning Vector Quantization (LVQ) \cite{kohonen1990self} and vector quantization methods \cite{linde1980algorithm}. With the rise of deep learning, however, prototypes have increasingly been represented in learned latent spaces, taking diverse forms and operating at different levels of abstraction. Early deep prototype methods used global feature embeddings as class representatives \cite{vinyals2016matching,snell2017prototypical,sung2018learning,li2018deep}, while subsequent approaches introduced localized or region-level prototypes to capture characteristic parts of an input \cite{van2017neural,dong2018few,chen2019this,nauta2021neural}. Other methods extended the representation beyond deterministic point prototypes, including probabilistic prototypes \cite{scott2019stochastic}, shared prototype dictionaries \cite{van2017neural,nauta2021neural}, and multiple prototypes or multi-scale prototype representations \cite{vinyals2016matching,li2018deep,chen2019this,li2020pcl,wang2021exploring}. These developments broaden the role of prototypes from simple representative points toward more structured representations that can capture different aspects of the underlying data.

Rather than providing an exhaustive literature review, this section establishes a foundational framework for the remainder of this survey. We begin by defining prototypes and outlining the general prototype learning pipeline. We then introduce a general organizing formulation that categorizes methods across three core components: representation, learning strategy, and utilization. Finally, we discuss related learning paradigms and highlight the importance of prototype learning in medical imaging.

\subsection{What Is Prototype Learning?}
\label{sec:what_is_prototype_learning}

The term \emph{prototype} traditionally refers to a representative example that captures the characteristic properties of a group of observations. In classical pattern recognition, a prototype may be an actual training sample, a class centroid, or a representative vector learned from data. A new sample can then be classified by measuring its similarity or distance to the prototypes associated with the candidate classes \cite{kohonen1990self,linde1980algorithm}.

In modern deep learning, prototypes are no longer restricted to raw training samples or simple centroids. Instead, they may live in a learned feature space and be optimized jointly with a neural network. In the literature, this learned representation space is variously referred to as the \emph{latent space} \cite{chen2019this}, \emph{embedding space} \cite{snell2017prototypical}, \emph{feature space} \cite{nauta2021neural}, or \emph{projected space} \cite{dong2018few}, depending on the formulation, to describe the $d$-dimensional space in which encoded inputs and prototypes are represented and compared.

For example, a prototype may represent a discriminative image patch in an interpretable classifier \cite{chen2019this}, a class-level mean embedding in few-shot learning \cite{snell2017prototypical}, or a sentence-level clinical finding in vision-language representation learning \cite{cheng2023prior}. These different forms of prototypes vary in their representation, but they share the same underlying role: they provide explicit reference structures against which new inputs can be compared.

In this survey, we define prototype learning as follows:

\begin{quote}
\emph{Prototype learning is a paradigm characterized \textbf{by the presence of explicit reference structures}, which summarize characteristic patterns shared by subsets of observations and serve as reference anchors within the learning or inference process.}
\end{quote}

Crucially, this definition distinguishes prototype learning from standard parametric deep learning. While conventional neural networks project latent features through unconstrained linear layers into abstract decision hyperplanes, prototype learning grounds predictions, auxiliary training objectives, or distributed client communication in direct comparisons against identifiable reference entities. In all cases, the defining boundary \emph{\textbf{lies in the presence of these explicit reference structures}} rather than unconstrained parameter projections. However, the comparison alone does not make a model more interpretable than a linear classifier; interpretability instead comes from how prototypes are constrained, grounded, and validated (Section~\ref{sec:explainability_of_prototypes}), not from the use of prototypes alone.

From this perspective, prototype learning should not be understood as a single algorithm but as a family of approaches that integrate representation learning with exemplar-based reasoning. By replacing abstract decision hyperplanes with explicit comparisons against representative anchors, prototype models offer distinct advantages over standard black-box baselines: they enable intrinsic interpretability \cite{chen2019this}, improve sample efficiency under limited supervision \cite{snell2017prototypical,vinyals2016matching}, and provide intuitive mechanisms for model inspection and clinical auditing \cite{barnett2021iaia,kim2021xprotonet}. These properties are particularly valuable in medical imaging, where diagnostic decisions must often be justified, expert annotations are scarce, and models must remain reliable under substantial cross-institutional distribution shifts.

\begin{figure*}[!t]
    \centering
    \includegraphics[width=1\linewidth]{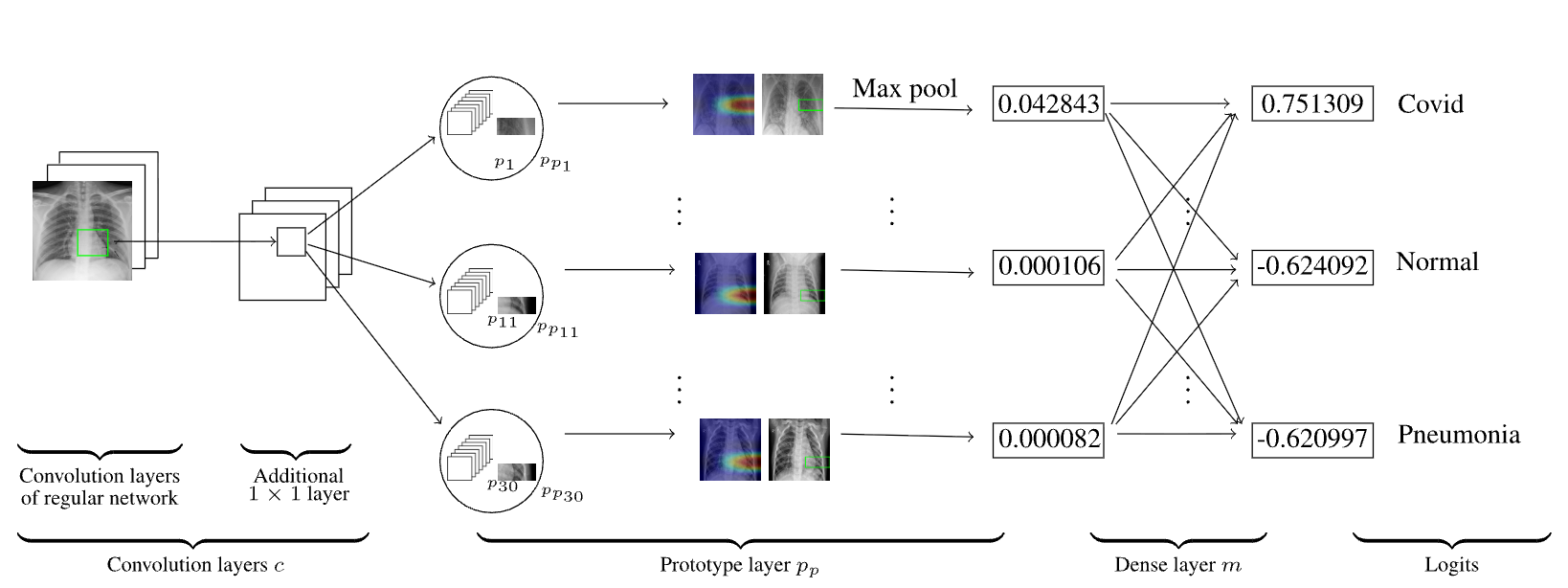}
    \caption{
    \textbf{General prototype learning pipeline in medical image classification.} 
    Using a chest X-ray model as an example, the pipeline consists of: feature extraction via convolutional layers; a prototype layer where the encoded features are compared against a set of learned prototypical parts (producing spatial similarity heatmaps); an aggregation step (max pooling) to quantify the presence of each prototype; and a final prediction stage where a fully connected dense layer maps these similarity scores to diagnostic logits (e.g., COVID, Normal, Pneumonia) (image from \cite{singh2021interpretable}).
    }
    \label{fig:overall_pipeline_of_prototype_learning}
\end{figure*}

\subsection{General Prototype Learning Pipeline}
\label{sec:prototype_pipeline}

Although prototype learning methods differ substantially in architecture and application, their overall operation can be described through a unified general pipeline, an example of which is illustrated for chest X-ray classification in Fig.~\ref{fig:overall_pipeline_of_prototype_learning}. Given an input $x$, the model first extracts a feature representation using an encoder. A set of prototypes is then constructed or maintained from the training data, learned parameters, or a separate update mechanism. The input representation is subsequently related to the prototype set through a comparison or interaction mechanism, and the resulting prototype-related information is used to produce the task-specific output.

Conceptually, the pipeline can be summarized as the sequence: $x \rightarrow$ feature extraction $\rightarrow$ prototype construction $\rightarrow$ prototype utilization $\rightarrow$ prediction $\rightarrow \hat{y}$. Each stage can be implemented in different ways depending on the task and model architecture. Specifically:

\begin{itemize}
    \item \textbf{$x$ -- Input.} The input may be a 2D image, 3D volume, histopathology image, whole-slide image, or multi-modal medical data.
    \item \textbf{Feature extraction.} The input is transformed into a feature representation using a suitable neural network, such as a CNN-based or Transformer-based encoder.
    \item \textbf{Prototype construction.} Prototypes can be computed from training or support samples, learned as model parameters, or obtained and updated using mechanisms such as clustering, on-the-fly computation, or exponential moving averages.
    \item \textbf{Prototype utilization.} The extracted features are related to the prototypes through a comparison or matching mechanism, such as a predefined distance, similarity function, or learned metric.
    \item \textbf{Prediction.} The prototype-related information is transformed into the final task output, such as a class label, segmentation mask, or other task-specific prediction.
\end{itemize}

The specific implementation of these stages varies substantially across prototype learning methods. The following section summarizes this pipeline through a general organizing formulation to describe these different design choices.

\subsection{General Organizing Formulation}
\label{sec:unified_formulation}

The general prototype learning pipeline described in Section~\ref{sec:prototype_pipeline} can be described using a general organizing formulation. Let $\mathcal{D}=\{(x_{i},y_{i})\}_{i=1}^{N}$ denote a training dataset, where $x_{i}\in\mathcal{X}$ is an input and $y_{i}\in\mathcal{Y}$ denotes the corresponding task target. When prototypes remain during the inference phase, the model prediction is written as
\begin{equation}
\hat{y}=g_{\phi}(S(f_{\theta}(x),\mathcal{P})),
\label{eq:unified_function}
\end{equation}
where $f_{\theta}$ denotes the feature encoder, $\mathcal{P}$ denotes the prototype set, $S$ represents the comparison or interaction between the input representation and the prototypes, and $g_{\phi}$ is the task-specific prediction function. In some models, a non-prototype pathway that bypasses $S$, such as a global classifier branch~\cite{wang2022knowledge}, also feeds the prediction; for these models, Eq.~\ref{eq:unified_function} describes only the prototype pathway. For methods that use prototypes only during training and discard them at inference \cite{li2020pcl,wang2021exploring,caron2020unsupervised,zhang2024disease,xu2025medical}, prototype-dependent training enters through the training objective (Eq.~\ref{eq:unified_loss}), whereas inference uses prototype-free prediction:
\begin{equation}
\hat{y}=g_{\phi}(f_{\theta}(x)).
\label{eq:prototype_free_prediction}
\end{equation}

The prototype set can be represented or obtained in different ways. For example, a prototype may be computed on-the-fly from a subset or support set $\mathcal{C}_k$ of training representations as
\begin{equation}
    \mathbf{p}_k(\theta) = \frac{1}{|\mathcal{C}_k|} \sum_{x_i\in\mathcal{C}_k} f_{\theta}(x_i),
    \label{eq:prototype_data_derived}
\end{equation}
where $\mathcal{C}_k$ may denote a class, cluster, support set, or other subset of samples \cite{snell2017prototypical}. In other approaches, prototypes are treated as learnable parameters in a $d$-dimensional latent space:
\begin{equation}
    \mathcal{P} = \{\mathbf{p}_k\in\mathbb{R}^{d}\}_{k=1}^{K},
    \label{eq:learnable_prototypes}
\end{equation}
and optimized jointly with the model parameters \cite{van2017neural,li2018deep}.

The prototype comparison function $S$ determines how the input representation is related to the prototype set. When prototypes are compared individually, this interaction can be written as
\begin{equation}
    s_k = S(\mathbf{z},\mathbf{p}_k), \qquad \mathbf{z}=f_{\theta}(x),
    \label{eq:prototype_comparison}
\end{equation}
where $s_k$ denotes the similarity or compatibility between the input representation $\mathbf{z}$ and the $k$-th prototype $\mathbf{p}_k$.

The model is trained using an overall objective that can be expressed generically as
\begin{equation}
    \mathcal{L} = \mathcal{L}_{\mathrm{task}} + \lambda_{\mathrm{proto}} \mathcal{L}_{\mathrm{proto}} + \lambda_{\mathrm{reg}} \mathcal{L}_{\mathrm{reg}},
    \label{eq:unified_loss}
\end{equation}
where $\mathcal{L}_{\mathrm{task}}$ denotes the primary task objective (e.g., cross-entropy loss for classification or Dice loss \cite{milletari2016v} for segmentation), $\mathcal{L}_{\mathrm{proto}}$ denotes a prototype-specific learning objective when explicitly used, and $\mathcal{L}_{\mathrm{reg}}$ denotes additional regularization terms.

The prototype-specific term $\mathcal{L}_{\mathrm{proto}}$ represents learning objectives that impose desired geometric properties on prototypes and their associated feature representations. For example, a common instantiation of $\mathcal{L}_{\mathrm{proto}}$ is a compactness (or clustering) objective, which encourages each training sample to remain close to at least one prototype associated with the same class \cite{li2018deep,chen2019this}. In this case, the prototype objective is defined as:
\begin{equation}
    \mathcal{L}_{\mathrm{proto}} = \frac{1}{N} \sum_{i=1}^{N} \min_{\mathbf{p}_k\in\mathcal{P}_{y_i}} \delta\!\left( f_{\theta}(x_i),\mathbf{p}_k \right),
    \label{eq:compactness_loss}
\end{equation}
where $\mathcal{P}_{y_i}$ denotes the prototypes associated with the target class $y_i$ and $\delta(\cdot,\cdot)$ is a distance function such as Euclidean distance or cosine distance.

Importantly, defining the overall loss $\mathcal{L}$ describes the geometric learning objectives, but it does not imply that the prototype set $\mathcal{P}$ is updated exclusively through gradient descent. To formalize this distinction, let $\Theta = \{\theta, \phi, \psi\}$ denote the trainable parameters of the neural network (including any learnable comparison parameters $\psi$), which are updated via standard gradient descent at iteration $t$ with learning rate $\eta_{\Theta}$:
\begin{equation}
    \Theta^{(t+1)} \leftarrow \Theta^{(t)} - \eta_{\Theta} \nabla_{\Theta} \mathcal{L}.
    \label{eq:theta_update}
\end{equation}
In contrast, the prototype set is obtained or updated via a generalized update function $\Phi$:
\begin{equation}
    \mathcal{P}^{(t+1)} \leftarrow \Phi\left(\mathcal{P}^{(t)}, f_\theta(\mathcal{D}), \nabla_{\mathcal{P}}\mathcal{L}\right),
    \label{eq:generalized_update}
\end{equation}
where $f_\theta(\mathcal{D}) = \{f_\theta(x_i)\}_{i=1}^{N}$ denotes the encoded feature representations of the training dataset (or current mini-batch). The formulation of $\Phi$ separates four prototype regimes:
\begin{enumerate}
    \item \textbf{Persistent trainable prototype parameters:} Prototypes are maintained as independently optimized parameters (Eq.~\ref{eq:learnable_prototypes}) updated via gradient-based optimization with a prototype-specific learning rate $\eta_{\mathcal{P}}$, where $\mathcal{P}^{(t+1)} \leftarrow \mathcal{P}^{(t)} - \eta_{\mathcal{P}} \nabla_{\mathcal{P}}\mathcal{L}$ \cite{van2017neural,li2018deep,chen2019this,arik2019protoattend}.
    \item \textbf{Differentiable, data-derived prototypes:} For a support-derived prototype $\mathbf{p}_k(\theta)$ computed on-the-fly from a support set $\mathcal{C}_k$ (Eq.~\ref{eq:prototype_data_derived}), there is no independently optimized prototype parameter. In methods such as Prototypical Networks \cite{snell2017prototypical} and few-shot segmentation frameworks \cite{dong2018few,wang2019panet,ouyang2020self}, these support means are part of the differentiable computation used to train the embedding.
    \item \textbf{Detached state estimates:} Prototypes have their gradients detached ($\nabla_{\mathcal{P}}\mathcal{L}$ is zeroed), and $\Phi$ updates $\mathcal{P}$ via state-estimation rules, such as particular exponential moving average (EMA) momentum updates \cite{li2021mopro,fan2022few,zhao2025medfedproto} or clustering implementations \cite{li2020pcl,yan2024buffalo}.
    \item \textbf{Frozen references:} Prototypes are prescribed prior to representation learning and kept fixed during model training ($\mathcal{P}^{(t+1)} \leftarrow \mathcal{P}^{(0)}$) without parameter gradients or data-driven updates \cite{mettes2019hyperspherical,yang2023neural,yu2023source,liu2026protosurv}.
\end{enumerate}
Thus, the learning objective $\mathcal{L}$ determines \emph{what} the model is encouraged to learn, whereas the update mechanism $\Phi$ determines \emph{how} the prototype representations are formed and evolved.

Together, these formulations provide the mathematical basis for the R-L-U framework introduced in the following section.

\subsection{Core Components of Prototype Architecture}
\label{sec:core_components}

\begin{figure*}[!t]
    \centering
    \includegraphics[width=1\textwidth]{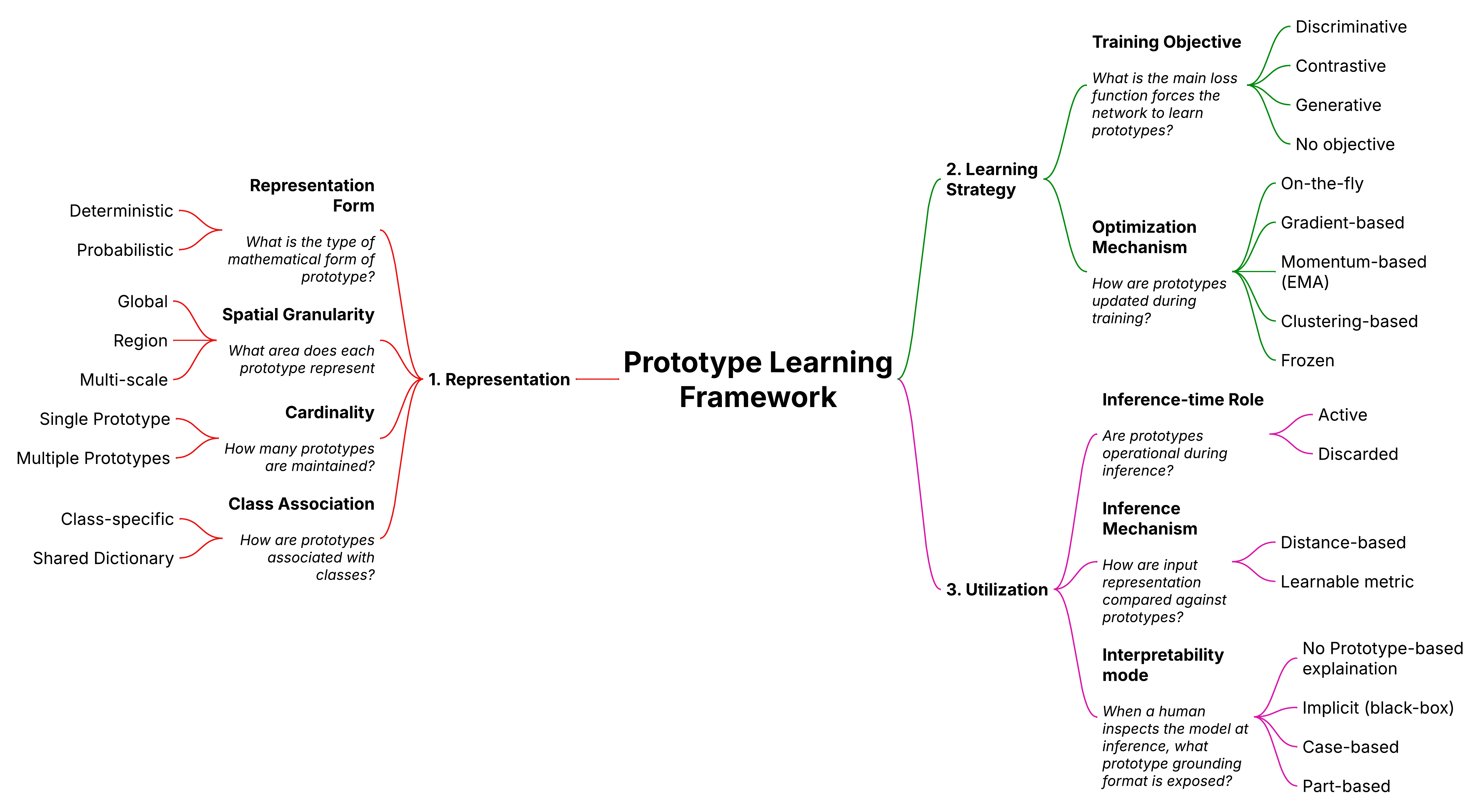}
    \caption{
    \textbf{Proposed taxonomy of prototype learning.} We organize existing prototype learning methods according to three complementary design dimensions: Prototype Representation, Learning Strategy, and Prototype Utilization. This taxonomy provides a unified perspective for analyzing prototype frameworks across different research communities and serves as the conceptual blueprint for the remainder of this survey.
    }
    \label{fig:taxonomy_tree}
\end{figure*}

The general organizing formulation presented in Section~\ref{sec:unified_formulation} suggests that prototype learning methods can be understood through three complementary design dimensions: Prototype Representation, Learning Strategy, and Prototype Utilization. We refer to this as the \emph{R-L-U design space} of prototype learning, where R denotes the prototype representation (dictating the structure of $\mathcal{P}$), L denotes the prototype learning strategy (dictating the objective $\mathcal{L}$ and update rule $\Phi$), and U denotes the prototype utilization mechanism (dictating inference presence, $S$, and $g_\phi$).

To comprehensively categorize how the medical imaging literature has adapted these variables, we deconstruct the R-L-U framework into a three-dimensional taxonomy tree, illustrated in Fig.~\ref{fig:taxonomy_tree}. Prototype methodologies in this review, including hybrid architectures that combine multiple design choices, can be characterized by mapping their components to these dimensions.

\subsubsection{Prototype Representation (R)}
The first dimension asks \emph{what} mathematical, spatial, and associative footprint the prototype encodes. This dimension determines the interpretability, flexibility, and clinical capacity of the prototype. We categorize representation across four axes:
\begin{itemize}
    \item \textbf{By Representation Form:} \emph{What is the mathematical form of a prototype?} It may be modeled as a point-based \emph{Deterministic} vector (e.g., Eq.~\ref{eq:learnable_prototypes}) or a \emph{Probabilistic} distribution (e.g., a Gaussian, Eq.~\ref{eq:probabilistic_prototype}) to capture feature variance and uncertainty.
    \item \textbf{By Spatial Granularity:} \emph{What area does each prototype represent?} The prototype may encode a \emph{Global} scan, a \emph{Region} (localized area), or a \emph{Multi-scale} hierarchy.
    \item \textbf{By Cardinality:} \emph{How many prototypes are maintained?} A target category may be represented by a \emph{Single Prototype} ($K=1$) or \emph{Multiple Prototypes} ($K>1$) to capture intra-class visual variance and multi-modal distributions.
    \item \textbf{By Class Association:} \emph{How are prototypes associated with classes?} Prototypes may be \emph{Class-Specific} (exclusively owned by individual categories or anatomical targets) or organized as a \emph{Shared Dictionary} of class-agnostic features reused across multiple classes.
\end{itemize}

\subsubsection{Prototype Learning Strategy (L)}
Broadly, the second dimension asks \emph{how} prototypes are supervised and updated during training. It further examines their adaptability to noisy medical data and their stability throughout the optimization process:
\begin{itemize}
    \item \textbf{Training Objective:} \emph{What is the main loss function that forces the network to learn prototypes?} It defines the geometric goal of the prototype-supervising objective within the overall loss function $\mathcal{L}$ (Eq.~\ref{eq:unified_loss}). Because multiple losses can coexist (e.g., pairing a discriminative task loss $\mathcal{L}_{\mathrm{task}}$ with a contrastive prototype loss $\mathcal{L}_{\mathrm{proto}}$), we apply a reproducible primary-objective rule based on how the prototype set $\mathcal{P}$ itself is supervised: coded as \emph{Generative} when prototypes condition a decoder to reconstruct images, features, or text; \emph{Contrastive} when $\mathcal{L}_{\mathrm{proto}}$ explicitly optimizes positive- versus negative-pair prototype separation (even alongside a discriminative task loss); \emph{Discriminative} when prototypes are supervised solely through downstream task prediction losses and distance margins; and \emph{No objective} when prototypes are frozen, since no loss updates them.
    \item \textbf{Optimization Mechanism:} \emph{How are prototypes updated during training?} It defines the continuous update rule $\Phi$ (Eq.~\ref{eq:generalized_update}) active during the main training loop (which distinguishes it from pre-training initialization). If prototypes are not persistent parameters across iterations but are computed dynamically from current support or batch samples, they are coded as \emph{On-the-fly}; if prototypes are persistent learnable parameters updated via backpropagation, they are coded as \emph{Gradient-based}; if persistent prototypes have detached gradients and are updated via exponential moving averages, they are coded as \emph{Momentum-based (EMA)}; if updated via cluster assignment steps, they are coded as \emph{Clustering-based}; and if prescribed prior to training without updates, they are coded as \emph{Frozen}.
\end{itemize}

\subsubsection{Prototype Utilization (U)}
The third dimension asks \emph{whether} and \emph{how} prototypes are operationalized during inference (Eq.~\ref{eq:unified_function} vs. Eq.~\ref{eq:prototype_free_prediction}). This dimension determines the computational role of prototypes inside the model and how prototype-based explanations are exposed to clinicians across three axes:
\begin{itemize}
    \item \textbf{Inference-Time Role:} \emph{Are prototypes operational during inference?} This binary axis records whether prototypes remain operational in the test-time prediction pathway (\emph{Active}, Eq.~\ref{eq:unified_function}) or are removed after training (\emph{Discarded}, Eq.~\ref{eq:prototype_free_prediction}) when used solely for training regularization.
    \item \textbf{Comparison Mechanism:} \emph{How are input representations compared against prototypes?} Interaction between encoded features and prototypes is achieved either via predefined \emph{Distance-based} metric calculations (Eqs.~\ref{eq:euclidean_similarity}--\ref{eq:cosine_similarity}) or via a trainable \emph{Learnable metric} (Eq.~\ref{eq:learnable_metric}).
    \item \textbf{Interpretability Mode:} \emph{When a human inspects the model at inference, what prototype grounding format is exposed?} We record the architectural grounding format separately from empirical explanation validity according to four mutually exclusive rules governed by the prototype's inference status and spatial granularity: (i) when prototypes are \emph{Discarded} after training, the model provides \emph{No prototype-based explanation}; (ii) when prototypes are active at inference (\emph{Active}) but no explicit prototype explanation is exposed, the model is \emph{Implicit}; (iii) when predictions are explained through retrieved whole patient examples (\emph{Global} granularity), the model is \emph{Case-based}; and (iv) when predictions are explained through localized image regions or patches (\emph{Region} or \emph{Multi-scale} granularity), the model is \emph{Part-based}.
\end{itemize}

This R-L-U design-space perspective is highly advantageous because it separates design questions that are often intertwined in individual papers. By decomposing prototype learning into this taxonomy, we obtain a structured framework for analyzing the evolution of the field and identifying underexplored algorithmic combinations.

\emph{It should be noted that the proposed taxonomy is intended to describe the design space of prototype learning rather than prescribe a fixed design sequence}. While certain design choices naturally co-occur in practice (for instance, discarded inference directly entails no prototype-based explanation at test time, frozen prototypes entail no training objective, and on-the-fly construction is predominantly paired with few-shot episodic training), many choices across these dimensions remain modular: a global prototype can be either single \cite{snell2017prototypical} or multiple \cite{allen2019infinite}, and a discriminative objective can be combined with either on-the-fly \cite{vinyals2016matching, snell2017prototypical} or gradient-based \cite{chen2019this, li2018deep} prototype updates.

\subsubsection{Taxonomic Coding Examples}
\label{sec:worked_coding_examples}
To illustrate how these rules deterministically assign labels across foundational and hybrid architectures (as summarized in Table~\ref{tab:landmark_papers} and Appendix Table~\ref{tab:appendix_rlu_coding}), we walk through five representative examples:
\begin{enumerate}
    \item \textbf{Prototypical Networks (ProtoNet) \cite{snell2017prototypical}:} Coded as \emph{Deterministic -- Global -- Single -- Class-Specific} (R), \emph{Discriminative -- On-the-fly} (L), and \emph{Active -- Distance-based -- Implicit} (U). Each class is represented by one class-specific, deterministic centroid that averages the global feature vector of each image. These prototypes are computed on-the-fly from support embeddings (Eq.~\ref{eq:prototype_data_derived}). During training, the network minimizes episodic cross-entropy (\emph{Discriminative}), and prototypes remain operational at inference (\emph{Active}) via squared Euclidean distance without exposing any visual explanations (\emph{Implicit}).
    
    \item \textbf{ProtoPNet \cite{chen2019this}:} Coded as \emph{Deterministic -- Region -- Multiple -- Class-Specific} (R), \emph{Discriminative -- Gradient-based} (L), and \emph{Active -- Distance-based -- Part-based} (U). The authors explicitly set 10 prototypes per class in their experiments (multiple class-specific prototypes); each prototype learns to mimic a region in the training image (region prototypes), and the main training loop updates persistent prototypes via backpropagation under cross-entropy (\emph{Gradient-based} and \emph{Discriminative}). During inference, prototypes are operational (\emph{Active}) via inverted $L_2$ distance and explain predictions through localized image patches (\emph{Part-based}).
    
    \item \textbf{Stochastic Prototype Embeddings (SPE) \cite{scott2019stochastic}:} Coded as \emph{Probabilistic -- Global -- Single -- Class-Specific} (R), \emph{Discriminative -- On-the-fly} (L), and \emph{Active -- Distance-based -- Implicit} (U). The support embeddings of a class are combined into one Gaussian class prototype (including mean and variance) before being sampled into a specific prototype vector (\emph{Probabilistic}, \emph{Single}, \emph{Class-Specific}); these prototypes are computed from the current episode (\emph{On-the-fly}). Training uses the classification loss only (\emph{Discriminative}), and queries are classified by their distance to the prototypes at inference using Euclidean distance (\emph{Active}, \emph{Distance-based}) without visual explanation (\emph{Implicit}).

    \item \textbf{Prototype-Oriented Contrastive Learning \cite{liu2023prototype}:} Coded as \emph{Deterministic -- Region -- Multiple -- Class-Specific} (R), \emph{Contrastive -- On-the-fly} (L), and \emph{Discarded -- Distance-based -- No prototype-based explanation} (U), with \emph{Shape \& Anatomy} as its clinical abstraction. The method extracts multiple deterministic feature vectors per class from meaningful local sub-regions of the image using spatially aware clustering to preserve structural priors (\emph{Shape \& Anatomy}, \emph{Deterministic}, \emph{Region}, \emph{Multiple}, \emph{Class-Specific}); these prototypes are dynamically generated per mini-batch rather than stored as persistent parameters (\emph{On-the-fly}). Training supervises the prototype set via a voxel-to-prototype contrastive loss to regularize the embedding space (\emph{Contrastive}), and voxel queries are compared to the prototypes during training using cosine similarity (\emph{Distance-based}), but the prototypes are strictly auxiliary and dropped during the standard sliding-window inference (\emph{Discarded}) without providing visual explanations (\emph{No prototype-based explanation}).

    \item \textbf{SimSID \cite{xiang2024exploiting}:} Coded as \emph{Deterministic -- Multi-scale -- Multiple -- Shared Dictionary} (R), \emph{Generative -- Gradient-based} (L), and \emph{Active -- Distance-based -- Implicit} (U), with \emph{Shape \& Anatomy} as its clinical abstraction. The method maintains a space-aware memory matrix acting as a shared dictionary of normal, recurrent visual patterns to perform feature-level in-painting (\emph{Shared Dictionary}). This memory matrix consists of multiple deterministic items distributed hierarchically across different resolution levels of the generator to capture anatomical structures (\emph{Shape \& Anatomy}, \emph{Deterministic}, \emph{Multi-scale}, \emph{Multiple}). The memory items are optimized end-to-end via backpropagation (\emph{Gradient-based}) under an image reconstruction framework designed to recover healthy anatomical features (\emph{Generative}). During inference, these memory items remain actively involved to generate the normal image used for anomaly discrimination (\emph{Active}), mapping extracted features to the closest items in the memory matrix via a distance constraint (\emph{Distance-based}). However, they function internally for feature in-painting rather than exposing explicit, part-based visual explanations to the clinician (\emph{Implicit}).
\end{enumerate}

\subsection{Explainability of Prototypes}
\label{sec:explainability_of_prototypes}

From an architectural perspective, prototype explainability generally operates in one of two main states:
\begin{itemize}
    \item \textbf{The model uses explicit reference representations.} In this state, prototypes act as latent class centroids (global prototypes) or regional features. While they ground the model mathematically, they remain abstract arrays of numbers in the latent space, providing an \emph{implicit} explanation. This is common in few-shot learning designs where prototypes are built dynamically and cannot be directly visualized.
    \item \textbf{Those references can be displayed.} To provide explicit visual evidence, prototypes must be tied to concrete data, allowing them to be displayed as localized patches (\emph{part-based}) or whole training examples (\emph{case-based}).
\end{itemize}

Neither state guarantees that the displayed evidence faithfully explains the prediction \cite{hoffmann2021looks,xudarme2023sanity} or that it is clinically meaningful and useful \cite{pathak2024prototype}. Each requires its own evaluation, such as faithfulness tests that remove the matched region or reader studies with clinicians \cite{nauta2023anecdotal,nauta2023co12}. Our taxonomy codes the grounding format exposed by the architecture (Section~\ref{sec:core_components}), not whether such evaluations were performed.

These two states raise a practical question: \textit{what makes a prototype choice more or less interpretable?} In practice, this depends entirely on how the model is designed across several components:
\begin{itemize}
    \item \textbf{Training pipeline:} If prototypes are simply added to a standard encoder-classifier network, they remain abstract. To be visualized, the pipeline must be designed to either periodically project and save a bank of real patches that are closest to the prototypes (e.g., ProtoPNet and its variants \cite{chen2019this}) or train a decoder simultaneously to generate images from the prototypes (e.g., PrototypeDNN \cite{li2018deep}).
    \item \textbf{Receptive field of the encoder:} To update prototypes, the network computes the distance between image patch features and the prototypes. However, in deep networks, these patch features have large, overlapping receptive fields. Consequently, a visualized patch might secretly rely on contextual pixels outside the displayed bounding box. To ensure the explanation is faithful, the encoder must be adjusted to produce strictly localized features.
    \item \textbf{Number of prototypes:} More is not always better. While too few prototypes limit the diversity of clinical features they can represent, an excessive number creates redundant explanations that overwhelm the user \cite{chen2019this}.
    \item \textbf{Comparison mechanism:} Even if every prototype is identical to a real patch, using a black-box learnable function to compare the distance between prototypes and test image features reduces mathematical transparency compared to using fixed Euclidean or cosine distances.
\end{itemize}

Furthermore, even if a prototype is visually clear and mathematically transparent, a critical challenge remains: \textit{how do we prevent prototypes from learning shortcuts?} Shortcut learning, where prototypes latch onto confounding features like radiographic text markers, image edges, or colored skin artifacts instead of true pathology \cite{gautam2023looks, nauta2023interpreting}, is a primary reason models fail to transfer well across hospitals. Forcing prototypes to represent real clinical anatomy requires explicit interventions:
\begin{itemize}
    \item \textbf{Data preprocessing:} Cropping images to the region of interest rather than feeding the full image helps prevent the model from learning background artifacts, ensuring it receives as little irrelevant noise as possible.
    \item \textbf{Spatial constraints:} Models can be constrained to extract features exclusively from true anatomical regions. In segmentation, prototypes are commonly constructed by utilizing lesion or tumor masks to filter out background features before the prototypes are computed \cite{wang2019panet}. In classification, a loss component can penalize prototypes that learn from healthy areas; for instance, the IAIA-BL approach \cite{barnett2021iaia} requires providing spatial masks during training to explicitly tell the model which areas are healthy and which are not.
    \item \textbf{Pruning shortcut prototypes:} In classification tasks, architectures can explicitly identify prototypes that represent shortcuts and prune them by setting their classification weights to zero, as demonstrated by PIP-Net \cite{nauta2023interpreting}.
\end{itemize}

By understanding these states and engineering constraints, it becomes clear that interpretable prototype learning is not an automatic byproduct of adding reference vectors to a model but a deliberate architectural design.

\subsection{Prototype Learning and Related Paradigms}
\label{sec:related_paradigms}

In the following discussion, we examine the relationships between prototype learning and related paradigms, beginning with metric learning and concluding with dictionary learning.

Metric learning \cite{chopra2005learning, weinberger2009distance} aims to learn an embedding space in which semantically similar samples are close and dissimilar samples are far apart. Prototype learning often relies on such an embedding space, but differs in that it introduces explicit reference structures. A metric learning model may classify samples using pairwise distances without maintaining persistent prototypes, whereas a prototype learning model uses a set of representative entities during training and/or inference.

Contrastive learning \cite{hadsell2006dimensionality, chen2020simple} learns representations by comparing positive and negative pairs. It can be used to initialize or improve prototype learning models, but contrastive representations are not necessarily prototype-based. A contrastive model becomes prototype-based only when explicit prototypes are constructed, updated, or used as anchors for reasoning.

Meta-learning \cite{finn2017model,vinyals2016matching,snell2017prototypical} aims to learn models that can rapidly adapt to new tasks or classes from limited data. Prototype learning is closely related to meta-learning in few-shot settings, where class prototypes are constructed from a support set and used to classify query samples. However, prototype learning does not inherently require a meta-learning framework; prototypes can also be learned or constructed in conventional supervised, self-supervised, or unsupervised settings.

Concept bottleneck models (CBMs) \cite{koh2020concept} route predictions through scores for named, human-defined concepts, such as specific clinical findings, whereas prototype methods route them through comparisons with reference representations. The two designs can overlap: the Locality-aware CBM \cite{jeon2025locality} learns one prototype per concept to localize where each concept appears in the image.

Dictionary learning and vector quantization \cite{olshausen1996emergence, van2017neural} represent data using a finite set of basis vectors or codebook entries. When these basis elements are used as representative patterns for classification, segmentation, reconstruction, or interpretation, they can be viewed as prototypes.

\subsection{Why Prototype Learning Is Particularly Relevant to Medical Imaging}
\label{sec:prototype_medical_relevance}
Prototype learning is particularly relevant to medical imaging because it naturally supports several properties that are important in clinical applications.

First, prototype-based models can provide explanations grounded in representative cases, regions, or disease patterns. This is related to the case-based reasoning commonly used in clinical image interpretation, where a new finding may be compared with previously observed appearances. Prototype-based approaches have adopted this principle explicitly across diverse modalities, linking predictions to representative image patterns and clinically relevant regions in digital mammography \cite{barnett2021iaia}, chest radiography \cite{kim2021xprotonet, singh2021interpretable}, volumetric brain magnetic resonance imaging (MRI) \cite{wei2024mprotonet}, and histological whole-slide images \cite{rymarczyk2022protomil, yang2023tpmil}. Furthermore, to ensure these visual explanations are diagnostically reliable, recent frameworks actively anchor prototypes to verified medical concepts \cite{huy2025interactive} and provide precise relevance maps that allow clinicians to audit the network's reasoning \cite{nauta2023interpreting, gautam2023looks}. Related forms of spatially grounded evidence also arise in medical image segmentation. EML \cite{wang2025edge} constructs an explicit inner-boundary prototype, CoW \cite{jiang2025concentrate} extracts additional hard prototypes from ambiguous support-image regions, and Proto-MSN \cite{yang2026proto} distinguishes boundary from non-boundary representations. These mechanisms make the image regions contributing to prototype matching more spatially identifiable, although they do not necessarily provide case-based or clinician-facing explanations.

Second, medical imaging datasets are often limited and expensive to annotate, particularly for tasks requiring expert pixel- or voxel-level labels. Prototype learning can support data-efficient learning by representing classes or structures through a limited set of representative features. This role is particularly evident in few-shot medical image segmentation. ALPNet \cite{ouyang2020self} learns adaptive local prototypes through superpixel-based self-supervised episodes, reducing its dependence on manually annotated training classes, while PAMI \cite{zhu2023partition} and RPT \cite{zhu2023few} use regional prototypes to make better use of the limited information available from a support set. Prototype-based supervision also extends beyond few-shot learning: CPCL \cite{wu2022all} transfers information between labeled and unlabeled images through forward and backward prototypical consistency. Prototype-based approaches have also been explored for few-shot domain adaptation, where prototypes help transfer knowledge across imaging domains with limited target-domain annotations \cite{wang2024pfmnet}. Furthermore, prototypes also enable networks to operate in completely label-free regimes, circumventing the need for pathological annotations entirely through unsupervised anomaly detection trained exclusively on healthy scans \cite{li2025diverse, wu2025anyad} or zero-shot anomaly localization \cite{wan2026beyond}.

\newlength{\repsubw}
\newlength{\repbodyw}

\begin{table*}[b]
\centering
\scriptsize
\caption{\textbf{Representative prototype learning architectures categorized using the proposed R-L-U framework and clinical abstraction taxonomy}. These methods are selected to illustrate the structural diversity of prototype representation, learning strategy, and utilization across the literature. \textbf{Bold text} highlights the specific taxonomic component that represents the primary structural innovation or distinguishing feature of each method. Deter. = Deterministic; Prob. = Probabilistic; Multi = Multiple Prototypes; CS = Class-Specific; SD = Shared Dictionary; No expl. = No prototype-based explanation.}
\label{tab:landmark_papers}
\renewcommand{\arraystretch}{1.35}
\setlength{\tabcolsep}{2pt}
\settowidth{\repsubw}{(Form -- Granularity -- Cardinality -- Class assoc.)}
\settowidth{\repbodyw}{Deter. -- \textbf{Multi-scale} -- Multi -- CS}
\begin{tabular*}{\textwidth}{@{\extracolsep{\fill}}ll>{\hspace{\dimexpr(\repsubw-\repbodyw)/2\relax}}lll@{}}
\hline
\textbf{Model (Reference)} &
\textbf{Clinical abstraction} &
\multicolumn{1}{c}{\textbf{Representation}} &
\multicolumn{1}{c}{\textbf{Learning Strategy}} &
\multicolumn{1}{c@{}}{\textbf{Utilization}} \\
 & &
\multicolumn{1}{c}{(Form -- Granularity -- Cardinality -- Class assoc.)} &
\multicolumn{1}{c}{(Objective -- Optimization)} &
\multicolumn{1}{c@{}}{(Role -- Comparison -- Interpretability)} \\
\hline
MatchingNet \cite{vinyals2016matching} & Class Identity & Deter. -- Global -- Multi -- CS & Discriminative -- \textbf{On-the-fly} & Active -- Distance-based -- Implicit \\
ProtoNet \cite{snell2017prototypical} & Class Identity & Deter. -- Global -- \textbf{Single} -- CS & Discriminative -- On-the-fly & Active -- Distance-based -- Implicit \\
VQ-VAE \cite{van2017neural} & Appearance & Deter. -- Region -- Multi -- \textbf{SD} & \textbf{Generative} -- Gradient-based & Active -- Distance-based -- Implicit \\
RelationNet \cite{sung2018learning} & Class Identity & Deter. -- Global -- Single -- CS & Discriminative -- On-the-fly & Active -- \textbf{Learnable metric} -- Implicit \\
PL + SEG \cite{dong2018few} & Shape \& Anatomy & Deter. -- \textbf{Region} -- Single -- CS & Discriminative -- On-the-fly & Active -- Learnable metric -- Implicit \\
PrototypeDNN \cite{li2018deep} & Appearance & Deter. -- Global -- Multi -- SD & Discriminative -- Gradient-based & Active -- Distance-based -- \textbf{Case-based} \\
ProtoPNet \cite{chen2019this} & Appearance & Deter. -- Region -- Multi -- CS & Discriminative -- Gradient-based & Active -- Distance-based -- \textbf{Part-based} \\
SPE \cite{scott2019stochastic} & Statistics & \textbf{Prob.} -- Global -- Single -- CS & Discriminative -- On-the-fly & Active -- Distance-based -- Implicit \\
PCL \cite{li2020pcl} & Class Identity & Deter. -- Global -- Multi -- SD & \textbf{Contrastive} -- Clustering-based & Discarded -- Distance-based -- No expl. \\
ProtoTree \cite{nauta2021neural} & Appearance & Deter. -- Region -- Multi -- \textbf{SD} & Discriminative -- Gradient-based & Active -- Distance-based -- Part-based \\
ContrastiveSeg \cite{wang2021exploring} & Class Identity & Deter. -- \textbf{Multi-scale} -- Multi -- CS & Contrastive -- On-the-fly & Discarded -- Distance-based -- No expl. \\
\hline
\end{tabular*}
\end{table*}

Third, medical images exhibit substantial variability across scanners, acquisition protocols, institutions, and patient populations. Consequently, models deployed at new hospitals frequently fail due to a combination of cross-center domain shifts and hospital-specific shortcut learning \cite{geirhos2020shortcut,bassi2024improving}. In cross-institution pelvic MRI segmentation, Li et al. \cite{li2023prototypical} combine local prototype matching with spatial registration to align support and query anatomy. PFMNet \cite{wang2024pfmnet} addresses few-shot domain adaptation by mapping target-domain class prototypes toward learned source-domain exemplar vectors, whereas Yu et al. \cite{yu2023source} align target features to class prototypes inherited from the source model when neither source images nor target labels are available. Furthermore, in decentralized clinical workflows where privacy constraints prohibit data sharing, prototypes serve as lightweight communication anchors, enabling federated networks to align multi-center representations and mitigate institutional concept shifts without exposing raw patient records \cite{zhao2025medfedproto,wang2025fedpmr}. 

These characteristics make prototype learning a promising framework for medical image analysis, particularly in settings involving limited supervision, heterogeneous data, and the need for interpretable predictions. At the same time, prototype-based models introduce their own challenges, including prototype quality, redundancy, stability, scalability, and clinical validity. These issues are examined in later sections of this review.

\subsection{Summary}
\label{sec:background_summary}

This section established a unified view of prototype learning as a representation learning paradigm centered on explicit reference structures. We defined a prototype as an explicit computational representation that summarizes characteristic information associated with a subset of data and can support learning or inference. We then described a general prototype learning pipeline and formulated it mathematically as the encoding of an input, its interaction with a prototype set, and the generation of a task-specific prediction. Building on this formulation, we introduced the R-L-U design space, which organizes prototype learning methods according to three complementary dimensions: prototype representation, learning strategy, and utilization.

We further positioned prototype learning relative to neighboring paradigms, including metric learning, contrastive learning, meta-learning, concept bottleneck models, and dictionary learning, and discussed why prototype-based approaches are particularly relevant to medical imaging, especially in settings involving limited supervision, heterogeneous data, and the need for interpretable predictions.

The R-L-U framework provides the conceptual foundation for the remainder of this survey. In the following sections, we use these dimensions to trace the evolution of prototype learning methods and comprehensively categorize, compare, and analyze their designs and applications in medical image analysis.

\section{Evolution of Prototype Learning Methodologies}
\label{sec:evolution_of_prototype_learning}

To ensure mathematical consistency across the diverse methodologies analyzed in this review, we maintain the unified notation established in Section~\ref{sec:unified_formulation}. Let $x$ denote an input medical image, $f_\theta(\cdot)$ represent a deep feature extractor yielding latent representations $\mathbf{z}$, and $\mathcal{P} = \{\mathbf{p}_1, \dots, \mathbf{p}_K\}$ represent the set of prototypes. 

Following the R-L-U taxonomy established in Section~\ref{sec:background}, this section comprehensively deconstructs the structural formulation of Prototype Representation, Learning Strategy, and Utilization across both general computer vision and medical imaging. To provide readers with a concrete starting point, \textbf{Table~\ref{tab:landmark_papers}} summarizes a representative selection of methodologies, specifically chosen to illustrate the architectural diversity and evolution of each taxonomic component.

\subsection{Innovation in Prototype Representation}
\label{sec:innovation_prototype_representation}

The representation of prototypes has evolved along four complementary axes: \textbf{representation form}, \textbf{spatial granularity}, \textbf{cardinality}, and \textbf{class association}. These axes describe, respectively, how a prototype is represented mathematically, what spatial extent it represents, how many prototypes are maintained, and whether prototypes are assigned to specific classes or shared across categories.

Across the literature, these design choices have evolved from simple and straightforward representations, such as global and single prototypes, toward more flexible and expressive structures, such as probabilistic, localized, multi-prototype, and shared representations.

\begin{figure*}[b]
    \centering
    \includegraphics[width=\textwidth]{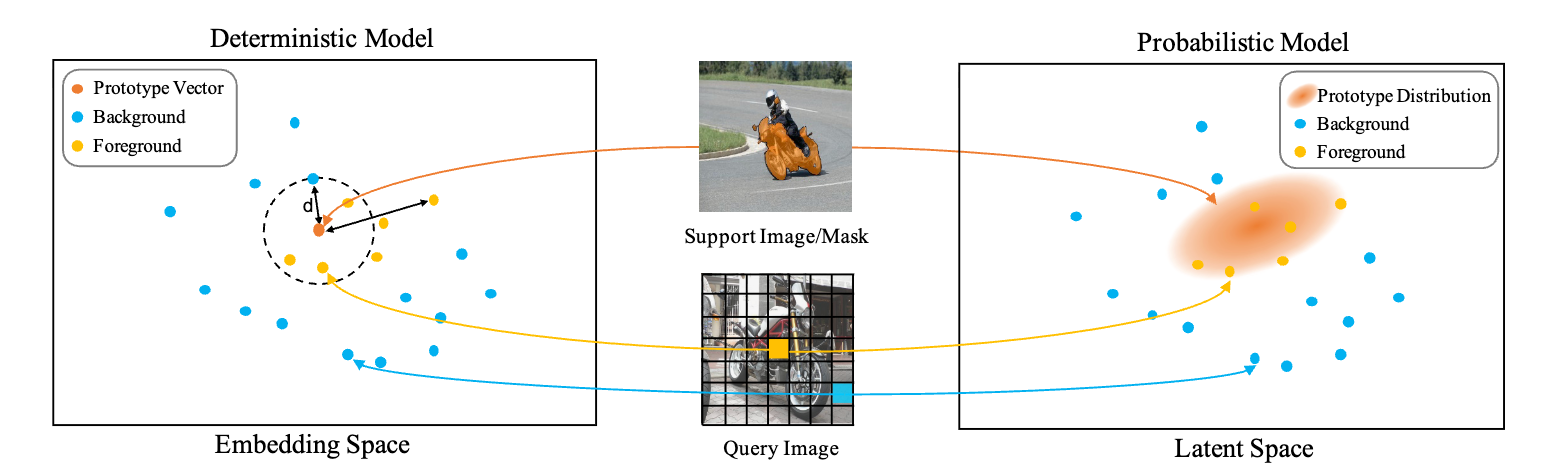}
    \caption{
    \textbf{Comparison of deterministic and probabilistic prototype representations in the latent space.} 
    \textbf{Left (Deterministic):} The class representation is parameterized as a single, fixed-point vector. While memory-efficient, a point estimate cannot adequately capture high intra-class variance or diagnostic ambiguity. 
    \textbf{Right (Probabilistic):} The prototype is formulated as a continuous probability distribution (e.g., a multivariate Gaussian). This distributional representation captures both the expected semantic centroid (mean) and the distributional variance arising from heterogeneous pathological appearances and imaging artifacts (image from \cite{wang2021variational}).
    }
    \label{fig:03_deterministic_vs_probabilistic_prototypes}
\end{figure*}
\subsubsection{Advancements in Representation Form}

The mathematical form of a prototype defines how clinical knowledge is physically encoded in the latent space. As illustrated in Fig.~\ref{fig:03_deterministic_vs_probabilistic_prototypes}, this design space is fundamentally divided into point-based deterministic representations and distributional probabilistic representations. While a single prototype vector provides a compact summary of support features, transitioning to a probabilistic distribution offers a richer representation capable of capturing both the central tendency and the spread of the features associated with a clinical class; whether this spread yields calibrated uncertainty must be evaluated separately. 

\textbf{Deterministic.}
In the deterministic paradigm, a prototype is usually formulated as a fixed-point vector $\mathbf{p}_k \in \mathbb{R}^{d}$ within the latent space. In those foundational architectures, such as Matching Networks \cite{vinyals2016matching}, Prototypical Networks \cite{snell2017prototypical}, and the original ProtoPNet \cite{chen2019this}, prototypes are all instantiated as deterministic representations.
Following these pioneering works, a vast body of literature across both general computer vision and medical image analysis has adopted this deterministic formulation \cite{nauta2021neural,wang2019panet,donnelly2021deformable,barnett2021iaia,sung2018learning,xu2026rethinking,huy2025interactive,liu2026openness,yu2021location}. During inference, the encoded image representation is compared with a prototype represented as a fixed point in the embedding space, typically by computing a geometric distance between the two vectors. How these prototypes are subsequently utilized during inference is discussed further in Section~\ref{sec:prototype_utilization}.

The deterministic formulation offers several practical advantages. Representing each prototype as a single vector is memory-efficient and computationally simple, and it integrates naturally with standard gradient-based optimization frameworks. These properties have contributed to its broad adoption across prototype-based architectures. However, the point-estimate formulation also imposes limitations when the underlying data exhibit substantial intra-class variability, ambiguity, or annotation uncertainty. This issue is particularly relevant in medical imaging, where pathological appearance may vary considerably across patients and acquisition conditions, while image quality, artifacts, and inter-observer disagreement can further complicate the characterization of disease patterns. A single prototype vector may therefore provide an incomplete representation of the distributional variability associated with a disease class.

\textbf{Probabilistic.}
To explicitly capture intra-class variance and clinical uncertainty, probabilistic frameworks formulate prototypes as continuous probability distributions over the latent space, most commonly parameterized as multivariate Gaussians:
\begin{equation}
    \mathbf{p}_k \sim \mathcal{N}(\boldsymbol{\mu}_k, \boldsymbol{\Sigma}_k),
    \label{eq:probabilistic_prototype}
\end{equation}
where the mean vector $\boldsymbol{\mu}_k$ represents the central tendency of the prototype in the latent space, while the covariance matrix $\boldsymbol{\Sigma}_k$ characterizes the variability and uncertainty surrounding that prototype. Rather than relying on a deterministic point estimate, an emerging line of research has shifted toward probabilistic representations \cite{scott2019stochastic, sun2021amortized, zhen2020learning, wang2021variational, kim2019variational}. Stochastic Prototype Embeddings (SPE) \cite{scott2019stochastic} extend Prototypical Networks \cite{snell2017prototypical} by representing individual image embeddings as Gaussian distributions and analytically combining them through precision-weighted aggregation to obtain a probabilistic class prototype. Similarly, Amortized Bayesian Prototype Meta-learning \cite{sun2021amortized} models the class prototype as a latent random variable and infers its distribution from the distributional representations of the support samples through a Bayesian formulation. In contrast, Zhen et al. \cite{zhen2020learning} first aggregate the support-set feature representations and then introduce an additional network that maps the aggregated representation to the mean and variance of the prototype distribution. Wang et al. \cite{wang2021variational} further formulate the prototype itself as a latent random variable and employ variational inference with an amortized inference network to infer its distribution from the support set. 

This capability is particularly valuable in medical imaging, where substantial intra-class variability may arise from heterogeneous pathological appearances across patients and acquisition conditions, while ambiguous anatomical boundaries and noisy observations introduce additional uncertainty. Accordingly, probabilistic prototypes have also been extended to medical imaging for different kinds of tasks \cite{yuan2024effective,raza2025ps3}. 

Despite these advantages, probabilistic representations introduce additional computational and optimization complexity, as modeling distributional parameters such as covariance matrices can increase memory requirements and training difficulty, particularly for high-dimensional medical data.

\subsubsection{Advancements in Spatial Granularity}

\begin{figure}[!t]
    \centering
    \includegraphics[width=1\linewidth]{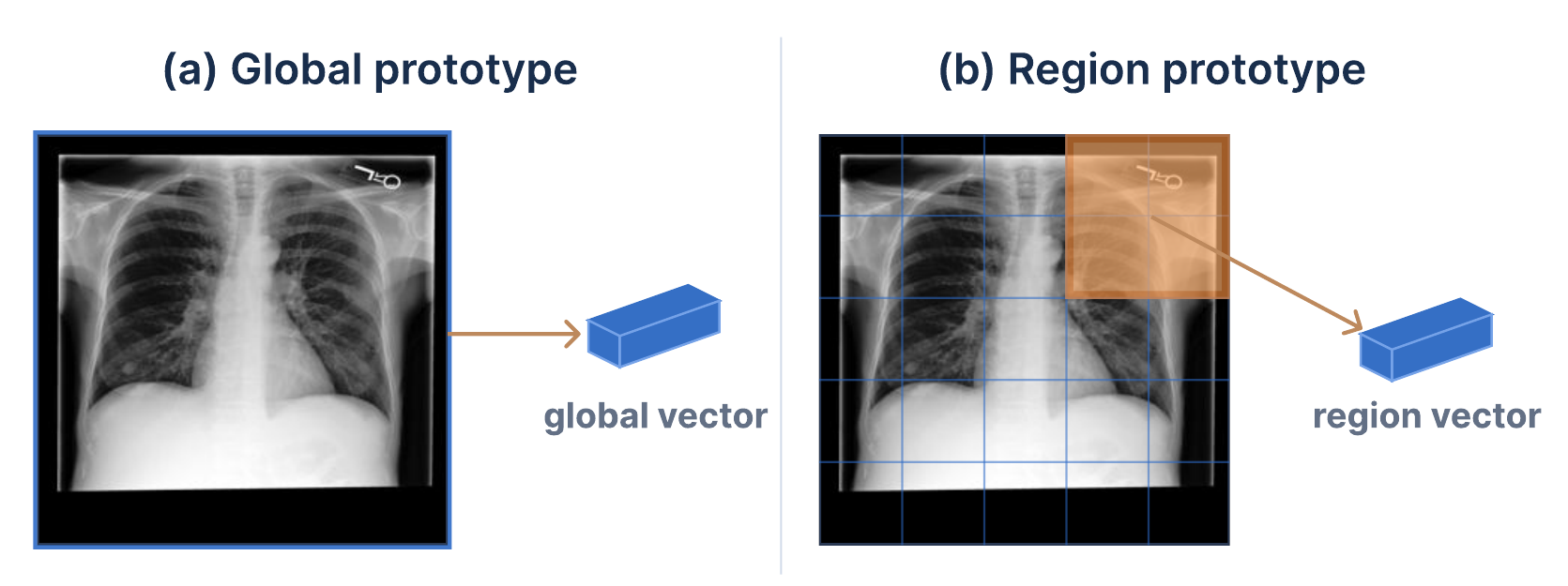}
    \caption{
    \textbf{Illustration of prototype spatial granularity in medical imaging.} 
    The spatial extent encoded by a prototype determines its localization capacity and clinical utility. 
    \textbf{(a)} The entire anatomical image is compressed into a single holistic vector (e.g., via global average pooling), which captures coarse image-level semantics but lacks fine-grained spatial interpretability. 
    \textbf{(b)} Prototypes restrict their receptive fields to specific sub-regions or patches, isolating localized diagnostic findings (e.g., lesions or tumors) from healthy background tissue to enable precise, part-based visual explanations.
    }
    \label{fig:visualize_prototype_in_2d_space}
\end{figure}

Spatial granularity dictates the spatial extent of the anatomical or semantic evidence encoded by a prototype. In general, prototype designs span three complementary spatial scopes: \textbf{global}, \textbf{region}, and \textbf{multi-scale}. As illustrated in Fig.~\ref{fig:visualize_prototype_in_2d_space}, the primary architectural distinction lies in whether a prototype compresses a holistic image or isolates a localized anatomical patch. More recently, frameworks have extended this spectrum toward multi-scale (hierarchical) representations.

\textbf{Global.}
In global formulations, each prototype represents an entire image or patient volume. Early prototype-based and metric-learning frameworks predominantly employed these global representations, where each prototype or reference representation captures image-level rather than spatially localized information \cite{vinyals2016matching, snell2017prototypical, sung2018learning, li2018deep}. Although these methods differ in how support representations are aggregated and utilized during inference, they share the same spatial-granularity property: their prototypes or reference representations encode \textbf{image-level rather than spatially localized information}.

In medical imaging, global prototypes have mainly been adopted in image-level classification settings where computational simplicity, few-shot support averaging, or lightweight federated communication is prioritized \cite{zhu2024sfpl, yu2025hierarchical,zhao2025medfedproto,liu2026openness}. However, compressing an entire scan into a single holistic vector makes it difficult to verify which visual features are actually encoded.

\textbf{Region.} As visually contrasted in Fig.~\ref{fig:visualize_prototype_in_2d_space}, because almost all diagnostic targets in medical imaging are confined to specific lesions or anatomical structures (even when a pathology spans an entire organ), global prototypes inevitably mix target features with irrelevant background regions. Consequently, subsequent architectures shifted toward region-level prototypes that restrict representation to localized image parts. Inspired by case-based reasoning \cite{li2018deep}, ProtoPNet \cite{chen2019this} explicitly learns prototypical image patches corresponding to characteristic visual parts. This localized formulation has been further extended to accommodate deformable spatial regions \cite{donnelly2021deformable} and hierarchical decision structures \cite{nauta2021neural}. In clinical practice, this allows networks to isolate localized diagnostic evidence without entangling predictions with healthy background tissue, enabling precise, part-based explanations for chest radiography \cite{kim2021xprotonet, singh2021interpretable}, digital mammography \cite{barnett2021iaia}, and neuroimaging \cite{wei2024mprotonet}. 

Concurrently, few-shot semantic segmentation drove the extension of prototypes from holistic image-level summaries to \textbf{spatially localized foreground regions}. Early frameworks extracted representative guidance features via masked average pooling over annotated support regions, using these prototypes to predict pixel-level query masks \cite{dong2018few, zhang2020sg, wang2019panet}. Subsequent methods further decomposed these foreground representations into multiple localized or part-aware prototypes to capture fine-grained intra-class structural variation \cite{liu2020part, yang2020prototype, yu2021location, zhu2023few}. By constraining the prototype's receptive field to pathological sub-regions, these region-level prototype architectures naturally support localized medical image segmentation by matching spatially restricted support representations with query regions \cite{ouyang2020self,zhu2023partition}. Importantly, spatial granularity refers to the spatial extent \emph{represented} by each prototype, not the downstream prediction resolution; a region-derived prototype is naturally equipped to produce dense, pixel-level segmentation masks.

Although some segmentation methods describe their representations as \textbf{pixel prototypes} \cite{zhou2022rethinking,du2022weakly}, we do not treat pixel-level as a separate spatial granularity in our taxonomy. In these methods, prototypes are derived from individual spatial locations in a latent feature map, whose receptive fields already cover local image regions rather than isolated raw pixels. We therefore categorize such representations as \textbf{region-level} prototypes rather than introducing a separate pixel-level category. This distinction ensures that spatial granularity reflects the actual spatial extent represented by each prototype.

\textbf{Multi-scale.} More recent studies have combined prototypes from different spatial resolutions, leading to multi-scale prototype formulations that capture both coarse and fine-grained features \cite{wang2021exploring, porta2025multi, gowda2025prototype}. In this setting, multi-scale prototypes do not represent a single spatial extent but combine information from multiple spatial scales.

\subsubsection{Advancements in Cardinality}
\label{sec:cardinality}

Here, cardinality refers to \textbf{how many prototypes are maintained} ($K=1$ vs. $K>1$). This axis is distinct from both spatial granularity (\emph{what area each prototype represents}) and class association (\emph{whether prototypes are class-specific or shared}). For example, a method can use either a \emph{single global prototype} to represent an entire image-level class or \emph{multiple global prototypes} to capture different modes within that same image-level class.

\textbf{Single prototype.} In early prototype architectures, models predominantly used a single prototype per target class due to its simplicity and computational efficiency. In few-shot classification, this typically manifests as a single global centroid summarizing the support images of each category \cite{snell2017prototypical,scott2019stochastic,sung2018learning, mettes2019hyperspherical, jetley2015prototypical}. Similarly, in early semantic segmentation frameworks, a single regional prototype is commonly extracted to represent the entirety of each foreground or background class \cite{dong2018few,wang2019panet,zhang2020sg}. While a single prototype provides a highly compact representation, it becomes a representational bottleneck when a class exhibits substantial intra-class variation, heterogeneous morphologies, or multi-modal data distributions.

\textbf{Multiple prototypes.} To overcome the limitations of a single centroid, prototype learning gradually shifted toward maintaining multiple prototypes. In few-shot classification, models introduced multiple global prototypes per class to effectively capture multi-modal class distributions and high intra-class variance \cite{allen2019infinite}. This paradigm shift occurred regardless of the spatial area represented by each prototype; for instance, in semantic segmentation, subsequent methods decomposed foreground representations into multiple localized prototypes to capture distinct visual patterns and fine-grained structural variations within the same class \cite{liu2020part,yang2020prototype,yu2021location,zhu2023few}.

Multiple prototypes are also widely used in interpretable prototype-based classification, where several prototypes represent different characteristic patterns of a class, either at the global level \cite{li2018deep} or at the regional level \cite{chen2019this,donnelly2021deformable}, as well as in unsupervised clustering and codebook frameworks \cite{caron2020unsupervised,li2020pcl,van2017neural}.

\subsubsection{Advancements in Class Association}
\label{sec:class_association}
Complementing cardinality, class association defines \textbf{how prototypes are linked to target categories}, distinguishing between \textbf{class-specific} prototype sets and \textbf{shared dictionary} pools.

\textbf{Class-specific.} In the majority of classification and segmentation frameworks, each prototype (whether single or multiple) is exclusively assigned to a single target class or anatomical structure \cite{snell2017prototypical,chen2019this,wang2019panet,donnelly2021deformable}. During training and inference, class-specific prototypes only accumulate evidence for their designated category, ensuring a direct one-to-one or many-to-one mapping between prototypes and diagnostic labels.

\textbf{Shared dictionary.} A further development moves from class-specific prototype sets toward shared prototype dictionaries, in which multiple prototypes ($K>1$) are maintained in a class-agnostic pool rather than exclusively assigned to individual classes. ProtoTree \cite{nauta2021neural} places prototypes at decision-tree branches rather than assigning them to individual classes, ProtoPool \cite{rymarczyk2022interpretable} explicitly learns a pool of reusable prototypes shared across categories, and VQ-VAE \cite{van2017neural} formulates its shared prototypes as a discrete \emph{embedding dictionary}. In clinical applications, this shared-pool formulation is useful for complex tasks, enabling normal tissue memory banks for unsupervised anomaly reconstruction \cite{xiang2023squid,xiang2024exploiting}, morphological prototype vocabularies for whole-slide images \cite{song2024morphological,raza2025ps3}, and unified cross-modal memory banks of canonical clinical findings for radiology report generation \cite{chen2021cross} and vision-language pre-training \cite{cheng2023prior}. A tree alone does not make prototypes shared, however: unlike ProtoTree, the prototypical decision tree of Yu et al.~\cite{yu2025hierarchical} anchors each node to one class of the diagnostic taxonomy, so its prototypes remain class-specific. 

\begin{figure}
    \centering
    \includegraphics[width=\columnwidth]{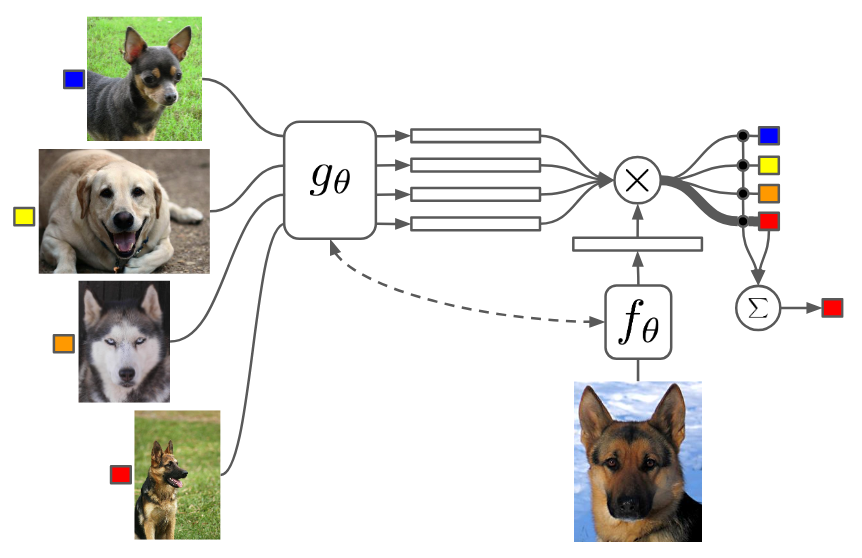}
    \caption{\textbf{Prototype-based inference in few-shot classification.}
    Although Vinyals et al. \cite{vinyals2016matching} do not explicitly use the term ``prototype,'' the support-sample representations serve as reference representations for classifying the query, reflecting the core prototype-based principle later formalized by Prototypical Networks \cite{snell2017prototypical} (image from \cite{vinyals2016matching}).
}
    \label{fig:03_prototype_use_during_inference}
\end{figure}
\subsection{Innovation in Prototype Learning Strategies}
\label{sec:innovation_in_prototype_learning_strategies}

Having defined the representation of prototypes, the next question is \textbf{how they are learned}. In the proposed taxonomy, prototype learning strategy considers two complementary aspects: the \textbf{training objective}, which describes the main loss used to shape the prototype representation, and the \textbf{optimization mechanism}, which describes how prototype values are obtained or updated during learning. These axes are independent design choices and can be combined in different ways.

\subsubsection{Shift in Training Objectives}
\label{sec:primary_learning_paradigm}

The training objective determines the main loss used to shape the relationship between inputs and prototypes, including \textbf{discriminative}, \textbf{contrastive}, and \textbf{generative} objectives. Frozen prototypes are coded as \emph{No objective}, since there is no loss that updates the prototypes during training at all.

\textbf{Discriminative.} Early prototype-based methods mainly adopt a discriminative paradigm, using prototypes to support the final task prediction, such as classification or segmentation \cite{vinyals2016matching,snell2017prototypical,chen2019this,li2018deep,dong2018few}. This direct use of reference representations for prediction is illustrated in Fig.~\ref{fig:03_prototype_use_during_inference}, where Vinyals et al. \cite{vinyals2016matching} compare the embedding of a query image with the embeddings of the support examples and use the resulting similarities for classification. Within the general formulation in Eq.~\ref{eq:unified_loss}, these methods are primarily driven by the task loss $\mathcal{L}_{\mathrm{task}}$, such as cross-entropy for classification, while prototype-specific and regularization terms may be added when needed.

\begin{figure}[b]
    \centering
   \includegraphics[width=\columnwidth]{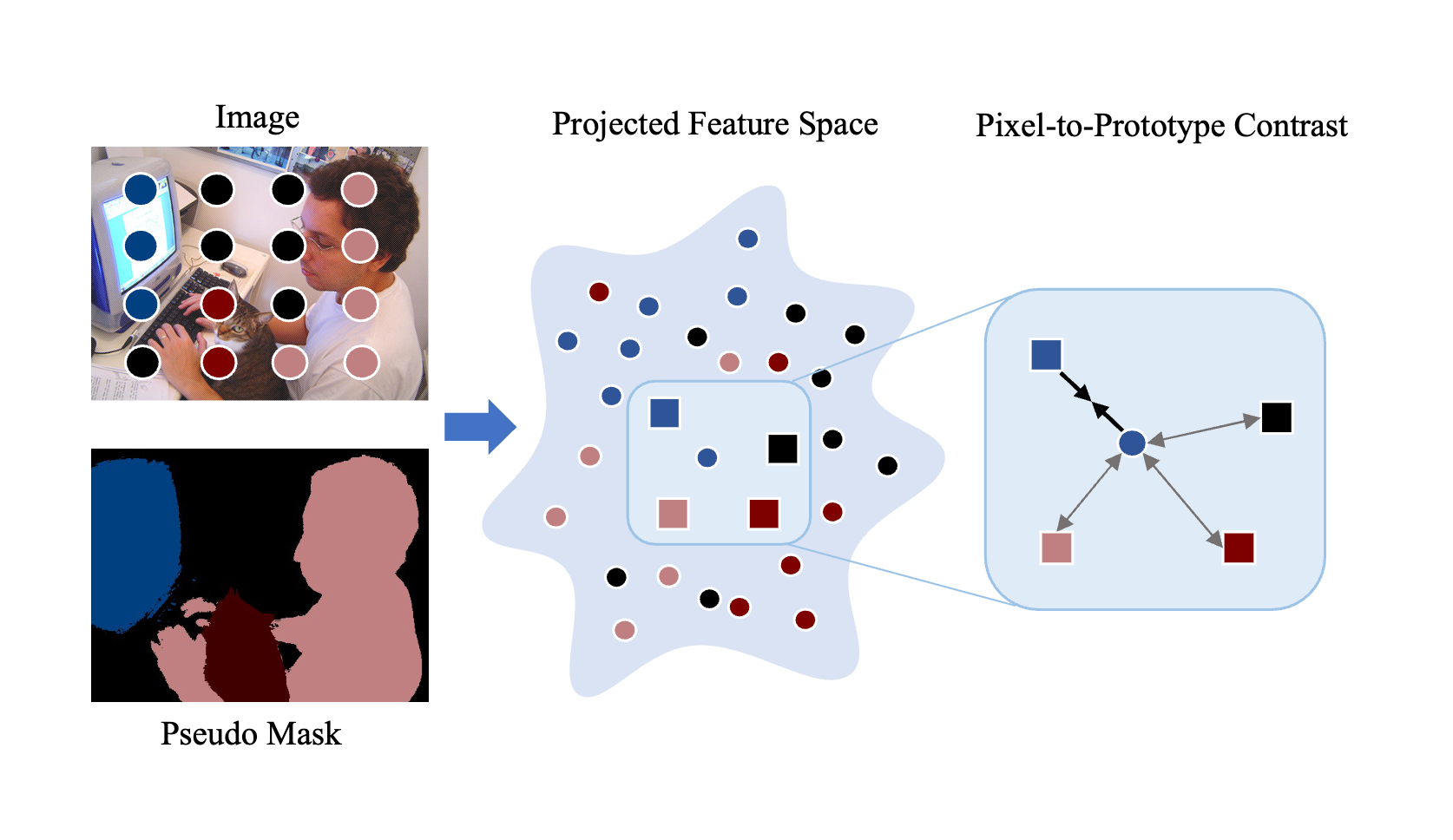}
    \caption{{\textbf{Prototype-based contrastive learning.}}
    Features are encouraged to align with prototypes of the same category while being separated from prototypes of other categories (image from \cite{du2022weakly}).}
    \label{fig:prototype_contrastive}
\end{figure}

\textbf{Contrastive.} To further structure the learned representation, later approaches began to combine prototype learning with contrastive objectives. In these methods, prototypes serve as reference representations for aligning samples with their corresponding classes while separating them from others. For example, Du et al. \cite{du2022weakly} explicitly perform pixel-to-prototype contrast, as illustrated in Fig.~\ref{fig:prototype_contrastive}. Rather than using the prototypes directly for inference, they use them during training to encourage features of different categories to become more separated in the representation space. In the formulation of Eq.~\ref{eq:unified_loss}, the prototype-related objective can instead be instantiated as a contrastive loss:
\begin{equation}
    \mathcal{L}
    =
    \mathcal{L}_{\mathrm{task}}
    +
    \lambda_{\mathrm{proto}}\mathcal{L}_{\mathrm{contrastive}}.
\end{equation}
Depending on the model design, these prototypes may remain part of the downstream prediction process or serve primarily as training references for representation learning \cite{li2020pcl,caron2020unsupervised}.

\textbf{Generative.} A different direction incorporates prototypes into \textbf{generative} objectives, where the learned representation is also required to preserve or model information from the input data. VQ-VAE \cite{van2017neural}, for example, uses an encoder that maps the input image into a grid of local feature vectors, then replaces each feature vector with its nearest prototype from a shared dictionary (codebook), and passes only these selected prototypes to a decoder to reconstruct the original image. More prototype-oriented approaches further combine prototype representations with variational or reconstruction objectives \cite{kim2019variational,gautam2022protovae}. In these settings, prototypes are shaped not only by class separation but also by the need to capture representative characteristics of the underlying data. In the medical domain, this paradigm naturally extends beyond spatial reconstruction to cross-modal generation and pre-training, where shared prototype matrices reconstruct sentence-level clinical findings \cite{cheng2023prior} or directly condition autoregressive decoders to generate clinical radiology reports \cite{chen2021cross, wang2022cross}.

Overall, prototype learning objectives have expanded from direct task supervision to objectives that more explicitly structure the representation space or model the underlying data distribution.

The optimization mechanism determines \emph{how prototype values are obtained or updated during learning}. Existing methods use several mechanisms, including \textbf{on-the-fly}, \textbf{gradient-based}, \textbf{momentum-based}, \textbf{frozen}, and \textbf{clustering-based} approaches.

\textbf{On-the-fly.} Early few-shot methods commonly construct prototypes on-the-fly from the available support examples rather than maintaining persistent prototype parameters \cite{snell2017prototypical, vinyals2016matching, scott2019stochastic, dong2018few, wang2019panet}. Prototypical Networks \cite{snell2017prototypical} provide the canonical example, where the class prototype is recomputed from the support embeddings for each episode. This allows immediate adaptation to new classes, but the resulting prototype depends directly on the available support examples. This corresponds to the data-derived prototype formulation introduced in Eq.~\ref{eq:prototype_data_derived}. Similar support-derived prototype mechanisms are also used in few-shot semantic segmentation \cite{dong2018few,wang2019panet}. This strategy avoids learning separate persistent prototypes for novel classes.

\textbf{Gradient-based.} To obtain persistent and task-specific prototypes, subsequent approaches commonly treat prototypes as learnable parameters and update them through gradient-based optimization \cite{van2017neural,li2018deep,chen2019this, arik2019protoattend}. This allows the prototypes and feature encoder to adapt jointly throughout training and provides greater flexibility than constructing prototypes only from the current support set. This mechanism has also been widely adopted in later contrastive approaches, where gradients from prototype-based contrastive objectives are used to jointly refine the prototypes and the feature encoder \cite{caron2020unsupervised, du2022weakly}.

\textbf{Momentum-based.} Beyond direct gradient updates, some methods use momentum-based mechanisms to obtain smoother prototype estimates from evolving feature representations. A typical exponential moving average (EMA) update is
\subsubsection{Advancements in Optimization Mechanisms}
\label{sec:optimization_mechanisms}
\begin{equation}
    \mathbf{p}_k^{(t)} = m \, \mathbf{p}_k^{(t-1)} + (1 - m) \, \bar{\mathbf{z}}_k^{(t)},
    \label{eq:momentum_update}
\end{equation}
where $m \in [0, 1)$ is the momentum coefficient controlling the contribution of the previous prototype state, and $\bar{\mathbf{z}}_k^{(t)}$ denotes the mean (or aggregated) feature representation of samples assigned to class or cluster $k$ in the current mini-batch at iteration $t$. Momentum-prototype works such as Li et al.~\cite{li2021mopro} and Fan et al.~\cite{fan2022few} use this type of update to reduce the influence of individual training samples and maintain more stable prototype estimates. In medical imaging, momentum updates are used, for example, in federated classification \cite{zhao2025medfedproto} and lesion detection \cite{bhat2025exemplar}.

\textbf{Frozen.} Some approaches prescribe the prototype geometry before representation learning and keep these prototypes fixed during model training. Hyperspherical Prototype Networks \cite{mettes2019hyperspherical} position class prototypes through data-independent optimization to achieve large-margin separation, whereas Neural Collapse Terminus \cite{yang2023neural} preassigns class anchors according to a simplex equiangular tight-frame structure. Training then optimizes the feature encoder to align samples with their corresponding fixed prototypes. This design avoids joint prototype updating and provides a stable embedding geometry, although the predefined prototypes may be less flexible in reflecting data-dependent class structure and intra-class variation. In medical imaging, frozen prototypes are rare: Yu et al. \cite{yu2023source} keep the source classifier weights fixed as prototypes during domain adaptation, ProtoSurv \cite{liu2026protosurv} computes class prototypes once in the space of a frozen computed tomography (CT) foundation model, PANTHER \cite{song2024morphological} clusters whole-slide image patches once with $k$-means and keeps the resulting prototypes fixed, and two image synthesis methods guide diffusion models with fixed prototypes \cite{shandilya2024improve,redekop2025prototype}.

\textbf{Clustering-based.} A related clustering-based mechanism forms or updates prototypes by grouping similar feature representations and using the resulting cluster centers as prototypes. PCL \cite{li2020pcl} updates prototypes through an expectation-maximization procedure. These methods allow prototypes to adapt to the structure of the learned feature space rather than being directly learned as independent parameters. In medical imaging, clustering is used to build prototypes from whole-slide image patches \cite{pan2023human}.

Overall, prototype values can be computed from the current data, learned as persistent parameters, kept fixed, or updated through momentum or clustering procedures. These mechanisms are complementary to the training objective and can therefore be combined in different ways.
\subsection{Innovation in Prototype Utilization}
\label{sec:prototype_utilization}

Once prototypes have been constructed and learned, they can be incorporated into the prediction process in different ways. In our general organizing formulation (Eq.~\ref{eq:unified_function}), prototype utilization is primarily determined by the comparison function $S$ and the prediction head $g_\phi$.

\subsubsection{Inference-Time Role}
\label{sec:inference_time_role}

\textbf{Active.}
In the classical prototype paradigm, learned or support-derived prototypes remain active components of the test-time decision pathway (Eq.~\ref{eq:unified_function}). Every unseen test scan is encoded and explicitly compared against or conditioned on the prototype set $\mathcal{P}$ to produce the final prediction \cite{snell2017prototypical,chen2019this,wang2019panet,chen2021cross}.

\textbf{Discarded.}
Not all prototype learning methods retain prototypes as explicit components of the inference process. In a growing class of representation-learning approaches, prototypes are introduced solely during training to regularize the feature encoder, for example, as cluster centers, contrastive targets, or auxiliary semantic anchors, and are removed at test time so that predictions rely directly on the trained encoder and task head (Eq.~\ref{eq:prototype_free_prediction}). In general computer vision, PCL \cite{li2020pcl} and ContrastiveSeg \cite{wang2021exploring} use prototypes to structure representations through prototype-based contrastive objectives, while SwAV \cite{caron2020unsupervised} uses prototypes to enforce cluster consistency across views. In medical imaging, this train-only regularizer role has been widely adopted in semi-supervised segmentation \cite{liu2023prototype,liu2026kp2l}, weakly supervised segmentation \cite{liu2025procns}, vision-language adaptation \cite{zhang2024disease}, and deformable image registration \cite{xu2025medical}.

\subsubsection{Advancements in Comparison Mechanisms}
\label{sec:prototype_inference}

Whether prototypes are evaluated during inference (Eq.~\ref{eq:unified_function}) or solely within the training stage, the comparison function $S$ determines how sample representations interact with prototypes.

\textbf{Distance-based.}
A common strategy is to compare a query representation $\mathbf{z}$ with a prototype $\mathbf{p}_k$ using a predefined distance or similarity function. Prototypical Networks \cite{snell2017prototypical}, for example, represent each class by the mean embedding of its support examples and perform classification according to the Euclidean distance between the query and each class prototype. The corresponding similarity can be written as
\begin{equation}
    S(\mathbf{z},\mathbf{p}_k)
    =
    -\|\mathbf{z}-\mathbf{p}_k\|_2^2.
    \label{eq:euclidean_similarity}
\end{equation}

Here, the negative distance is expressed as a similarity score, with larger values indicating stronger correspondence. Euclidean distance has also been adopted in a broad range of prototype-based classification methods \cite{mensink2013distance,chen2019this}. Another common choice is cosine similarity, particularly when representations are normalized or explicitly constrained to a hyperspherical space \cite{vinyals2016matching,mettes2019hyperspherical,donnelly2021deformable,zhang2020sg,wang2019panet}:
\begin{equation}
    S(\mathbf{z},\mathbf{p}_k)
    =
    \frac{\mathbf{z}^{\top}\mathbf{p}_k}
    {\|\mathbf{z}\|_2\|\mathbf{p}_k\|_2}.
    \label{eq:cosine_similarity}
\end{equation}
The defining property of this category is that the comparison geometry is prescribed rather than learned specifically for the target task. Such mechanisms are attractive for their simplicity, low computational overhead, and relatively direct interpretation: the prediction can be related explicitly to the distance or similarity between an observation and a prototype. However, a fixed metric also imposes a particular geometry on the representation space and may be insufficient when class-specific relationships are more complex than the selected metric can express \cite{sung2018learning,cao2022learning,li2023renap}.

\textbf{Learnable metric.} A natural extension is to replace the predefined comparison function with a trainable compatibility function. Rather than assuming a fixed geometric relationship between a query and a prototype, these methods learn how the two should be compared:
\begin{equation}
    S(\mathbf{z}, \mathbf{p}_k) = h_\psi(\mathbf{z}, \mathbf{p}_k),
    \label{eq:learnable_metric}
\end{equation}
where $h_\psi$ denotes a trainable matching network parameterized by $\psi$. Relation Networks~\cite{sung2018learning} provide an early representative example of this formulation, in which a learned relation module determines the compatibility between representations. Learnable metrics have also been used in other prototype-based methods \cite{dong2018few,li2023renap,cao2022learning}.

Compared with fixed metrics, learnable comparison functions can adapt the notion of similarity to the target task and therefore capture relationships that may not be well described by Euclidean or cosine geometry. This flexibility, however, comes with additional model parameters and computational cost. Moreover, because the comparison function itself is learned, the resulting score is generally less transparent than a predefined distance or similarity measure. Thus, this design represents a transition from prototypes as fixed reference points toward prototypes as components of a learned decision function.

Overall, this progression reflects a shift from prescribing the geometry of prototype-query comparison to learning the comparison function itself. Distance-based methods rely on a fixed similarity rule, whereas learnable-metric methods allow the model to determine prototype-query compatibility from the data.

\subsubsection{Interpretability of Prototypes}
\label{sec:prototype_interpretability}

Prototype utilization and prototype interpretability represent related but distinct design choices. A model may use prototypes directly in prediction without making their semantic meaning explicit, or it may expose prototype information as part of the explanation. We therefore characterize prototype interpretability along four modes: no prototype-based explanation, implicit, part-based, and case-based.

\textbf{No prototype-based explanation.}
When prototypes are discarded after the training phase (\emph{Discarded} inference-time role), the deployed test-time predictor $g_\phi(f_\theta(x))$ does not reference prototypes during inference \cite{li2020pcl,caron2020unsupervised,zhang2024disease,liu2023prototype,xu2025medical}. Because prototypes are absent from the inference pathway, the model provides no prototype-based explanation for its test-time predictions.

\textbf{Implicit.}
In the implicit setting, prototypes participate in the prediction process but are not explicitly associated with a human-interpretable semantic component. For example, a prototype may represent the centroid of a class in an embedding space, with its contribution determined through a distance or similarity score \cite{snell2017prototypical,mettes2019hyperspherical}. Although the prototype-based decision structure can make the source of a prediction more explicit than a conventional end-to-end classifier, the semantic meaning of an individual prototype is not necessarily observable. Thus, interpretability in this setting primarily comes from exposing prototype-level similarity rather than from assigning a human-interpretable concept to each prototype.

\textbf{Part-based.}
Part-based approaches associate prototypes with localized regions or characteristic components of an input. ProtoPNet \cite{chen2019this} introduced prototypical parts that represent recurring visual patterns and explained predictions by identifying image regions that resemble the learned prototypes while still having comparable accuracy. ProtoPShare \cite{rymarczyk2021protopshare} enables prototypical parts to be shared across classes, while Deformable ProtoPNet \cite{donnelly2021deformable} allows the relative positions of prototypical parts to adapt to the input. ProtoTree \cite{nauta2021neural} similarly uses prototypical parts within a decision-tree structure to provide localized evidence for hierarchical decisions. These approaches make the connection between a prediction and specific visual evidence more explicit, providing explanations in terms of \emph{where} and \emph{what} in the input resembles the learned prototype.

\textbf{Case-based.}
Case-based approaches ground prototype information in representative instances or examples from the training data. Instead of interpreting a prototype only as an abstract feature vector, the model can present a concrete training example as evidence for a prediction. It should be noted that while some part-based studies refer to prototypes as ``case-based reasoning'' because prototypical patches originate from training scans \cite{barnett2021iaia,wei2024mprotonet}, we strictly reserve \textbf{case-based} for prototypes that can be retrieved or projected onto holistic patient examples instead of just an area or a region so that case-based and part-based grounding never overlap.

Prototype-based case reasoning was developed in foundational models such as PrototypeDNN \cite{li2018deep} and ProtoAttend \cite{arik2019protoattend}, and has been applied in medical imaging for exemplar-grounded lung nodule grading \cite{gallee2023interpretable}, multi-modal survival prediction \cite{yang2026protomm}, and continuous brain and spine age regression \cite{hesse2024prototype,bazargani2025progrespine}.

\subsection{Clinical Abstractions of Prototypes}
\label{sec:clinical_abstraction_of_prototypes}

Having examined the innovations across each core component of prototype-based methods, a fundamental question arises: \textit{at the highest level of abstraction, what are prototypes made of?} While the R-L-U taxonomy provides a bottom-up deconstruction of \emph{how} a method operates mechanically across granular design choices, relying on R-L-U alone requires readers to infer what a prototype work represents as a whole.

To provide this top-down summarization, we use \textbf{Clinical Abstraction} as an additional holistic annotation axis (coded alongside R-L-U in Table~\ref{tab:landmark_papers} and Appendix Table~\ref{tab:appendix_rlu_coding}) that categorizes the overarching clinical or domain concept represented by each method into seven categories:
\begin{itemize}
    \item \textbf{Class Identity:} Prototypes capture categorical centroids or class-conditional clusters in latent space to establish discriminative decision boundaries for metric learning across both image-level classification and dense segmentation (e.g., Matching Networks \cite{vinyals2016matching}, Prototypical Networks \cite{snell2017prototypical}, Relation Networks \cite{sung2018learning}, PANet \cite{wang2019panet}).

    \item \textbf{Statistics:} Prototypes capture probability distributions (such as mean and variance) rather than fixed single vectors to model diagnostic uncertainty and clinical ambiguity (e.g., probabilistic and variational prototype frameworks \cite{scott2019stochastic, sun2021amortized, zhen2020learning, wang2021variational, yuan2024effective}).

    \item \textbf{Appearance:} Prototypes capture visual features such as tissue textures, lesion patches, or reference images, represented either as raw image regions or as latent feature vectors (e.g., ProtoPNet \cite{chen2019this}, IAIA-BL \cite{barnett2021iaia}, XProtoNet \cite{kim2021xprotonet}).
    
    \item \textbf{Shape \& Anatomy:} Instead of local textures, prototypes capture spatial structure, organ boundaries, or coordinate layouts by enforcing spatial constraints on prototypes during training to preserve anatomical structure (e.g., Deformable ProtoPNet \cite{donnelly2021deformable}, PPNet \cite{liu2020part}, ALPNet \cite{ouyang2020self}, SimSID \cite{xiang2024exploiting}).
    
    \item \textbf{Semantics:} In multi-modal settings (such as images paired with reports or clinical records), prototypes capture language descriptions, clinical concepts, or medical terminology to connect visual findings with diagnostic text (e.g., cross-modal prototypes in report generation \cite{chen2021cross, wang2022cross} and vision-language representation learning \cite{cheng2023prior}).
    
    \item \textbf{Population:} Prototypes capture data distributions or feature summaries across different hospitals and patient cohorts, allowing models to learn across institutions without sharing raw patient data (e.g., FedProto \cite{tan2022fedproto}, MedFedProto \cite{zhao2025medfedproto}, FedPMR \cite{wang2025fedpmr}, Buffalo \cite{yan2024buffalo}). 
    
    \item \textbf{Function \& Dynamics:} Prototypes capture motion or physiological changes over time across sequential frames rather than static snapshots, extending dynamic video prototypes from general computer vision (e.g., DPNet \cite{trinh2021interpretable}) to model cyclic cardiac motion in clinical echocardiography (e.g., ProtoASNet \cite{gu2025protoasnet}).
\end{itemize}

These seven constituents provide a unified conceptual perspective to characterize what prototypes fundamentally represent across diverse architectures. While the R-L-U taxonomy deconstructs the operational mechanics of prototype representation, learning, and utilization, these abstractions define the core clinical or semantic role of the learned prototypes. For example, they differentiate region-level segmentation prototypes that capture organ geometry (\emph{Shape \& Anatomy}, e.g., \cite{ouyang2020self}) from those that represent generic category centroids (\emph{Class Identity}, e.g., \cite{tian2023implicit}), or differentiate global case-based prototypes that encode visual lesion traits (\emph{Appearance}, e.g., \cite{gallee2023interpretable}) from those that fuse multi-modal patient records (\emph{Semantics}, e.g., \cite{yang2026protomm}).

Building upon this methodological foundation, Section~\ref{sec:medical_image_applications} examines prototype methods across key medical imaging tasks, comprehensively analyzing how specific design choices in representation, learning strategy, and utilization address distinct clinical challenges.

\section{Medical Imaging Applications}
\label{sec:medical_image_applications}

\begin{figure*}[!t]
    \centering
    \includegraphics[width=0.95\linewidth]{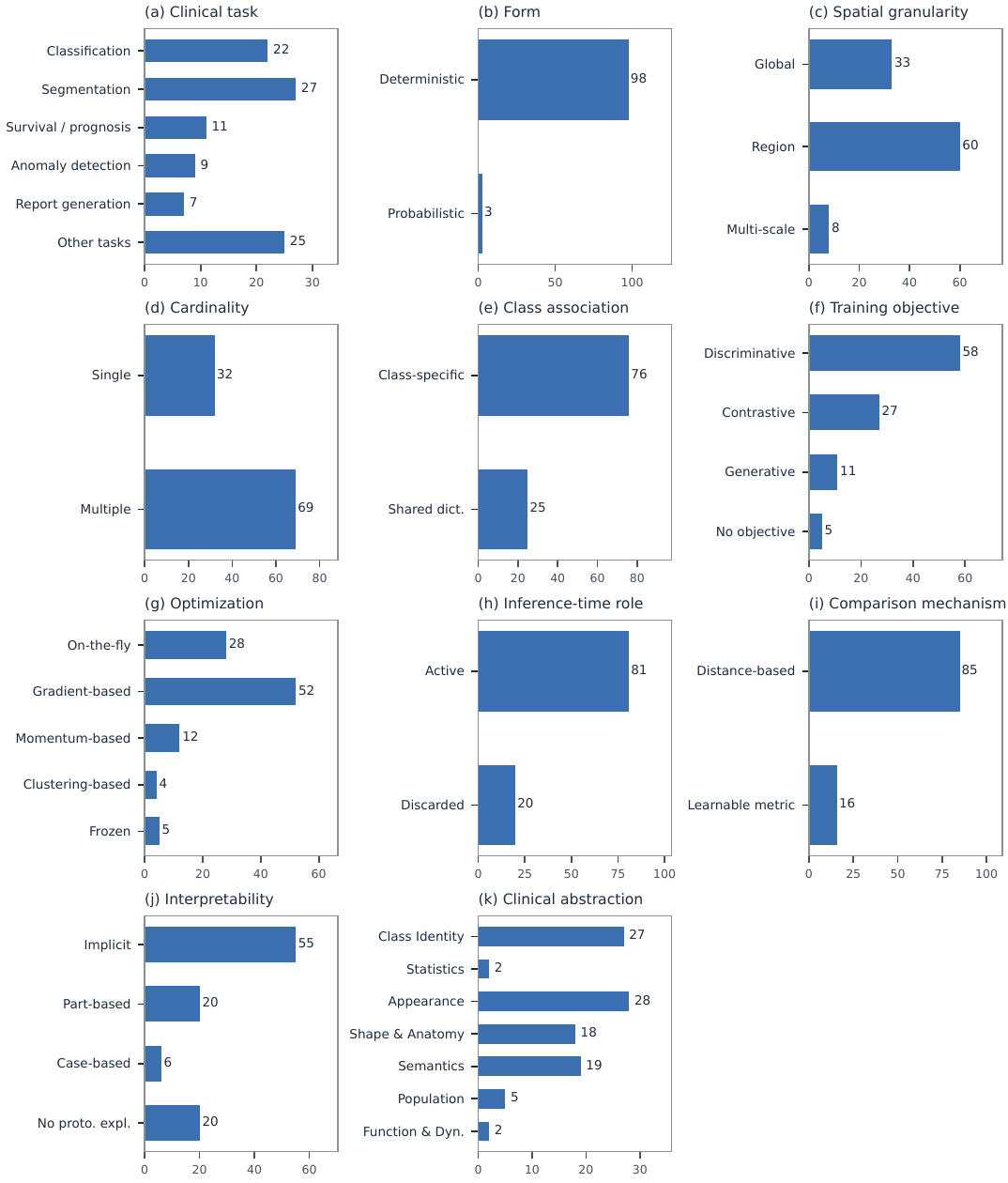}
    \caption{
    \textbf{Quantitative distribution of the 101 core medical prototype learning studies across clinical tasks and the R-L-U taxonomy.} 
    (\textbf{a}) Distribution of studies across clinical tasks. 
    (\textbf{b--e}) Prototype Representation choices across mathematical form, spatial granularity, cardinality, and class association. 
    (\textbf{f--g}) Prototype Learning Strategy distributions across training objectives and optimization mechanisms; \textit{Frozen} prototypes remain rare ($n=5$) and have no training objective (\textit{No obj.}). 
    (\textbf{h--j}) Prototype Utilization distributions across inference-time roles, comparison mechanisms, and interpretability modes. (\textbf{k}) Clinical abstraction.
    }
    \label{fig:quantitative_rlu}
\end{figure*}

\begin{figure*}[!b]
    \centering
    \includegraphics[width=0.95\linewidth]{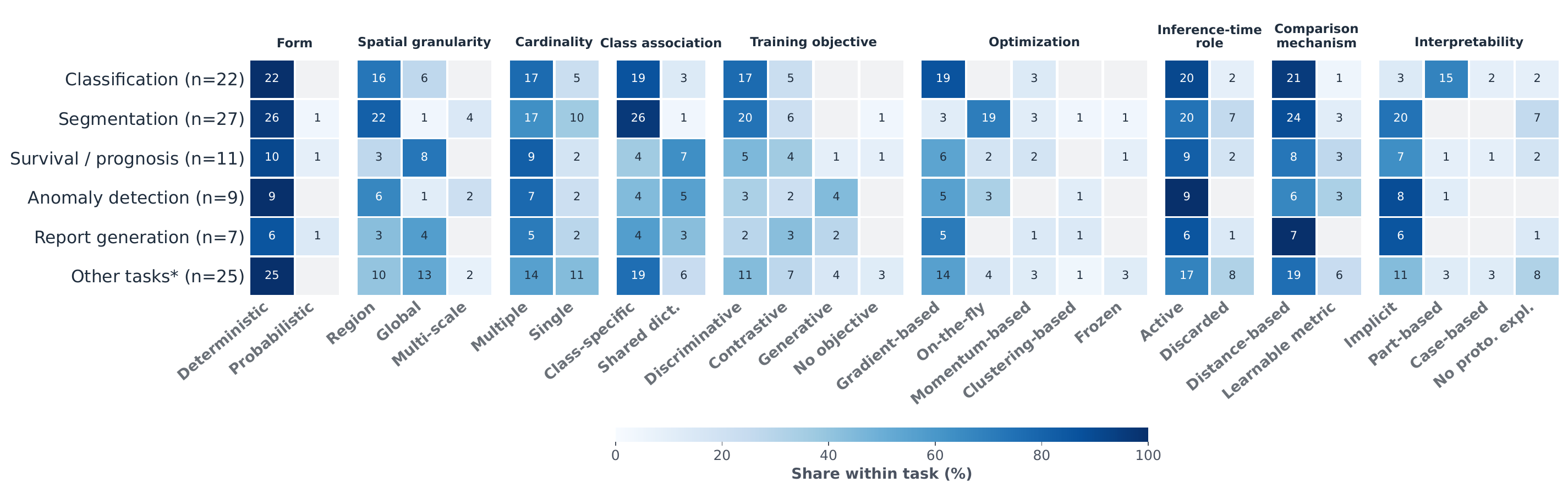}
    \caption{
    \textbf{Cross-tabulated breakdown of R-L-U components across clinical tasks ($N = 101$).} 
    Cell values denote the number of studies ($n$), while color intensity reflects the within-task share (0--100\%). 
    \textit{Other tasks} include regression, image synthesis, representation learning and pre-training, open-set recognition and OOD detection, lesion and cell detection, landmark localization, retrieval, registration, and MR reconstruction.}
    \label{fig:quantitative_across_task}
\end{figure*}

In this section, we analyze how prototype learning has been applied to medical imaging. For each study, we identify (i) the clinical problem it addresses, (ii) the foundational prototype architecture it builds upon, and (iii) the learning setting in which it operates (e.g., few-shot adaptation, federated learning, or weakly supervised learning). Table~\ref{tab:medical_applications} summarizes these clinical characteristics and core innovations across the reviewed literature. Rather than repeating categorical taxonomic labels for every study in the narrative, we provide the complete paper-by-paper R-L-U coding for all 101 core studies in Appendix Table~\ref{tab:appendix_rlu_coding} and summarize their macro-level distributions in Fig.~\ref{fig:quantitative_rlu} and Fig.~\ref{fig:quantitative_across_task}. General-vision methods are mentioned only as background and are not counted among the 101 core studies.

Taken together, Fig.~\ref{fig:quantitative_rlu} and Fig.~\ref{fig:quantitative_across_task} highlight how clinical requirements shape prototype design across the 101 core studies. Since the five core criteria were applied only to classification and segmentation, some of the cross-task differences in these figures also reflect how the studies were selected. The core studies cover segmentation ($n=27$), classification ($n=22$), survival and prognosis prediction ($n=11$), anomaly detection ($n=9$), report generation ($n=7$), and other tasks ($n=25$). While deterministic ($98/101$), region-level ($60/101$), multi-prototype ($69/101$), and class-specific ($76/101$) representations are widely favored, learning and utilization strategies split sharply by task (Fig.~\ref{fig:quantitative_across_task}). All codes follow the mechanism described in each paper's method section rather than the authors' terminology.

Classification predominantly optimizes persistent prototypes via gradient descent ($19/22$) to deliver explicit part-based ($15/22$) or case-based ($2/22$) visual explanations. Conversely, segmentation ($19/27$ on-the-fly) and anomaly detection focus more on data efficiency and representation learning, in which prototypes act as dynamic support references or consistency regularizers. These counts describe how often each design choice is used in the core studies and do not show that a choice improves accuracy, robustness, or the quality of explanations.

We organize the remainder of this section by clinical task. Within each task, we trace the development of prototype-based methods and examine how their designs are adapted to address specific clinical challenges.


\subsection{Medical Image Classification and Disease Diagnosis}
\label{sec:image_classification}

In general vision, Snell et al. \cite{snell2017prototypical} introduce Prototypical Networks for few-shot metric classification, using Euclidean distances to class centroids computed on-the-fly from support samples. 

More recently, to address structured diagnostic taxonomies, Yu et al. \cite{yu2025hierarchical} restructure prototype layers into a prototypical decision tree whose nodes are each anchored to one class of the diagnostic taxonomy, capturing coarse-to-fine pathological relationships in skin lesion classification.

In federated learning, Tan et al. \cite{tan2022fedproto} introduce the general-vision FedProto framework by adapting the principle of Prototypical Networks \cite{snell2017prototypical} to distributed settings. Rather than transmitting large, uninterpretable neural network weights across distributed clients, FedProto replaces standard parameter aggregation with lightweight class-centroid prototype communication between decentralized clients. Building upon this communication scheme, Zhao et al. \cite{zhao2025medfedproto} develop MedFedProto, which extends federated prototypical learning into a semi-supervised regime by incorporating dual-level momentum-based exponential moving average (EMA) prototype updates and prototype-guided alignment to exploit abundant unlabeled clinical scans. 
Prototype-based federated learning has also been applied to MR reconstruction (Section~\ref{sec:emerging_frontiers}). 

Shifting from global class centroids to localized spatial representations, Chen et al. \cite{chen2019this} propose ProtoPNet for natural images, optimizing multiple localized prototypes per class to deliver intrinsic part-based interpretability. Barnett et al. \cite{barnett2021iaia} adapt ProtoPNet to digital mammography in IAIA-BL, replacing max-pooling with top-$k$ average pooling when reducing region similarity maps to similarity scores and aligning part-based prototypes with clinical BI-RADS marginal features. Likewise, Singh and Yow \cite{singh2021interpretable} deploy multiple rectangular and square prototypes larger than 1×1, providing patch explanations (activation heatmaps) for interpretable COVID-19 detection.

To overcome the rigid spatial granularity of standard patch prototypes, Kim et al. \cite{kim2021xprotonet} modify ProtoPNet to construct XProtoNet, introducing an occurrence module that compares variable-sized regional features against prototypes rather than constraining representations to static 1×1 latent patches, which boosts detection accuracy in chest radiography. In a related effort to constrain feature representations, Djoumessi et al. \cite{djoumessi2024actually} introduce Proto-BagNet by replacing the conventional backbone with a BagNet feature extractor to enforce small, fixed receptive fields. Additionally, they replace global max-pooling with average top-$k$ pooling over region similarity maps to aggregate evidence from multiple discriminative areas. Transitioning from purely convolutional backbones to attention mechanisms, Demir et al. \cite{demir2024explainable} extend XProtoNet \cite{kim2021xprotonet} by replacing the conventional distance metric with a learnable metric mechanism, using self-attention to match dynamically pooled regional queries directly against prototype keys.

Beyond rigid 2D planar patches, Wei et al. \cite{wei2024mprotonet} introduce MProtoNet, an interpretable model for brain tumor classification in 3D MRI. This method inflates standard 2D spatial operations into 3D volumetric convolutions within a ResNet backbone, utilizing the add-on module from ProtoPNet \cite{chen2019this} and the mapping module from XProtoNet \cite{kim2021xprotonet} to extract localized 3D region prototypes for glioma grading. Extending prototype learning to spatiotemporal video data, Gu et al. propose ProtoASNet \cite{gu2025protoasnet}, capturing cyclic cardiac motion patterns via dynamic echocardiography prototypes coupled with prototype-based abstention to account for predictive uncertainty in aortic stenosis grading.

To improve semantic grounding and clinical interpretability, recent approaches anchor prototypes directly to verified medical concepts rather than unconstrained feature clusters. Gallée et al. \cite{gallee2023interpretable} extend ProtoPNet by anchoring prototypes to attribute capsule vectors rather than convolutional feature maps, using the nearest prototype's ground-truth label to determine attribute ratings during inference. Addressing ProtoPNet's semantic gap and rigid class-budget allocation, Nauta et al. \cite{nauta2023interpreting} adapt PIP-Net \cite{nauta2023pip}, which learns a shared dictionary of patch prototypes via self-supervised contrastive learning; applied to musculoskeletal radiographs and dermoscopic images, this enables clinicians to audit data quality and debug shortcut learning by directly disabling confounded prototypes. While ProtoPNet pairs image patches directly with disease labels, it still requires clinicians to interpret what radiological finding a patch actually captures. Based on that limitation, Huy et al. \cite{huy2025interactive} assign prototypes to explicit clinical concepts (e.g., lung opacity, cardiomegaly) before predicting disease, ensuring each learned patch represents a verified medical finding rather than an arbitrary visual pattern.

To reconcile part-based interpretability with the representational power of unconstrained models, Wang et al. \cite{wang2022knowledge} deploy knowledge distillation, transferring representations from an unconstrained black-box classifier into an interpretable student prototype network and combining them into an ensemble during inference. They subsequently extend this in InterNRL \cite{wang2024interpretable}, enforcing spatial transformation equivariance on prototype similarity maps to regularize the global classifier through reciprocal pseudo-labeling. Addressing visualization fidelity, Gautam et al. introduce Prototypical Relevance Propagation (PRP) \cite{gautam2023looks} and evaluate it on chest X-rays \cite{gautam2022demonstrating}. This framework refines part-based interpretability in ProtoPNet by replacing heuristic upsampled heatmaps with precise, layer-wise relevance propagation maps to ensure visual attributions faithfully reflect the network's reasoning.
\begin{figure*}[!b]
    \centering
    \includegraphics[width=0.9\linewidth]{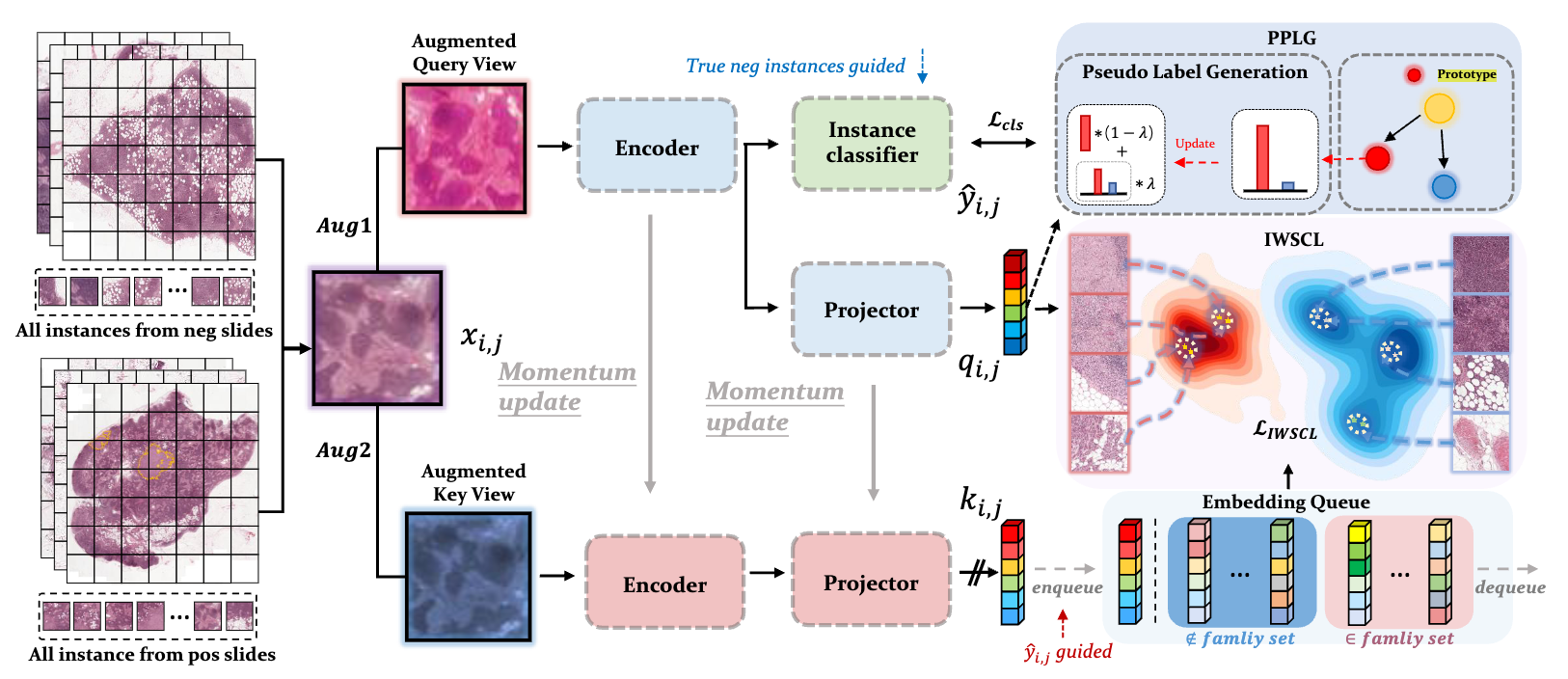}
    \caption{
    \textbf{Prototype-guided pseudo-labeling for instance-level WSI classification.} 
    Patches from negative and positive slides are augmented into a query view and a key view, encoded by a query branch and a momentum-updated key branch, and trained with instance-level weakly supervised contrastive learning (IWSCL) using an embedding queue. In the prototype-based pseudo-label generation (PPLG) module, one prototype per class (positive and negative) is updated with momentum from the instance embeddings, and each instance's pseudo-label is refined from its similarity to these prototypes to supervise the instance classifier ($\mathcal{L}_{cls}$). The prototypes are used only during training, and inference relies on the instance classifier alone (image from \cite{qu2024rethinking}).
    }
    \label{fig:qu2024rethinking_pipeline}
\end{figure*}

In whole-slide image (WSI) classification, several multiple instance learning (MIL) frameworks adapt prototype mechanisms to handle weakly annotated histological bags. Rymarczyk et al. \cite{rymarczyk2022protomil} adapt ProtoPNet in ProtoMIL by combining patch-level prototype similarities with attention pooling to aggregate instance evidence into bag-level predictions. To overcome ProtoMIL's reliance on sparse representative patches, Yang et al. propose TPMIL \cite{yang2023tpmil}, which employs trainable prototypes optimized end-to-end via gradient descent with a clustering regularization loss during training while preserving them at inference to generate distance-based interpretability heatmaps. Focusing on dual-level supervision, Qu et al. \cite{qu2024rethinking} modify multiple instance learning by introducing prototype-guided instance classification to effectively accomplish both instance-level and bag-level classification (Fig.~\ref{fig:qu2024rethinking_pipeline}).

For complex diagnostic structures exhibiting multi-label co-occurrence and severe class skewness, recent frameworks leverage hybrid prototype formulations. Wang et al. \cite{wang2025cross} develop CIPL for multi-label classification, learning global cross-image disease centroids alongside intra-image localized patch prototypes to decouple co-occurring thoracic and fundus pathologies. For long-tailed recognition, Zhu et al. \cite{zhu2024sfpl} introduce SFPL, a framework that establishes global prototypes to capture majority-class heterogeneity, updating them via a contrastive learning paradigm with momentum-based tracking in real time via PCL \cite{li2020pcl}.

Finally, beyond uni-modal setups, multi-modal supervision can be leveraged during training to ground visual interpretations. For example, Zhang et al. \cite{zhang2024disease} proposed Disease Prototype Learning (DPL), where learnable prototypes are initialized solely from clinical text descriptions and updated via gradient descent. In their framework, prototypes serve only as geometric regularizers during training to structure the vision-language model (VLM) latent space and mitigate overfitting in data-scarce settings, and they are discarded at test time so that classification relies entirely on the adapted visual encoder and a linear head. Meanwhile, QPMIL-VL \cite{gou2025queryable} instead maintains a queryable prototype pool in WSI classification, in which a prototype is a pair of key (representing the aggregation of many image patch features) and prompt (representing text) vectors living in a vision-language feature space.

\noindent\textbf{Summary.}
Across disease classification ($n=22$), prototype architectures have evolved along two main tracks: metric-based global centroids for data-scarce or federated environments and localized part-based networks (building on ProtoPNet) that provide intrinsic visual explanations (Fig.~\ref{fig:quantitative_across_task}). Recent works have progressively adapted rigid 2D patch constraints through 3D volumetric and spatiotemporal methods and have used prototypes to summarize whole-slide images. Most studies learn persistent prototypes by gradient descent ($19/22$) and keep them active at inference ($20/22$), and $17$ of the $22$ provide part-based ($15$) or case-based ($2$) explanations, the highest share of any task.

\subsection{Medical Image Segmentation}
\label{sec:image_segmentation}

The cost of dense expert annotation has made few-shot segmentation a major application of prototype learning in medical imaging. PANet \cite{wang2019panet} established a general approach in which foreground and background prototypes are computed from labeled support images and matched against query pixels. Ouyang et al. \cite{ouyang2020self} adapted this approach to medical imaging in ALPNet, using superpixel pseudo-labels for self-supervised episodic training and adaptive local prototype pooling to address foreground-background imbalance. This allowed the model to learn transferable representations without manually annotated training classes. Many later few-shot methods report Dice scores on the abdominal CT, abdominal MRI, and cardiac MRI benchmark introduced with ALPNet. We do not compare these scores directly, because the methods differ in self-supervision (superpixels or supervoxels) and other training and evaluation details, and not all of them state which of the benchmark's two settings they use.

\begin{figure}[!b]
    \centering
    \includegraphics[width=0.9\linewidth]{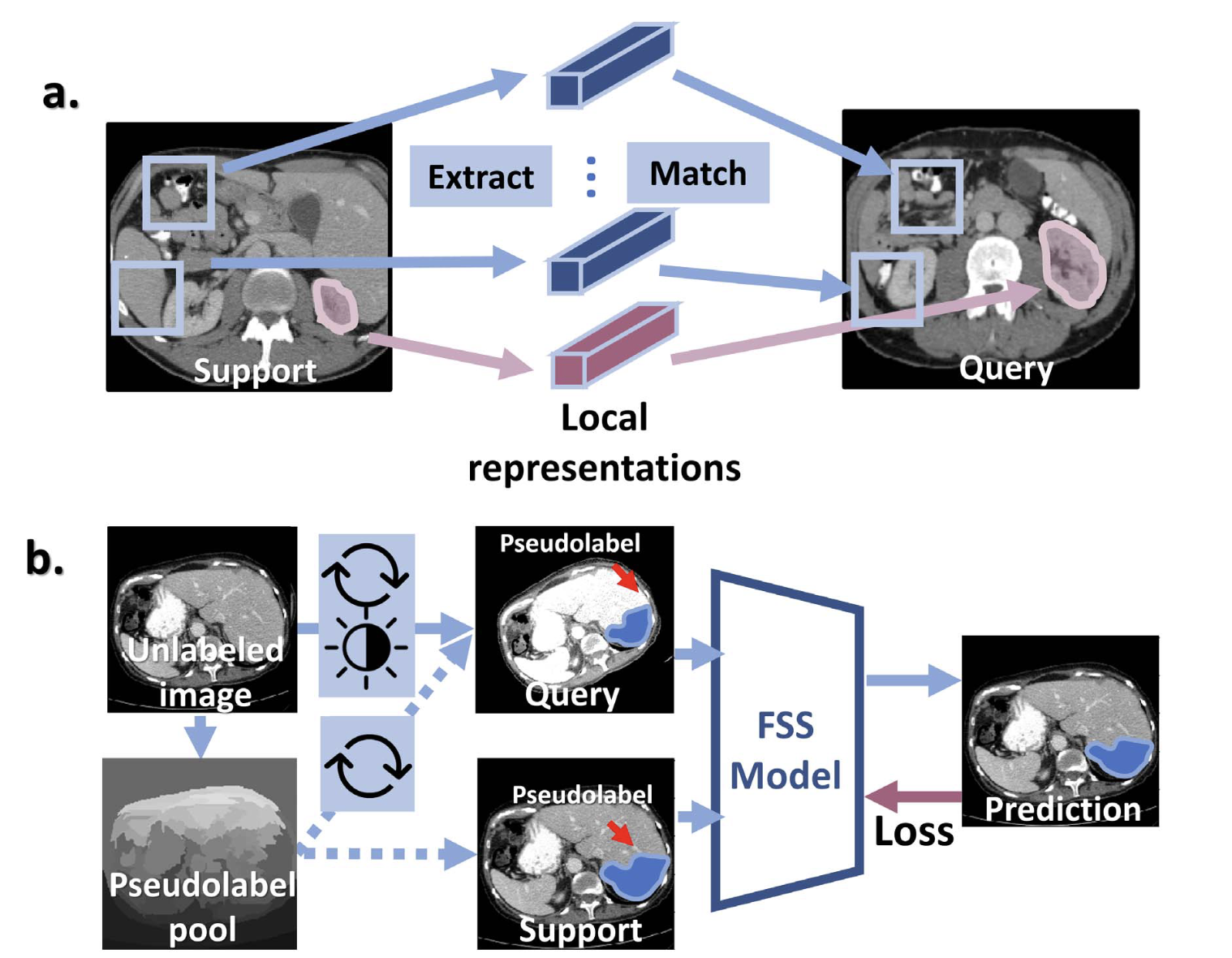}
    \caption{
    \textbf{Local prototype matching for annotation-efficient medical image segmentation}, illustrated by ALPNet \cite{ouyang2020self}. Rather than representing each class with one global embedding, the model derives local support representations and matches them to spatially corresponding query regions. This is particularly important for heterogeneous anatomical structures and background tissue, whose appearances cannot be adequately summarized by a single prototype. ALPNet learns this mechanism through superpixel-based self-supervised episodes: pseudo-labeled images provide support--query pairs, while appearance and geometric transformations of the query mimic inter-patient variation. The local matching principle motivates later multi-prototype segmentation methods that explicitly retain regional anatomical variation (image from \cite{ouyang2020self}).}
    \label{fig:intuition_alpnet}
\end{figure}

For organs with heterogeneous appearance, averaging all foreground features into one prototype can hide differences between anatomical sub-regions. Fig.~\ref{fig:intuition_alpnet} illustrates why local matching is useful: it preserves spatially specific foreground and background information that would be lost when an entire class is compressed into a single vector. Yu et al. \cite{yu2021location} compute local prototypes on a grid of the support image and match each query location only to nearby prototypes. PAMI \cite{zhu2023partition} retains this regional information through multiple foreground prototypes and debiases their contributions before combining predictions. RPT \cite{zhu2023few} instead selects and refines regional information to rectify a global prototype. Following ALPNet, Zhao et al. \cite{zhao2024dual} replace global prototypes with on-the-fly part prototypes that complement local representations in self-supervised one-shot segmentation. DSPNet \cite{tang2025few} also accounts for the different characteristics of foreground and background: on-the-fly foreground prototypes are fused across channels, while sparse channel-aware regulation handles complex background features.

Local prototypes may still inadequately represent the transition between an organ or lesion and its surroundings. EML \cite{wang2025edge} specifically samples this transition by eroding the support mask and applying norm-weighted pooling to construct an inner-boundary prototype, which is combined with other foreground prototypes for query matching. CoW \cite{jiang2025concentrate} takes a different approach, using support self-prediction errors to identify poorly represented foreground and background regions. It constructs hard prototypes from these regions and fuses their matching scores, with additional boundary supervision. In supervised lesion segmentation, Proto-MSN \cite{yang2026proto} incorporates boundary/non-boundary prototype learning and foreground/background subclustering to refine the features used by its main segmentation module.

Other few-shot methods improve segmentation by using relationships among classes or image regions. PMCR \cite{zhang2024prototype} combines optimal-transport matching between representative support and query prototypes with class-relation reasoning. PGRNet \cite{huang2024prototype} uses support-conditioned prototypes derived from horizontal and vertical perspectives to guide information exchange within a query graph. The refined query features are then decoded into a segmentation mask. Here, prototypes help determine how features interact, rather than serving only as references for independent pixel classification.

Differences between hospitals and imaging modalities pose a further problem: support features may not correspond well to the anatomy or appearance of the query image. For cross-institution pelvic MRI, Li et al. \cite{li2023prototypical} address anatomical correspondence through 3D support-to-query registration, local prototype matching, and support-mask conditioning. PFMNet \cite{wang2024pfmnet} focuses on appearance differences, using a small labeled target set to map target-domain class prototypes toward learned source-domain exemplar vectors. This mapping transforms target features into a source-like space for decoding; the encoder, decoder, and exemplar vectors are jointly optimized. Without source images or target labels, Yu et al. \cite{yu2023source} keep the source classifier weights fixed as class prototypes and align target features to them. Meanwhile, DAPSAM \cite{wei2024prompting} stores domain-adaptive prototypes in a learnable memory and uses them to prompt the Segment Anything Model (SAM) \cite{kirillov2023segment} for domain generalization.

Prototype learning is also used to exploit unlabeled images within a training cohort. CPCL \cite{wu2022all} introduces forward and backward prototypical consistency between labeled and unlabeled images. IEPAlign \cite{tian2023implicit} builds on this approach with multiple image-derived prototypes, prediction voting, and explicit feature alignment. Liu et al. \cite{liu2023prototype} use spatially aware foreground prototypes in voxel-to-prototype contrastive learning, while KP$^{2}$L \cite{liu2026kp2l} incorporates shape, texture, and semantic information through multi-scale prototype predictions. For weakly supervised segmentation from sparse labels, ProCNS \cite{liu2025procns} uses class prototypes to refine pseudo-labels and suppress label noise during training.

Such supervision depends on how reliably the prototypes represent ambiguous or noisy image regions. Yuan et al. \cite{yuan2024effective} model pixel features and class prototypes as Gaussian distributions, using uncertainty-aware matching and Bayesian updates in semi-supervised segmentation. Evidential Prototype Learning \cite{he2025evidential} uses evidential fusion and voxel-level uncertainty masking to limit unreliable supervision. Uncertainty, however, is only one concern when a disease or anatomical structure has several distinct appearances. Liu et al. \cite{liu2026embracing} address this heterogeneity with multiple intensity-aligned prototypes, refine the prototypical representation space, and transfer the resulting information to a dual-branch segmentation network. In fully supervised segmentation, SIPL \cite{liang2024advancing} instead adapts a single class prototype to each image by fusing a learnable common prototype with an instance-specific prototype pooled under self-generated pseudo-masks.

A smaller group of studies considers how prototypes can support different annotation and deployment workflows. Proto2Seg \cite{pan2023human} uses histopathology patches to form multiple regional prototypes, which a pathologist assigns to specific classes to generate pseudo-masks for training before the prototypes are discarded at inference. ProGIS \cite{ge2026progis} builds a prototype from a single user scribble and uses it to find all tissue regions of the same type in a slide. PRIME \cite{zhu2025prime} considers the addition of new segmentation classes after initial training. It combines class prototypes with structural alignment and pixel-level contrastive learning, using agreement between the old and current models to select pixels for updating old-class prototypes.

\noindent\textbf{Summary.}
In medical image segmentation ($n=27$), prototype learning serves primarily to combat dense annotation scarcity in few-shot, semi-supervised, weakly supervised, and domain-adaptive regimes. Rather than learning fixed prototype parameters across the dataset, segmentation frameworks predominantly extract regional prototypes on-the-fly from support masks or current batches (Fig.~\ref{fig:quantitative_across_task}), evolving from single global foreground vectors toward multi-prototype local, boundary-aware, and uncertainty-weighted representations. Accordingly, $19$ of the $27$ studies build prototypes on-the-fly and $26$ use class-specific prototypes, while none exposes part- or case-based explanations; in $7$ studies, the prototypes are discarded after training.

\subsection{Survival and Prognosis Prediction}
\label{sec:survival_prediction}
Survival prediction estimates the time to an event, such as death or recurrence, from follow-up data that are often censored. In computational pathology, the input is usually a gigapixel whole-slide image (WSI) together with a genomic profile, and prototypes compress thousands of patches and genes into a few representative features that can be compared across patients. Liu et al. \cite{liu2025prototype} propose ProSurv, a framework that predicts patient survival using tissue images and genomic data, even when one modality is missing. ProSurv creates separate ``prototype banks'' of typical pathology and genomic features grouped by patient survival times. When a patient lacks one data type (such as genomic data), the model uses their available data as a query to pull matching features from the missing modality's prototype bank via cross-attention. This reconstructs the missing information, allowing the model to make reliable survival predictions across varied clinical settings. Yu et al. \cite{yu2026prime} address missing modalities during self-supervised multi-modal pre-training with one learnable prototype bank shared by WSIs, bulk RNA sequencing, and pathology reports. Each observed modality is softly assigned to the prototypes by cosine similarity, the assignments are averaged into a patient-level consensus, and the missing modality is imputed as the weighted sum of prototypes.

Other methods derive prototypes from the morphology of each slide. PS3 \cite{raza2025ps3} computes on-the-fly prototypes for each modality of WSI, text report, and gene features before fusing them using a multi-modal fusion module to predict survival risk. Wang et al. \cite{wang2026protosurv} similarly cluster each slide's patch features on-the-fly into morphological prototypes and fuse them with pathway features. TTA \cite{liu2026together} instead learns one set of prototypes shared across the dataset and by both WSI patches and genomic pathways; tokens of both modalities are assigned to these prototypes by cosine similarity with optimal transport, and the tokens grouped under each prototype are fused for survival prediction.

\begin{figure*}[!t]
    \centering
    \includegraphics[width=0.9\linewidth]{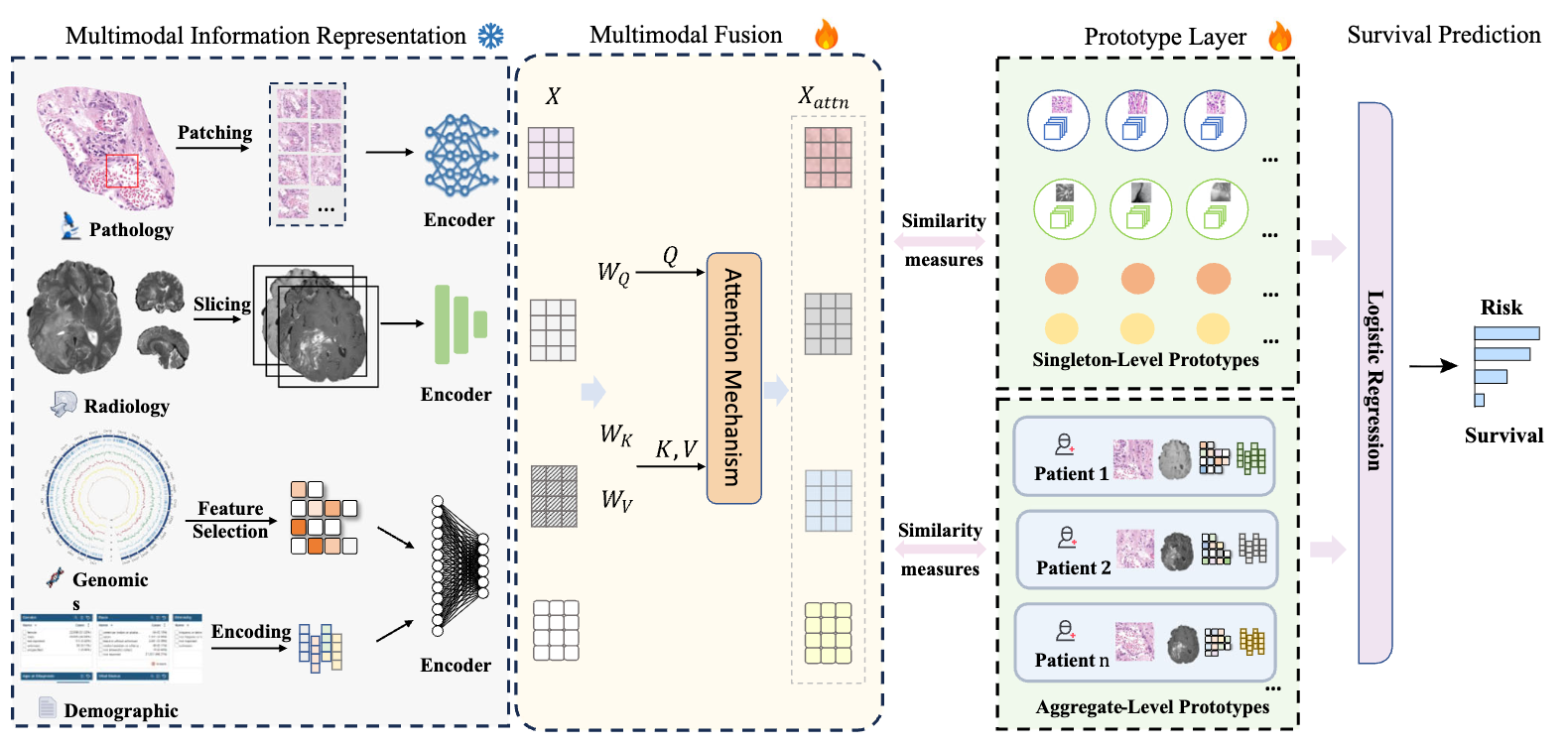}
    \caption{
    \textbf{Multi-modal prototype reasoning for survival prediction.} 
    In ProtoMM, shown here as a representative architecture, frozen uni-modal encoders embed pathology, radiology, genomic, and demographic data, which an attention-based module fuses into a patient representation. This representation is compared with learned singleton-level prototypes (one modality) and aggregate-level prototypes (fused patient cases), and the resulting similarity scores serve as a case-based explanation and are passed to a linear (logistic regression) layer to predict survival risk (image from \cite{yang2026protomm}).
    }
    \label{fig:protomm_pipeline}
\end{figure*}

In cancer survival analysis, several methods learn prototypes to enhance prognostic interpretability. For instance, FeatProto \cite{jiang2026multimodal} maintains a class-specific dictionary of multi-modal prototypes, which are updated via momentum (EMA) and matched through distance metrics to guide prediction. ProtoMM \cite{yang2026protomm} learns prototypes at both the multi-modal and the single-modality level to provide case-based explanations for brain cancer survival (Fig.~\ref{fig:protomm_pipeline}). ProtoSurv-X \cite{kumari2026protosurvx} summarizes radiomic, deep, and clinical features of glioma MRI with a shared memory of learnable prototypes.

In radiology and longitudinal data, prototypes mainly address limited labels or unstable training. ProtoSurv by Liu et al. \cite{liu2026protosurv} computes one prototype per prognosis class from a frozen CT foundation model and uses them to pseudo-label unlabeled scans. DuoProto \cite{yu2026learning} aligns class prototypes between the multi-phase and single-phase CT branches to empower the main network to perform robust early recurrence prediction. ImProNCDE \cite{wang2026impro} pulls the latent state of a neural controlled differential equation toward learned prototypes at each visit to stabilize longitudinal ophthalmic prognosis.

\noindent\textbf{Summary.} In survival and prognosis prediction ($n=11$), prototypes mainly compress whole-slide images and molecular data and often encode clinical or molecular semantics ($6/11$), and seven of the eleven studies share one prototype dictionary across classes or risk groups instead of using class-specific prototypes. Part-based ($1/11$) and case-based ($1/11$) explanations are still uncommon (Fig.~\ref{fig:quantitative_across_task}).
\subsection{Medical Anomaly Detection and Localization}
\label{sec:anomaly_detection}


In unsupervised anomaly detection, normal prototypes extracted from each image can suffer from prototype collapse, where a small subset of dominant prototypes suppresses representation diversity. To address this limitation, Li et al. \cite{li2025diverse} propose DNP-ConFormer, which couples a trainable encoder with a momentum-updated teacher to stabilize domain adaptation while deploying a lightweight Prototype Extractor. Under the supervision of a novel Diversity-Aware Alignment (DAA) loss that enforces balanced prototype utilization, their framework extracts informative, diverse normal prototypes and injects them into the decoder via cross-attention to guide the reconstruction process. Extending prototype-guided reconstruction to incomplete multi-sequence MRI, Wu et al. \cite{wu2025anyad} develop AnyAD, which extracts intrinsic normal prototypes (INPs) from each input in a unified architecture designed to handle arbitrary and missing modality subsets without retraining.

Instead of extracting prototypes from each image, another group of methods stores normal anatomical patterns in learned memory matrices. SQUID \cite{xiang2023squid} maintains space-aware memory queues to dynamically store features of normal image patches as prototypes across training iterations. These cached representations guide feature-level in-painting alongside masked shortcuts to ease feature aggregation. To reduce optimization complexity, SimSID \cite{xiang2024exploiting} replaces the memory queues with a learnable space-aware memory matrix coupled with teacher-student distillation, effectively removing masked shortcut overhead while relaxing the requirement for disease-free training data. Similarly, SP-Mamba \cite{pan2025spmamba} learns position-aware prototypes that model the regular anatomy of radiographs.

When abnormal images are available for training, prototypes can also represent the anomalies themselves. In supervised and few-shot settings, UPformer \cite{huang2024uncertainty} stores prototypes of anomalous lesions and estimates uncertainty at their boundaries, while Lee et al. \cite{lee2025fewshot} perform few-shot anomaly detection by constructing normal and abnormal prototypes computed as the mean of text-aligned BiomedCLIP embeddings and evaluating test images based on Euclidean distance to these prototypes. Kobayashi and Yamaguchi \cite{kobayashi2024simple} adapt PIP-Net \cite{nauta2023pip} to anomaly detection on bone radiographs; because no local pattern belongs only to normal images, each part prototype is scored only for the abnormal class, so fewer prototypes are needed. In weakly supervised disease detection, where labels are given only for bags of instances, She et al. \cite{she2024memory} note that normal-only prototypes can miss subtle anomalies and therefore learn two memory banks, one of normal and one of abnormal patterns, together with losses that separate normal from abnormal features; they evaluate on prostate WSIs and lung CT.

\begin{figure*}[!t]
    \centering
    \includegraphics[width=0.9\linewidth]{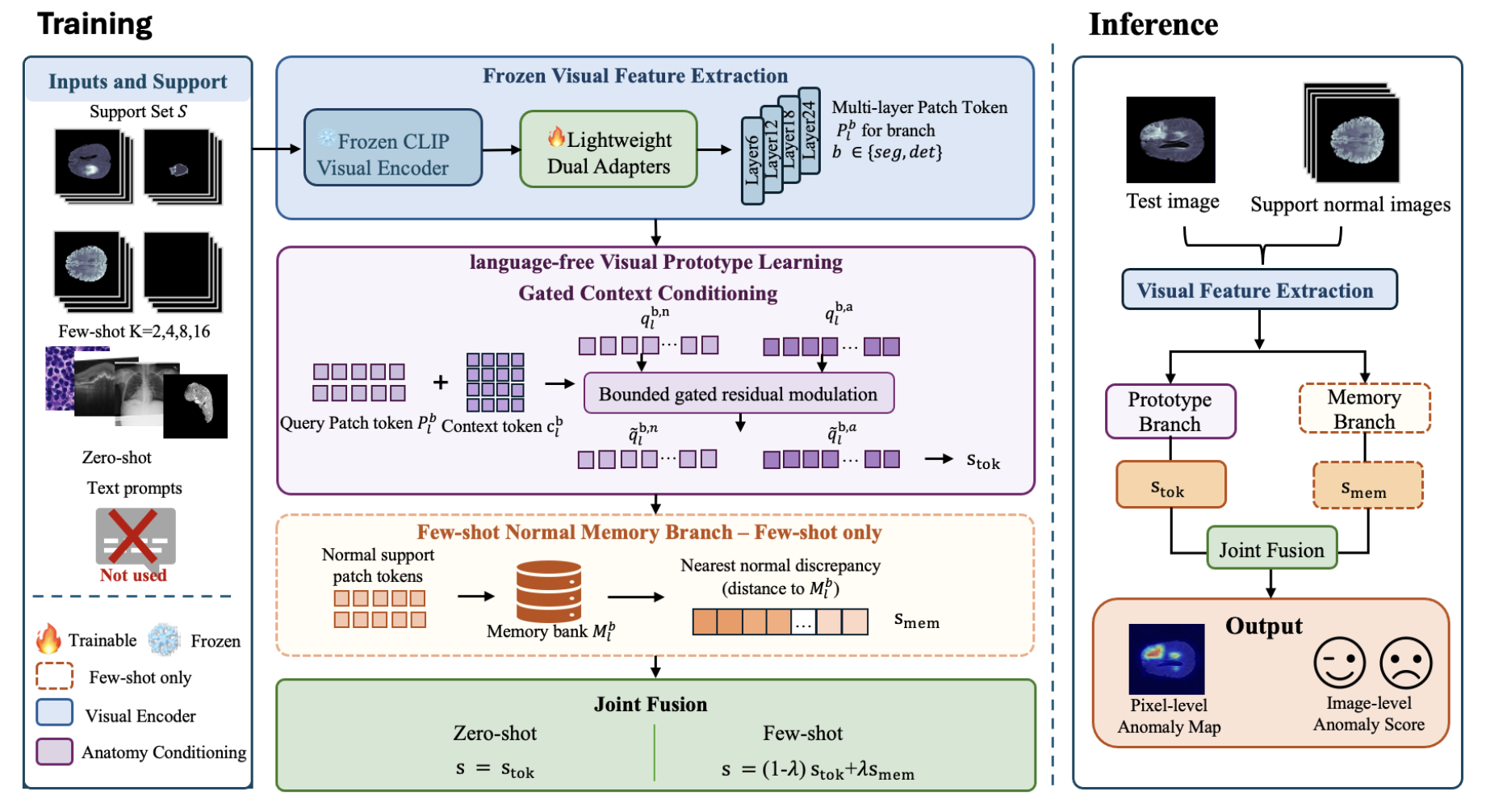}
    \caption{
    \textbf{Context-conditioned visual prototypes for zero- and few-shot anomaly detection.} 
    In ReCAP, shown here as a representative architecture, a frozen CLIP visual encoder with lightweight adapters extracts patch tokens from several layers. Instead of text prompts, one learnable normal and one abnormal prototype per layer are shifted toward each input by a bounded gated modulation of its global context, and patch-level anomaly scores are computed from cosine similarity to these prototypes. In the few-shot setting, a memory bank of normal support patches adds a nearest-normal distance score, which is fused with the prototype score to produce pixel-level anomaly maps and image-level anomaly scores (image from \cite{wan2026beyond}).
    }
    \label{fig:recap_pipeline}
\end{figure*}

In zero-shot anomaly detection, Wan et al. \cite{wan2026beyond} propose ReCAP (Fig.~\ref{fig:recap_pipeline}) for medical anomaly detection, learning normal and abnormal visual prototypes directly within the CLIP visual space together with its adapters. To adapt to diverse anatomical structures across imaging modalities, the framework applies a bounded gated modulation to dynamically shift these base prototypes using a global image context descriptor.

\noindent\textbf{Summary.} 
In medical anomaly detection ($n = 9$), prototypes are primarily designed to model compact reference manifolds of healthy tissue or canonical normal-abnormal states without requiring dense lesion labels. Within our R-L-U taxonomy (Fig.~\ref{fig:quantitative_across_task}), these methods rely almost entirely on implicit utilization ($8/9$), splitting between generative reconstruction pipelines that query shared memory banks or on-the-fly intrinsic normal prototypes, and discriminative frameworks that score anomalies by their distance to normal and abnormal prototypes. Four of the nine studies are trained with a generative objective ($4/9$), and five of the nine store prototypes in a shared dictionary.

\subsection{Medical Report Generation}
\label{sec:report_generation}

In radiology report generation, early efforts exploring aligned latent representations introduced memory mechanisms that function identically to prototypical representations. Specifically, Chen et al. \cite{chen2021cross} proposed a Cross-modal Memory Network (CMN), where a shared memory matrix acts conceptually as a bank of cross-modal prototypes. Although termed a ``shared memory,'' each slot in the matrix serves as a latent multi-modal prototype that captures radiological visual patterns alongside standardized diagnostic phrasing. In their framework, both visual and textual queries interact with this shared bank through a key-value addressing mechanism using top-$k$ routing to retrieve relevant prototypes for both the encoder and decoder. Building upon the idea of shared cross-modal representations, Wang et al. \cite{wang2022cross} explicitly formalized this paradigm through prototype learning in XPRONET. Unlike the purely parametric memory slots in \cite{chen2021cross}, XPRONET constructs a cross-modal prototype matrix by clustering pre-trained visual and linguistic representations guided by clinical pseudo-labels. These discrete prototypes are refined end-to-end via gradient descent. During generation, extracted image tokens query the shared prototype matrix to retrieve corresponding prototypes, which are fused with the visual features to condition a Transformer decoder.

Extending the multi-modal prototype in report generation to federated learning, Yan et al. \cite{yan2024buffalo} introduce Buffalo, which projects visual patch features into the language embedding space and clusters each modality's representations independently via K-Means, modeling each cluster as a Gaussian prototype with the cluster mean and standard deviation. Clients then upload these local prototypes and trainable model parameters to the server for aggregation, rather than communicating full foundation model parameters, enabling collaborative report generation and medical visual question answering (Med-VQA) without exposing raw patient records.

Pellegrini et al. \cite{pellegrini2026prototype} further extend prototypes into fine-grained structured radiology reporting with an auxiliary branch that adds prototype-based guidance at inference. The framework keeps an external bank with one image-derived prototype per answer option and uses the retrieved prototypes to correct the predicted logits. For long-tailed findings, CSMH-RRG \cite{yang2026clinical} learns hyperbolic prototypes for each finding and state and uses the predicted states to prompt the report decoder. Dia-LLaMA \cite{chen2025dia} uses a similar design for chest CT: it learns a normal and an abnormal prototype for each disease with a contrastive loss, and the disease states predicted from feature-prototype similarity are written into diagnostic prompts for a LLaMA2-7B report generator.

AP-ISG \cite{zhang2024attribute} instead anchors text prototypes to predefined clinical attribute concepts from the Chest ImaGenome dataset \cite{wu2021chest}, such as diseases, anatomical findings, and tubes and lines, using GloVe word embeddings. These prototypes, after being projected into the visual feature space, are updated by gradient descent with a contrastive loss between the prototypes and image region features. During inference, the similarity between the prototypes and the region features of a test image is used to add attribute nodes to a scene graph that conditions the GPT-2 report decoder.

\noindent\textbf{Summary.}
In radiology report generation ($n = 7$), prototypes act as cross-modal bridges that align recurring visual patterns with standardized diagnostic phrasing. In terms of R-L-U design (Fig.~\ref{fig:quantitative_across_task}), this task is characterized by cross-modal prototypes that remain implicit during inference ($6/7$). All seven papers compare features with prototypes by a distance function, and three store them in a shared cross-modal dictionary.

\subsection{Other Clinical and Anatomical Tasks}
\label{sec:emerging_frontiers}

Beyond classification, segmentation, survival prediction, anomaly detection, and report generation, prototype learning has been adapted to a diverse spectrum of specialized clinical tasks, ranging from multi-modal pre-training to geometric coordinate regression.

In multi-modal pre-training, MGCA \cite{wang2022multi} learns a shared bank of 500 cross-modal prototypes over global chest X-ray and radiology report embeddings to capture disease-level semantics alongside its instance- and token-level alignment modules. During pre-training, each global image and report representation is softly assigned to these prototypes (via Sinkhorn-Knopp) and trained to predict the other's assignment, pulling different patients with the same condition toward the same prototype instead of pushing them apart as negatives. Although these prototypes are discarded after pre-training, adding them consistently improved downstream performance across all nine settings of the authors' ablation. Cheng et al.~\cite{cheng2023prior} introduce PRIOR to ground visual representations in report semantics using sentence-level prototypes. Instead of relying on unconstrained text embeddings, their framework maintains a learnable prototype memory bank that captures canonical clinical findings. Each report sentence is dynamically assigned to its nearest prototype via an explicit distance loss, pulling the prototype toward shared clinical concepts. During pre-training, learnable queries interact with image features to reconstruct and predict this unordered set of sentence prototypes in parallel, encouraging the visual encoder to capture diverse abnormalities. CoMA \cite{kong2026concept} splits each report into concept clauses that describe individual findings and adds a bi-level prototype alignment, wherein learnable cross-modal prototypes act as shared cluster centers to pull together unpaired image and text features at both the instance and concept levels. PANTHER \cite{song2024morphological} instead clusters patch features into fixed morphological prototypes with $k$-means and summarizes each slide with a Gaussian mixture around them, giving a compact slide representation for subtyping and survival prediction.

In image-report retrieval, Gowda et al. \cite{gowda2025prototype} improve medical image-report search by introducing multi-scale prototypes. Instead of comparing an entire X-ray to a full report all at once, their framework extracts representative local and global prototypes on-the-fly directly from the image patches and report sentences. By evaluating how well these visual and textual prototype distributions align using cosine similarity, the model estimates retrieval uncertainty to adaptively re-rank results and prioritize high-confidence reports. Santos et al. \cite{santos2025toward} address the generalization limits of foundation models by training a projection head with a modified prototypical loss, which restructures the embedding space into compact clusters for highly efficient histopathology image retrieval. For unsupervised cross-modal hashing between PET and MRI, UEPA \cite{yao2024unsupervised} derives category prototypes by ensemble clustering and uses them as self-supervised guidance, combined with higher-order similarities between samples, to learn the hash codes.

Bhat et al. \cite{bhat2025exemplar} introduce Exemplar Med-DETR (EM-DETR), integrating prototype learning into object detection for digital mammography. The network extracts visual features from ground-truth lesion regions and aggregates them via an attention-pooling Transformer and an exponential moving average memory bank to form class-specific prototypes. For semi-supervised cervical cancer cell detection, Chai et al. \cite{chai2022deep} align the proposal features of unlabeled images with learnable class proxies and align the class prototypes of labeled and unlabeled data to reduce the effect of noisy pseudo-labels, using a memory bank that keeps labeled prototypes across batches.

For anatomical landmark localization, Xu et al. \cite{xu2026rethinking} adopt a single-prototype-per-class design, learning a deterministic, semantic prototype vector for each target keypoint via end-to-end gradient descent. During iterative inference, the network extracts regional feature patches centered on current estimates and computes a cosine similarity map against the corresponding prototype. These similarity maps serve as attention weights over a fixed offset field, progressively decoding coordinate displacements through a prototype-guided spatial expectation rather than uninterpretable, black-box regression. Lately, Dai et al. \cite{dai2026learning} introduce prototype learning into functional landmarks in ultrasound, using a gated deformable prototype alignment module that registers site-specific anatomical templates into the feature stream so that topological priors guide feature extraction despite acoustic shadowing and speckle noise. 

Xu et al. \cite{xu2025medical} enhance deformable medical image registration by incorporating SAM-guided prototype learning during training. Using segmentation masks from SAM \cite{kirillov2023segment}, the framework applies masked average pooling over encoder feature maps to derive anatomical prototypes for both fixed and moving images. By minimizing cosine distance to pull same-class features together, push different-class features apart, and align matching prototypes across images, the network learns domain-invariant representations. These prototypes function solely as auxiliary regularizers during training and are discarded at inference, enabling the network to directly predict dense deformation fields from the refined feature representations. For renal tumors imaged in two contrast-enhanced CT phases and T1-weighted MRI, Wang et al. \cite{wang2024unsupervised} assign one prototype to each modality to store its style; a prototype loss makes the encoder extract modality-specific style features, the moving image is translated into the modality of the fixed image, and affine and fine registration modules then align the tumor.

Hesse et al. \cite{hesse2024prototype} adapt prototype learning to continuous regression for explainable brain age prediction from neuroimaging data (such as structural MRI and fetal ultrasound). Departing from classification-based prototype networks, their framework (ExPeRT) creates a set of learnable global prototypes with fixed, evenly spaced target labels. During training, these prototypes are updated using gradient descent through a distance metric loss and periodically replaced with the closest real training examples. At test time,

\onecolumn
\begingroup
\scriptsize
\renewcommand{\arraystretch}{1.25}
\setlength{\aboverulesep}{0pt}
\setlength{\belowrulesep}{0pt}
\setlength{\LTcapwidth}{\textwidth}


\setlength{\tabcolsep}{4pt}
\begin{xltabular}{\textwidth}{
  >{\raggedright\arraybackslash\hspace{0pt}}p{2.1cm}
  >{\raggedright\arraybackslash\hspace{0pt}}p{1.5cm}
  >{\raggedright\arraybackslash\hspace{0pt}}p{1.6cm}
  >{\raggedright\arraybackslash\hspace{0pt}}p{2.1cm}
  >{\raggedright\arraybackslash\hspace{0pt}}p{2.7cm}
  >{\raggedright\arraybackslash\hspace{0pt}}X}
\caption{
\textbf{Comprehensive categorization of prototype learning methodologies across medical imaging.} 
Methods are comprehensively cataloged according to their clinical \textbf{Task}, primary imaging \textbf{Modality}, anatomical \textbf{ROI / Target}, and evaluation \textbf{Dataset}. Row shading visually differentiates the primary clinical tasks. The \textbf{Taxonomic Innovation} column deconstructs the core algorithmic contribution of each framework, highlighting how specific prototype attributes (Representation, Learning Strategy, or Utilization) are adapted to overcome targeted clinical constraints. In-house institutional cohorts are denoted where public benchmarks are unavailable. The table lists the 101 core studies.
}
\label{tab:medical_applications} \\
\toprule
\textbf{Ref} & \textbf{Task} & \textbf{Modality} & \textbf{ROI / Target} & \textbf{Dataset} & \textbf{Taxonomic Innovation} \\
\midrule
\endfirsthead
\toprule
\textbf{Ref} & \textbf{Task} & \textbf{Modality} & \textbf{ROI / Target} & \textbf{Dataset} & \textbf{Taxonomic Innovation} \\
\midrule
\endhead
\bottomrule
\endfoot
\bottomrule
\endlastfoot
\rowcolor{taskClassif} Yu et al. \cite{yu2025hierarchical} & Classif. & Dermoscopy & Skin lesion & In-house Molemap dataset & Hierarchical prototypical decision tree routing coarse-to-fine pathological categories \\

\rowcolor{taskClassif} Barnett et al. \cite{barnett2021iaia} & Classif. & Mammography & Malignant breast cancer & CBIS-DDSM \cite{sawyerlee2016cbisddsm}, in-house & Part-based prototypes with top-$k$ pooling aligned with clinical BI-RADS margins \\
\rowcolor{taskClassif} Singh and Yow \cite{singh2021interpretable} & Classif. & X-ray & COVID-19 & COVID-19 Image Data Collection \cite{cohen2020covid} & Square and rectangular region prototypes highlighting class-indicative thoracic patches \\
\rowcolor{taskClassif} Kim et al. \cite{kim2021xprotonet} & Classif. & X-ray & Chest diseases & NIH Chest X-ray \cite{wang2017chestx} & Auxiliary occurrence module that predicts soft spatial weighting maps instead of a fixed-size feature \\
\rowcolor{taskClassif} Djoumessi et al. \cite{djoumessi2024actually} & Classif. & OCT & Drusen lesions & OCT2017 \cite{kermany2018identifying} & Part-based prototypes with top-$k$ pooling \\
\rowcolor{taskClassif} Demir et al. \cite{demir2024explainable} & Classif. & X-ray & Thoracic pathologies & NIH Chest X-ray \cite{wang2017chestx} & Self-attention instead of cosine similarity to compute the distance between prototypes and image features \\
\rowcolor{taskClassif} Wei et al. \cite{wei2024mprotonet} & Classif. & MRI & Glioma sub-regions & BraTS \cite{menze2014multimodal} & 3D multi-parametric prototype learning for volumetric brain tumor classification \\
\rowcolor{taskClassif} Gu et al. \cite{gu2025protoasnet} & Classif. & Echocardiography & Aortic stenosis severity & TMED-2 \cite{huang2022tmed}, in-house & Spatiotemporal video prototypes capturing cyclic motion with prototype-based abstention loss \\
\rowcolor{taskClassif} Gall{\'e}e et al. \cite{gallee2023interpretable} & Classif. & CT & Lung nodules & LIDC-IDRI \cite{armato2011lung} & Attribute-level prototype allocation with privileged clinical guidance \\
\rowcolor{taskClassif} Nauta et al. \cite{nauta2023interpreting} & Classif. & Dermoscopy, X-ray & Skin lesion, musculoskeletal & ISIC 2019 \cite{codella2019skin}, MURA \cite{rajpurkar2017mura}, HIP, ANKLE & Self-supervised prototypes with human-guided pruning capabilities \\
\rowcolor{taskClassif} Huy et al. \cite{huy2025interactive} & Classif. & X-ray, Dermoscopy & Chest diseases, skin lesions & TBX11K \cite{liu2020rethinking}, VinDr-CXR \cite{nguyen2022vindr}, ISIC 2018 \cite{codella2018skin} & ProtoPNet variant that binds prototypes to explicit clinical concepts rather than unconstrained latent patches \\
\rowcolor{taskClassif} Gautam et al. \cite{gautam2022demonstrating} & Classif. & X-ray & Pneumonia detection & CheXpert \cite{irvin2019chexpert}, NIH Chest X-ray \cite{wang2017chestx} & Layer-wise Relevance Propagation in place of ProtoPNet's blurry upsampled heatmaps \\
\rowcolor{taskClassif} Wang et al. \cite{wang2022knowledge} & Classif. & Mammography & Breast cancer detection & ADMANI \cite{frazer2022admani}, CMMD \cite{cui2021cmmd} & Knowledge distillation from a conventional classifier to ProtoPNet to form an ensemble model \\
\rowcolor{taskClassif} Wang et al. \cite{wang2024interpretable} & Classif. & Mammography, OCT, MRI & Breast cancer, retinal disease, brain tumor & ADMANI \cite{frazer2022admani}, CMMD \cite{cui2021cmmd}, NEH \cite{sotoudeh2022multi}, BraTS \cite{menze2014multimodal} & Reciprocal student-teacher co-training between conventional classifier and ProtoPNet \\

\rowcolor{taskClassif} Qu et al. \cite{qu2024rethinking} & Classif. & WSI & Histological carcinoma &  Camelyon16 \cite{ehteshami2017diagnostic}, TCGA-LUAD \cite{cancer2014comprehensive}, TCGA-LUSC \cite{cancer2012comprehensive}, in-house Cervical Cancer & Dynamic prototypes generating pseudo-labels for WSI instance classification \\
\rowcolor{taskClassif} Rymarczyk et al. \cite{rymarczyk2022protomil} & Classif. & Histopathology, WSI & Multi-tissue carcinoma, retina & Bisque Breast Cancer \cite{gelasca2008evaluation}, Colon Cancer \cite{sirinukunwattana2016locality}, Camelyon16 \cite{ehteshami2017diagnostic}, TCGA-NSCLC \cite{bakr2018radiogenomic}, TCGA-RCC \cite{akin2016tcga} & ProtoPNet adapted to WSI classification with attention pooling \\
\rowcolor{taskClassif} Yang et al. \cite{yang2023tpmil} & Classif. & WSI & Brain tumor, lung cancer & TCGA-LUAD \cite{cancer2014comprehensive}, TCGA-LUSC \cite{cancer2012comprehensive}, TCGA Brain Tumor \cite{clark2013cancer} & Trainable prototypes for attention-guided feature regularization and distance-based interpretability heatmaps \\
\rowcolor{taskClassif} Gou et al. \cite{gou2025queryable} & Classif. & Multi-modal (Histopathology + Text) & Cancer & TCGA-NSCLC \cite{bakr2018radiogenomic}, TCGA-RCC \cite{akin2016tcga}, TCGA-BRCA \cite{lingle2016tcga}, TCGA-ESCA \cite{lucchesi2016cancer} & Queryable prototype pool in a vision-language MIL framework for WSI classification \\

\rowcolor{taskClassif} Zhu et al. \cite{zhu2024sfpl} & Classif. & CT, X-ray & Multi-organ diseases & LIDC-IDRI \cite{armato2011lung}, Tuberculosis Diagnosis \cite{liu2020rethinking,jaeger2014two}, in-house EGFR & Uniformly distributed prototypes learned through mixture-weighted, real-time updates \\
\rowcolor{taskClassif} Wang et al. \cite{wang2025cross} & Classif. & X-ray, Fundus & Chest, retina & NIH Chest X-ray \cite{wang2017chestx}, ODIR \cite{li2020benchmark} & ProtoPNet adapted to multi-label medical classification, with cross-image co-attention to disentangle co-occurring disease features \\

\rowcolor{taskClassif} Zhao et al. \cite{zhao2025medfedproto} & Classif. & Ultrasound, Microscopy, Dermoscopy, OCT, Histopathology, X-ray, CT & Fetal heart, blood cells, breast, skin lesions, retina, colon, chest, abdomen & In-house CHD-US, MedMNIST v2 \cite{yang2023medmnist} & FedProto extended to semi-supervised learning via EMA prototype smoothing and multi-loss alignment \\

\rowcolor{taskClassif} Zhang et al. \cite{zhang2024disease} & Classif. & Multi-modal (X-ray, Histopathology + Text) & Cancer and COVID-19 & PanNuke \cite{gamper2019pannuke}, COVIDx \cite{wang2020covid} & Trainable prototypes initialized from clinical text prompts that regularize the VLM latent space during training and are discarded at inference \\
\midrule

\rowcolor{taskSeg} Ouyang et al. \cite{ouyang2020self} & Seg. & CT, MRI & Abdominal organs, cardiac structures & BTCV \cite{landman2015bcv}, CHAOS \cite{kavur2021chaos}, MS-CMRSeg \cite{zhuang2019multivariate} & Self-supervised few-shot learning with superpixel pseudo-labels and adaptive local prototype pooling \\
\rowcolor{taskSeg} Yu et al. \cite{yu2021location} & Seg. & CT & Abdominal organs & VISCERAL \cite{jimenez2016cloud} & Location-sensitive local prototypes from an overlapping support grid, matched only to the query locations each grid cell covers \\
\rowcolor{taskSeg} Zhu et al. \cite{zhu2023partition} & Seg. & CT, MRI & Abdominal organs, cardiac structures & BTCV \cite{landman2015bcv}, CHAOS \cite{kavur2021chaos}, MS-CMRSeg \cite{zhuang2019multivariate} & Regional foreground prototypes with two-way debiasing and prediction aggregation for few-shot segmentation \\
\rowcolor{taskSeg} Zhu et al. \cite{zhu2023few} & Seg. & CT, MRI & Abdominal organs, cardiac structures & BTCV \cite{landman2015bcv}, CHAOS \cite{kavur2021chaos}, MS-CMRSeg \cite{zhuang2019multivariate} & Regional prototype selection and iterative refinement to rectify the global class representation \\
\rowcolor{taskSeg} Zhao et al. \cite{zhao2024dual} & Seg. & MRI & Left and right kidneys & CHAOS \cite{kavur2021chaos} & Self-supervised one-shot segmentation combining local pooling with clustered part prototypes \\
\rowcolor{taskSeg} Tang et al. \cite{tang2025few} & Seg. & CT, MRI & Abdominal organs, cardiac structures & BTCV \cite{landman2015bcv}, CHAOS \cite{kavur2021chaos}, MS-CMRSeg \cite{zhuang2019multivariate} & Few-shot segmentation with channel-wise fusion of clustered foreground prototypes and sparse background regulation \\
\rowcolor{taskSeg} Liang et al. \cite{liang2024advancing} & Seg. & CT, MRI & Abdominal organs, lung, brain & BTCV \cite{landman2015bcv}, Lung \cite{antonelli2022msd}, BraTS \cite{menze2014multimodal} & Fully supervised segmentation using a fusion of features from a learnable class prototype and an instance-specific prototype \\
\rowcolor{taskSeg} Yang et al. \cite{yang2026proto} & Seg. & Ultrasound, X-ray & Breast tumors, thyroid nodules, pneumothorax & BUSI \cite{alDhabyani2020busi}, DDTI \cite{pedraza2015ddti}, SIIM-ACR \cite{siimacr2019pneumothorax} & Supervised segmentation with auxiliary boundary prototypes and foreground/background subclustering to refine the main prototype segmentor \\
\rowcolor{taskSeg} Wang et al. \cite{wang2025edge} & Seg. & CT, MRI & Abdominal organs, cardiac structures & BTCV \cite{landman2015bcv}, CHAOS \cite{kavur2021chaos}, MS-CMRSeg \cite{zhuang2019multivariate} & Few-shot segmentation using erosion-derived inner-boundary prototypes combined with other foreground prototypes for query matching \\
\rowcolor{taskSeg} Jiang and Zhang \cite{jiang2025concentrate} & Seg. & CT, MRI & Abdominal organs, cardiac structures & BTCV \cite{landman2015bcv}, CHAOS \cite{kavur2021chaos}, MS-CMRSeg \cite{zhuang2019multivariate} & Support-error mining of hard foreground/background prototypes, with similarity-map fusion and boundary supervision \\
\rowcolor{taskSeg} Zhang et al. \cite{zhang2024prototype} & Seg. & CT, MRI & Abdominal organs, cardiac structures, prostate & BTCV \cite{landman2015bcv}, CHAOS \cite{kavur2021chaos}, MS-CMRSeg \cite{zhuang2019multivariate}, Prostate-MRI collection \cite{huang2023vector} & Few-shot segmentation combining optimal-transport prototype matching with class-relation reasoning \\
\rowcolor{taskSeg} Huang et al. \cite{huang2024prototype} & Seg. & CT, MRI & Abdominal organs, cardiac structures, pancreas & BTCV \cite{landman2015bcv}, CHAOS \cite{kavur2021chaos}, MS-CMRSeg \cite{zhuang2019multivariate}, MSD \cite{antonelli2022msd} & Asymmetric support-conditioned prototypes that guide query graph refinement before segmentation decoding \\
\rowcolor{taskSeg} Li et al. \cite{li2023prototypical} & Seg. & MRI & Male pelvic structures & Multi-institution pelvic MRI cohort \cite{li2023prototypical} & Cross-institution few-shot segmentation combining spatial registration, support-mask conditioning, and local prototype matching \\
\rowcolor{taskSeg} Wang and Zheng \cite{wang2024pfmnet} & Seg. & Fundus, CT, MRI, Endoscopy & Optic disc/cup, cardiac structures, polyps & REFUGE \cite{orlando2020refuge}, RIM-ONE \cite{fumero2011rimone}, CT--MR cardiac collection \cite{wu2020cfdistance,zhuang2016multiscale}, CVC-ClinicDB \cite{BERNAL201599}, Kvasir-SEG \cite{jha2019kvasir}, CVC-EndoSceneStill \cite{vazquez2017endoluminal} & Few-shot domain adaptation using target prototypes and learned source exemplars to construct a feature mapping for segmentation \\
\rowcolor{taskSeg} Yu et al. \cite{yu2023source} & Seg. & CT, MRI & Abdominal organs & CHAOS \cite{kavur2021chaos}, BTCV \cite{landman2015bcv} & Source-free adaptation aligning target features to frozen prototypes from the source classifier through bidirectional transport and contrastive learning \\
\rowcolor{taskSeg} Wei et al. \cite{wei2024prompting} & Seg. & MRI, Fundus & Prostate, eye fundus & Prostate \cite{liu2020shape}, RIGA+ \cite{almazroa2018retinal,decenciere2014feedback} & Domain-adaptive prototypes from a learnable memory turned into prompts for SAM in single-source domain generalization \\

\rowcolor{taskSeg} Xu et al. \cite{wu2022all} & Seg. & CT, MRI & Kidney, brain tumor & KiTS19 \cite{heller2021kits19}, BraTS \cite{menze2014multimodal} & Semi-supervised learning through forward and backward prototypical consistency between labeled and unlabeled images \\
\rowcolor{taskSeg} Tian et al. \cite{tian2023implicit} & Seg. & Endoscopy, Fundus, Dermoscopy & Polyps, optic discs, skin lesions & Kvasir-SEG \cite{jha2019kvasir}, RIM-ONE \cite{fumero2011rimone}, ISIC 2019 \cite{codella2019skin}, CVC-ColonDB \cite{bernal2012polyp} & Semi-supervised supervision transfer through multiple prototypes, prediction voting, and implicit/explicit alignment \\
\rowcolor{taskSeg} Liu et al. \cite{liu2023prototype} & Seg. & MRI, CT & Brain tumor, left atrium, liver & BraTS \cite{menze2014multimodal}, LA \cite{xiong2021leftatrium}, LiTS \cite{bilic2023liver} & Semi-supervised learning with spatially aware foreground prototypes and voxel-to-prototype contrastive regularization \\
\rowcolor{taskSeg} Liu et al. \cite{liu2026kp2l} & Seg. & MRI, CT & Left atrium, pancreas, brain tumor & LA \cite{xiong2021leftatrium}, Pancreas-CT \cite{roth2016pancreasct}, BraTS \cite{menze2014multimodal} & Semi-supervised transfer of shape, texture, and semantic knowledge through multi-scale prototype prediction consistency \\
\rowcolor{taskSeg} Yuan et al. \cite{yuan2024effective} & Seg. & MRI & Left atrium, cardiac structures, prostate & LA \cite{xiong2021leftatrium}, ACDC \cite{bernard2018deep}, PROMISE12 \cite{litjens2014promise12} & Semi-supervised segmentation with Gaussian pixel/class representations, uncertainty-aware matching, and Bayesian prototype updates \\
\rowcolor{taskSeg} He et al. \cite{he2025evidential} & Seg. & MRI, CT & Left atrium, pancreas, aorta & LA \cite{xiong2021leftatrium}, Pancreas-CT \cite{roth2016pancreasct}, ImageTBAD \cite{yao2021imagetbad} & Semi-supervised learning with evidential prototype fusion and voxel-level uncertainty masking to reduce unreliable supervision \\
\rowcolor{taskSeg} Liu et al. \cite{liu2025procns} & Seg. & OCTA, Fundus, Endoscopy, Histopathology, MRI & Retina, colon, nuclei, heart, brain & sOCTA \cite{wang2021deep}, RIM-ONE \cite{fumero2011rimone}, Kvasir-SEG \cite{jha2019kvasir}, WO \cite{kumar2017dataset}, ACDC \cite{bernard2018deep}, BraTS \cite{menze2014multimodal} & Weakly supervised segmentation with prototype-based regional affinity and noise suppression of sparse-label pseudo-masks \\
\rowcolor{taskSeg} Pan et al. \cite{pan2023human} & Seg. & Histopathology & Tissue types & Camelyon16 \cite{ehteshami2017diagnostic}, BCSS \cite{amgad2019structured} & Multiple regional prototypes from histopathology patches, which are then named by a pathologist and queried to produce dense pseudo-masks \\

\rowcolor{taskSeg} Ge et al. \cite{ge2026progis} & Seg. & Histopathology (ROI, WSI) & Tissue types & BCSS \cite{amgad2019structured}, GCTS \cite{gao2022unsupervised}, Camelyon16 \cite{ehteshami2017diagnostic}, TCGA-LUAD \cite{cancer2014comprehensive}, TCGA-LUSC \cite{cancer2012comprehensive} & Prototype built from a single user scribble to find all tissue regions of the same type in a slide \\

\rowcolor{taskSeg} Liu et al. \cite{liu2026embracing} & Seg. & MRI, CT & Left atrium, pancreas, brain tumor & LA \cite{xiong2021leftatrium}, Pancreas-CT \cite{roth2016pancreasct}, BraTS \cite{menze2014multimodal} & Intensity-aligned heterogeneous prototypes and prototypical-space refinement that transfer intra-class information to a semi-supervised segmentor \\
\rowcolor{taskSeg} Zhu et al. \cite{zhu2025prime} & Seg. & CT & Incrementally added abdominal organs & BTCV \cite{landman2015bcv}, MACT \cite{gibson2018mact} & Class-incremental segmentation using prototype-structure alignment, pixel-level contrastive learning, and consensus-based old-class prototype updates \\
\midrule

\rowcolor{taskSurvival} Liu et al. \cite{liu2025prototype} & Survival Pred. & Multi-modal (WSI, Genomics) & Pan-cancer (breast, bladder, stomach, colorectal) & TCGA (BRCA \cite{lingle2016tcga}, BLCA \cite{kirk2016tcga}, STAD \cite{lucchesi2016tcga}, CRAD) & Modality-specific prototype banks learned via risk contrastive learning for cross-modal translation \\
\rowcolor{taskSurvival} Raza et al. \cite{raza2025ps3} & Survival Pred. & Multi-modal (WSI, Reports, Transcriptomics) & Pan-cancer & Six TCGA datasets \cite{weinstein2013cancer} & On-the-fly prototypes for WSI, report, and genomic features, combined by a fusion module \\
\rowcolor{taskSurvival} Liu et al. \cite{liu2026together} & Survival Pred. & Multi-modal (WSI, Genomics) & Pan-cancer & Multiple TCGA cohorts \cite{weinstein2013cancer} & Shared histology-genomics prototypes assigned by unbalanced optimal transport, balancing alignment and distinctiveness \\

\rowcolor{taskSurvival} Yu et al. \cite{yu2026prime} & Survival Pred. & Multi-modal (WSI, RNA-seq, Reports) & Pan-cancer & TCGA (32 cancer types) \cite{weinstein2013cancer} & Shared learnable prototype memory bank that imputes missing modalities as prototype mixtures during multi-modal pre-training \\

\rowcolor{taskSurvival} Wang et al. \cite{wang2026protosurv} & Survival Pred. & Multi-modal (WSI, Genomics) & Pan-cancer, gastric & Four TCGA datasets \cite{weinstein2013cancer}, in-house gastric cohort & On-the-fly morphological prototypes for each slide, fused with pathway-guided genomic embeddings by gated attention \\

\rowcolor{taskSurvival} Jiang et al. \cite{jiang2026multimodal} & Survival Pred. & Multi-modal (WSI, Genomics) & Pan-cancer & TCGA (LUAD, BLCA, BRCA, UCEC) \cite{weinstein2013cancer} & Class-specific prototypes updated via momentum and matched through distance metrics to guide prediction \\

\rowcolor{taskSurvival} Yang et al. \cite{yang2026protomm} & Survival Pred. & Multi-modal (Pathology, Radiology, Genomics, Demographics) & Brain cancer & TCGA \cite{pedano2016tcga,scarpace2016radiology}, TCIA \cite{clark2013cancer}, BraTS \cite{menze2014multimodal} & Prototypes learned at both the multi-modal and single-modality levels to provide case-based explanations for brain cancer survival \\

\rowcolor{taskSurvival} Kumari et al. \cite{kumari2026protosurvx} & Survival Pred. & Multi-modal (MRI, Patient Age) & High-grade glioma & BraTS 2019/2020 \cite{menze2014multimodal} & Shared memory of learnable prototypes of a combination of radiomic and patient-age features \\

\rowcolor{taskSurvival} Liu et al. \cite{liu2026protosurv} & Prognosis Pred. & CT & Lung cancer & In-house HUAXI (507), LUNG1 \cite{aerts2014data} & Class prototypes pre-computed from a CT foundation model to pseudo-label and calibrate unlabeled scans \\

\rowcolor{taskSurvival} Wang et al. \cite{wang2026impro} & Prognosis Pred. & Longitudinal OCT & Retina & OCT4DME \cite{zhang2026predicting}, GRAPE \cite{huang2023grape}, in-house & Learnable prognosis prototypes that stabilize latent trajectories of a neural controlled differential equation \\

\rowcolor{taskSurvival} Yu et al. \cite{yu2026learning} & Recurrence Pred. & Multi-phase CT & Liver & In-house HCC cohort & Cross-branch alignment of class prototypes that transfers multi-phase knowledge to a single-phase deployment branch \\
\midrule

\rowcolor{taskAnomaly} Li et al. \cite{li2025diverse} & Anomaly Det. & OCT, Fundus, Dermoscopy, MRI & Retina, skin, brain & OCT2017 \cite{kermany2018identifying}, APTOS2019 \cite{karthik2019aptos}, ISIC 2019 \cite{codella2019skin}, in-house BrainMRI & Diverse Normal Prototypes (DNPs) extracted via cross-attention with a DAA loss to guide reconstruction \\
\rowcolor{taskAnomaly} Wu et al. \cite{wu2025anyad} & Anomaly Det. & Multi-sequence MRI & Brain & BraTS \cite{menze2014multimodal}, MU-Glioma-Post \cite{de20242024}, Pretreat-MetsToBrain-Masks \cite{ramakrishnan2024large} & INP-guided decoder with feature distribution alignment to handle missing-modality inputs \\
\rowcolor{taskAnomaly} Wan et al. \cite{wan2026beyond} & Anomaly Det. & MRI, CT, OCT, X-ray, Histopathology & Brain, liver, retina, chest, lymph node & BrainMRI \cite{baid2021rsna}, LiverCT \cite{bilic2023liver}, RESC and OCT2017 \cite{hu2019automated,kermany2018identifying}, NIH Chest X-ray \cite{wang2017chestx}, HIS \cite{ehteshami2017diagnostic} & Normal and abnormal prototypes learned in CLIP visual space and re-centered on each input via bounded gated modulation \\
\rowcolor{taskAnomaly} Xiang et al. \cite{xiang2024exploiting} & Anomaly Det. & X-ray & Medical lesions & ZhangLab Chest X-ray \cite{kermany2018identifying}, CheXpert \cite{irvin2019chexpert}, COVIDx \cite{wang2020covid} & Space-aware normal prototypes learned in a hierarchical memory matrix to guide feature-level in-painting for anatomical consistency \\
\rowcolor{taskAnomaly} Kobayashi and Yamaguchi \cite{kobayashi2024simple} & Anomaly Det. & X-ray & Musculoskeletal & MURA \cite{rajpurkar2017mura} & PIP-Net scoring restricted to anomaly evidence, improving detection with fewer part prototypes \\
\rowcolor{taskAnomaly} Lee et al. \cite{lee2025fewshot} & Anomaly Det. & Ultrasound & Breast & BUSI \cite{alDhabyani2020busi} & Normal and abnormal prototypes from projected BiomedCLIP embeddings for distance-based classification \\
\rowcolor{taskAnomaly} Huang et al. \cite{huang2024uncertainty} & Anomaly Det. & Endoscopy & Colon & Kvasir-SEG \cite{jha2019kvasir}, CVC-ClinicDB \cite{BERNAL201599}, CVC-ColonDB \cite{bernal2012polyp}, ETIS \cite{silva2014toward}, CVC-EndoSceneStill \cite{vazquez2017endoluminal} & Prototype memory bank for diverse anomalous lesions, coupled with Bayesian uncertainty estimation for ambiguous boundaries \\

\rowcolor{taskAnomaly} Pan and Lu \cite{pan2025spmamba} & Anomaly Det. & X-ray, MRI, CT, OCT & Anatomical regions & ZhangLab Chest X-ray \cite{kermany2018identifying}, CheXpert \cite{irvin2019chexpert}, Uni-Medical \cite{he2024mambaad} & Multiple regional prototypes that exploit the structural regularity of radiographs in a Mamba-based detector \\

\rowcolor{taskAnomaly} She et al. \cite{she2024memory} & Anomaly Det. & WSI, CT & Prostate cancer, lung tumor & PANDA-MIL \cite{wu2023contrastive}, in-house LUNG-MIL & Dual memory banks of normal and abnormal patterns with contrastive and triplet feature separation losses \\

\midrule

\rowcolor{taskReportGen} Chen et al. \cite{chen2021cross} & Report Gen. & Multi-modal (X-ray + Text) & Chest & IU X-ray \cite{demner2016preparing}, MIMIC-CXR \cite{johnson2019mimic} & Shared cross-modal memory matrix (prototypes) with top-$k$ retrieval to align visual and textual representations \\
\rowcolor{taskReportGen} Wang et al. \cite{wang2022cross} & Report Gen. & Multi-modal (X-ray + Text) & Chest &  IU X-ray \cite{demner2016preparing}, MIMIC-CXR \cite{johnson2019mimic} & Shared cross-modal prototype matrix queried to condition the report decoder \\
\rowcolor{taskReportGen} Yan et al.~\cite{yan2024buffalo} & Report Gen. & Multi-modal (X-ray, CT, MRI + Text) & Chest and multi-organ & IU X-ray \cite{demner2016preparing}, VQA-RAD \cite{lau2018dataset}, SLAKE \cite{liu2021slake} & Cross-modal prototype imputation to handle missing modalities in federated learning \\
\rowcolor{taskReportGen} Pellegrini et al.~\cite{pellegrini2026prototype} & Structured Report Gen. & Multi-modal (X-ray + Text) & Chest & MIMIC-CXR \cite{johnson2019mimic}, Rad-ReStruct \cite{pellegrini2023rad} & Prototype bank refreshed with an EMA image encoder that adds a residual logit correction to the predictions \\

\rowcolor{taskReportGen} Yang et al. \cite{yang2026clinical} & Report Gen. & Multi-modal (X-ray + Text) & Chest & MIMIC-CXR \cite{johnson2019mimic}, IU X-ray \cite{demner2016preparing} & Hyperbolic finding-state prototypes for long-tailed findings, whose predicted states prompt the report decoder \\

\rowcolor{taskReportGen} Chen et al. \cite{chen2025dia} & Report Gen. & Multi-modal (CT + Text) & Chest & CTRG-Chest-548K \cite{tang2024work} & Learnable normal and abnormal prototypes per disease whose predicted states become diagnostic prompts for a large language model \\
\rowcolor{taskReportGen} Zhang et al. \cite{zhang2024attribute} & Report Gen. & Multi-modal (X-ray + Text) & Chest & MIMIC-CXR \cite{johnson2019mimic} with Chest ImaGenome \cite{wu2021chest}, IU X-ray \cite{demner2016preparing} & Attribute prototypes from word embeddings matched to pairs of anatomical regions, adding attribute nodes to a scene graph that guides the report decoder \\

\midrule

\rowcolor{taskEmerging} Song et al. \cite{song2024morphological} & Rep. Learning & WSI & Cancer subtyping and survival & 13 public WSI datasets & Fixed $k$-means morphological prototypes with a per-slide Gaussian mixture forming a compact, task-agnostic representation \\

\rowcolor{taskEmerging} Wang et al. \cite{wang2022multi} & Rep. Learning & Multi-modal (X-ray + Text) & Chest & MIMIC-CXR \cite{johnson2019mimic}, CheXpert \cite{irvin2019chexpert}, RSNA Pneumonia \cite{shih2019augmenting}, COVIDx \cite{wang2020covid}, Object-CXR, SIIM-ACR \cite{siimacr2019pneumothorax} & Shared bank of learnable cross-modal prototypes for disease-level image--report alignment \\
\rowcolor{taskEmerging} Cheng et al. \cite{cheng2023prior} & Rep. Learning & Multi-modal (X-ray + Text) & Chest & MIMIC-CXR \cite{johnson2019mimic} & Learnable sentence-level prototype memory bank via parallel decoder queries \\
\rowcolor{taskEmerging} Gowda et al. \cite{gowda2025prototype} & Image-Report Retr. & X-ray & Chest, multi-organ & MIMIC-CXR \cite{johnson2019mimic}, RadImageNet \cite{mei2022radimagenet}, NIH Chest X-ray \cite{wang2017chestx}, CheXpert \cite{irvin2019chexpert}, MURA \cite{rajpurkar2017mura} & Multi-scale image and text prototypes computed on-the-fly to estimate match confidence and adaptively re-rank cross-modal retrievals \\
\rowcolor{taskEmerging} Bhat et al. \cite{bhat2025exemplar} & Lesion Detection & Mammography, X-ray & Breast, chest, coronary & VinDr-Mammo \cite{nguyen2023vindr}, CMMD \cite{cui2021cmmd}, VinDr-CXR \cite{nguyen2022vindr}, ARCADE \cite{popov2023arcade} & Single region prototype per class maintained via EMA memory bank and interleaved with text queries for attention-based detection \\
\rowcolor{taskEmerging} Xu et al. \cite{xu2026rethinking} & Landmark Loc. & X-ray, Ultrasound & Cephalometric, fetal head, fetal heart & ISBI 2015 \cite{wang2016benchmark}, IUGC \cite{BAI2026103960}, in-house FHU & Learnable landmark prototypes as semantic references to guide progressive coordinate offset correction \\
\rowcolor{taskEmerging} Xu et al. \cite{xu2025medical} & Registration & CT, MRI & Abdomen, heart & Learn2Reg \cite{hering2022learn2reg}, ACDC \cite{bernard2018deep} & SAM-guided anatomical prototypes from masked average pooling that regularize encoder features through contrastive learning \\
\rowcolor{taskEmerging} Hesse et al. \cite{hesse2024prototype} & Age Pred. & MRI, Ultrasound & Brain & INTERGROWTH-21st \cite{papageorghiou2014international}, IXI \cite{ixidataset} & Global prototypes learned via a distance metric loss and spatial patch-to-patch matching \\

\rowcolor{taskEmerging} Lisazo et al. \cite{lisazo2026learning} & Age Pred. & MRI & Brain & ADNI \cite{jack2008alzheimer}, AIBL \cite{ellis2009australian}, CamCAN \cite{shafto2014cambridge}, MIRIAD \cite{malone2013miriad}, NACC \cite{beekly2007national} & Age-conditioned latent prototypes decoded into brain atlases \\

\rowcolor{taskEmerging} Bazargani et al. \cite{bazargani2025progrespine} & Age Pred. & MRI & Spine & In-house & 3D target-anchored prototypes compared by matching local anatomical patches for spine age estimation \\

\rowcolor{taskEmerging} Ghamary et al. \cite{ghamary2026protoefnet} & Ejection Fraction Pred. & Echocardiography & Left ventricle & EchoNet-Dynamic \cite{ouyang2019echonet} & Ejection fraction estimated from echocardiography videos by comparing spatiotemporal features with learnable prototypes \\
\rowcolor{taskEmerging} Nicolson et al. \cite{nicolson2025human} & Gest. Age Pred. & Ultrasound & Fetal head & INTERGROWTH-21st \cite{papageorghiou2014international}, INTERBIO-21st \cite{kennedy2018deep} & ProtoPNet adapted to binned gestational age with prototypes shared across bins and pruned for sparse explanations \\

\rowcolor{taskEmerging} Liu et al. \cite{liu2026openness} & Open-set Recog. & Histopathology, Microscopy, Dermoscopy & Tissue patches, blood cells, skin lesions & MedMNIST v2 (PathMNIST, BloodMNIST, DermaMNIST) \cite{yang2023medmnist} & Multiple prototypes per class with open-space regularization for open-set recognition \\

\rowcolor{taskEmerging} Cai et al. \cite{cai2026adaptive} & Open-set Recog. & Endoscopy, Dermoscopy, OCT & Gastrointestinal tract, skin lesions, retina & HyperKvasir \cite{borgli2020hyperkvasir}, ISIC 2018 \cite{codella2018skin}, Retinal OCT-C8 \cite{he2023interpretable} & Multiple learnable prototypes for each known class in open-set recognition \\

\rowcolor{taskEmerging} Sun et al. \cite{sun2026sculpt} & OOD Detection & WSI & Cancer tissue & BRACS \cite{brancati2022bracs}, BRACS7, PANDA \cite{bulten2022artificial} & Tissue prototypes from learnable causal prompts for slide-level OOD detection \\

\rowcolor{taskEmerging} Chang et al. \cite{chang2025enhancing} & Image Synthesis & Multi-contrast MRI & Glioma & BraTS 2021 \cite{baid2021rsna} & Contrast-specific prototypes that drive the synthesis of missing MRI contrasts with consistent tumor edges \\

\rowcolor{taskEmerging} Redekop et al. \cite{redekop2025prototype} & Image Synthesis & Histopathology & Cancer & CPIA \cite{ying2025cpia}, PLCO \cite{zhu2013prostate}, PANDA \cite{bulten2022artificial}, in-house & Pre-computed tissue prototypes that condition a latent diffusion model via classifier guidance \\

\rowcolor{taskEmerging} Shandilya et al. \cite{shandilya2024improve} & Image Synthesis & Microscopy, Dermoscopy & Bone marrow, skin lesions & Bone Marrow \cite{matek2021expert}, HAM10000 \cite{tschandl2018ham10000} & Prototype-guided diffusion to improve the medical plausibility of generated images \\

\rowcolor{taskEmerging} Santos et al. \cite{santos2025toward} & Image Retrieval & Histopathology, Dermoscopy & Glomerulus, skin, ovary & HAM10000 \cite{tschandl2018ham10000}, Ovarian Cancer \cite{kasture2021ovarian} & Projection head trained with a modified prototypical loss that restructures the embedding space for image retrieval \\

\rowcolor{taskEmerging} Yao et al. \cite{yao2024unsupervised} & Image Retrieval & PET, MRI & Brain & ADNI2 \cite{jack2008alzheimer}, OASIS3 \cite{lamontagne2019oasis} & Category prototypes from ensemble clustering used as self-supervised guidance \\

\rowcolor{taskEmerging} Wang et al. \cite{wang2025fedpmr} & MR Reconstruction & MRI & Brain, knee & BraTS \cite{menze2014multimodal}, fastMRI \cite{knoll2020fastmri}, IXI \cite{ixidataset} & Federated exchange of global feature prototypes for personalized reconstruction in each site \\

\rowcolor{taskEmerging} Kong et al. \cite{kong2026concept} & Rep. Learning & Multi-modal (X-ray + Text) & Chest & MIMIC-CXR \cite{johnson2019mimic} & Bi-level cross-modal alignment using a shared dictionary of learnable instance and concept prototypes \\

\rowcolor{taskEmerging} Chai et al. \cite{chai2022deep} & Cell Detection & Cytology & Cervical cancer cells & In-house & Memory bank that aligns labeled and unlabeled class prototypes \\

\rowcolor{taskEmerging} Dai et al. \cite{dai2026learning} & Landmark Loc. & Ultrasound, X-ray & Neuro-fascial interfaces, skull & Deqi-US-1k, ISBI 2015 \cite{wang2016benchmark}, ISBI 2023 \cite{khalid2022cepha29} & Class-specific prototypes aligned via a learnable metric to enforce anatomical shape priors \\

\rowcolor{taskEmerging} Wang et al. \cite{wang2024unsupervised} & Registration & CT, MRI & Renal tumors, abdominal organs & In-house, CHAOS \cite{kavur2021chaos} & One style prototype per modality that guides translation of the moving image before affine and fine tumor registration \\

\bottomrule
\end{xltabular}
\endgroup
\twocolumn

\noindent the model calculates patch-level Earth Mover's Distance \cite{rubner2000earth} between the input image and nearby prototypes, predicting the final value through a distance-weighted average of their labels.

Later regression methods follow the same idea. Lisazo et al. \cite{lisazo2026learning} learn age-conditioned prototypes in latent space and decode them into brain atlases, predicting age from the proximity of a scan to these prototypes. Following ExPeRT, ProgreSpine \cite{bazargani2025progrespine} estimates spine age from 3D MRI as a weighted average of the ages of the nearest whole-scan prototypes. ProtoEFNet \cite{ghamary2026protoefnet} estimates ejection fraction from echocardiography videos by comparing spatiotemporal features with learnable prototypes, each labeled with an ejection fraction value. For gestational age estimation from fetal head ultrasound, Nicolson et al. \cite{nicolson2025human} discretize 13--42 weeks into 13 bins of about two weeks and adapt ProtoPNet with shared prototypes across neighboring bins. They prune small weights so that four displayed prototypes cover about 80\% of the model's reasoning on average, and even evaluate the model in a clinician reader study.

For open-set recognition and out-of-distribution (OOD) detection, prototypes describe the known classes, and samples far from all prototypes are rejected. OAMPL \cite{liu2026openness} and PSW-AMPN \cite{cai2026adaptive} represent each known class with multiple prototypes; OAMPL also synthesizes hard unknown features and pushes them away from the known prototypes. Meanwhile, SCULPT \cite{sun2026sculpt} learns tissue prototypes via optimizable causal prompts on pure vision MIL backbones, detecting OOD whole-slide images based on their causal alignment with these prototypes.

For image synthesis and reconstruction, prototypes mainly serve as conditioning signals. CSEC-Net \cite{chang2025enhancing} extracts contrast-specific prototypes to guide the synthesis of missing MRI contrasts. Redekop et al. \cite{redekop2025prototype} pre-compute pathology patch embeddings to form prototypes that condition a latent diffusion model, and IMPROVE \cite{shandilya2024improve} conditions a diffusion model using class prototypes learned in a separate stage. For accelerated MR reconstruction, FedPMR \cite{wang2025fedpmr} shares feature prototypes instead of model weights in personalized federated learning.

\noindent\textbf{Summary.} 
Finally, the other clinical tasks ($n = 25$) show that prototype learning extends to representation learning and pre-training, retrieval, detection, localization, registration, regression, open-set recognition, image synthesis, and reconstruction. Implicit prototypes are the most common choice here ($11/25$), and eight studies discard their prototypes after training, whereas the regression studies provide case-based ($3/5$) or part-based ($2/5$) explanations (Fig.~\ref{fig:quantitative_across_task}).

\section{Practical Guidelines for Prototype Learning in Medical Imaging}
\label{sec:practical_guidelines}
\begin{table*}[!b]
\centering
\footnotesize
\renewcommand{\arraystretch}{1.22}
\setlength{\tabcolsep}{4pt}
\newcommand{\guiderule}{\arrayrulecolor{black!40}\specialrule{0.3pt}{1pt}{1pt}\arrayrulecolor{black}}

\caption{\textbf{Choosing a prototype design.} Recurring needs in the core studies, the tasks in which they typically arise, the R-L-U components that provide each ability, example studies, and the main limitations. For brevity, the design column names only the components that drive the ability; the remaining components vary across studies. The table reflects how the core studies use each design, not head-to-head comparisons.}
\label{tab:design_guidance}
\begin{tabularx}{\textwidth}{@{}>{\raggedright\arraybackslash}p{2.6cm} >{\raggedright\arraybackslash}p{3.6cm} >{\raggedright\arraybackslash}p{2.7cm} >{\raggedright\arraybackslash}p{2.35cm} >{\raggedright\arraybackslash}X@{}}
\toprule
\textbf{Need} & \textbf{Typical tasks} & \textbf{Suitable design choices} & \textbf{Example studies} & \textbf{Main limitations} \\
\midrule
Explaining predictions to clinicians & Classification (Section~\ref{sec:image_classification}); age regression (Section~\ref{sec:emerging_frontiers}) &
Multiple + region-level or global + active prototypes &
\citeleft\citen{barnett2021iaia,singh2021interpretable,kim2021xprotonet,wei2024mprotonet,gu2025protoasnet,nauta2023interpreting,gallee2023interpretable,hesse2024prototype,bazargani2025progrespine,nicolson2025human}\citeright &
Explanations still need faithfulness tests and reader studies (Section~\ref{sec:evaluating_models}), and prototypes can learn shortcuts (Section~\ref{subsec:aligning_clinical_knowledge}); 2D patch designs do not carry over directly to 3D or WSI (Section~\ref{subsec:scaling_volumetric_gigapixel}) \\ \guiderule
Recognizing new classes from a few examples & Few-shot classification (Section~\ref{sec:image_classification}), segmentation (Section~\ref{sec:image_segmentation}), and anomaly detection (Section~\ref{sec:anomaly_detection}) &
On-the-fly + distance-based prototypes &
\citeleft\citen{snell2017prototypical,ouyang2020self,yu2021location,zhu2023partition,zhu2023few,zhao2024dual,tang2025few,jiang2025concentrate,wang2025edge,li2023prototypical,lee2025fewshot}\citeright &
The prototypes stay implicit, so they give no explanation; dense matching is costly in 3D (Section~\ref{subsec:scaling_volumetric_gigapixel}) \\ \guiderule
Learning from few labeled images & Semi- and weakly supervised segmentation (Section~\ref{sec:image_segmentation}), classification (Section~\ref{sec:image_classification}), and prognosis (Section~\ref{sec:survival_prediction}) &
Class-specific + distance-based prototypes &
\citeleft\citen{wu2022all,tian2023implicit,liu2023prototype,liu2026kp2l,he2025evidential,liu2026embracing,liu2025procns,qu2024rethinking,zhao2025medfedproto,liu2026protosurv}\citeright &
Most studies discard the prototypes after training, so they give no explanation \\ \guiderule
Detecting abnormalities without lesion labels & Anomaly detection (Section~\ref{sec:anomaly_detection}) &
Shared dictionary + generative + active prototypes &
\citeleft\citen{xiang2024exploiting,li2025diverse,wu2025anyad,pan2025spmamba}\citeright &
Prototypes can collapse onto a few dominant patterns; prototypes built from the test image can absorb abnormal tissue (Section~\ref{subsec:ood_handling}) \\ \guiderule
Handling whole-slide images & Slide classification (Section~\ref{sec:image_classification}); survival prediction (Section~\ref{sec:survival_prediction}) &
Multiple + region-level or global prototypes &
\citeleft\citen{rymarczyk2022protomil,song2024morphological,wang2026protosurv,raza2025ps3,liu2026together,sun2026sculpt,liu2025prototype,jiang2026multimodal}\citeright &
Comparing every patch with every prototype is costly at this scale (Section~\ref{subsec:scaling_volumetric_gigapixel}); most survival models keep the prototypes implicit \\ \guiderule
Combining images with other modalities & Multi-modal survival prediction (Section~\ref{sec:survival_prediction}), report generation (Section~\ref{sec:report_generation}), and vision-language pre-training (Section~\ref{sec:emerging_frontiers}) &
Shared dictionary prototypes &
\citeleft\citen{liu2026together,yu2026prime,yang2026protomm,gou2025queryable,chen2021cross,wang2022cross,cheng2023prior,kong2026concept,wang2022multi}\citeright &
Mostly implicit, so the prototypes do not show which evidence drove a prediction or a sentence \\ \guiderule
Rejecting inputs unlike the training data & Open-set recognition and OOD detection (Section~\ref{sec:emerging_frontiers}) &
Multiple + class-specific + active prototypes &
\citeleft\citen{liu2026openness,cai2026adaptive,sun2026sculpt}\citeright &
Only four core studies reject unknown inputs, and only one of them gives part-based explanations (Section~\ref{subsec:ood_handling}) \\ \guiderule
Training across hospitals without sharing data & Federated classification (Section~\ref{sec:image_classification}), report generation (Section~\ref{sec:report_generation}), and MR reconstruction (Section~\ref{sec:emerging_frontiers}) &
Global prototypes &
\citeleft\citen{zhao2025medfedproto,yan2024buffalo,wang2025fedpmr}\citeright &
Shared prototypes may still reveal patient information, so studies should test for such leaks and state how they prevent them (Section~\ref{subsec:preserving_patient_privacy}) \\ \guiderule

Building on a foundation model & Report generation (Section~\ref{sec:report_generation}), anomaly detection (Section~\ref{sec:anomaly_detection}), and segmentation (Section~\ref{sec:image_segmentation}) &
Global prototypes &
\citeleft\citen{wei2024prompting,lee2025fewshot,gou2025queryable,liu2026protosurv,chen2025dia}\citeright &
The prototypes inherit the model's notion of similarity and its biases; none of these studies exposes part- or case-based explanations yet \\
\bottomrule
\end{tabularx}
\end{table*}

Sections~\ref{sec:evolution_of_prototype_learning} and~\ref{sec:medical_image_applications} describe how prototype learning has been designed and applied in medical imaging. In this section, we turn these findings into practical guidance: Section~\ref{sec:choosing_design} helps readers choose a prototype design for a given need, and Section~\ref{sec:evaluating_models} discusses how to evaluate a prototype model and its explanations.

\subsection{Choosing a Prototype Design}
\label{sec:choosing_design}
In this section, we point out the main components that drive each recurring need, so that readers can quickly choose a practical prototype design for their own task. Table~\ref{tab:design_guidance} summarizes these needs together with the tasks in which they typically arise, example studies, and their main limitations. We organize the table by need rather than by task, because the same need, such as learning from few labeled images, often appears in several tasks. Note that the table shows what each component provides, not which design performs best, since results from different papers are not directly comparable (Section~\ref{sec:evaluating_models}). Below, we briefly explain suitable choices for each need.

\textbf{Explaining predictions to clinicians.} Since each prototype can be shown as only one image region or one whole image, a model needs multiple prototypes to cover the different patterns of a class. If we want to show prototypes as image parts, they should be region-level, which gives part-based explanations. Meanwhile, if we want to show whole training cases instead, they should be global prototype, which gives case-based explanations. In both cases, the prototypes must stay active at inference, because we can only explain a prediction with the prototypes that actually produced it.

\textbf{Recognizing new classes from a few examples.} Here, the prototypes are built on the fly from the few labeled support images, so a new class can be defined at test time. Because each query is then matched to these prototypes by distance, the model does not need retraining or new class weights for classes it has never seen.

\textbf{Learning from few labeled images.} Here, we compute one or more prototypes per class from the labeled images, so each class has its own reference. Unlabeled pixels or images are then assigned to the class of their closest prototype, which gives us pseudo-labels or consistency targets without extra annotation.

\textbf{Detecting abnormalities without lesion labels.} When no lesion labels are available, the prototypes can only describe normal anatomy. They are therefore kept in a shared dictionary of normal patterns rather than in one set per class, and the model is trained with a generative objective to reconstruct normal images from them. At inference, the prototypes stay active, so regions that cannot be reconstructed from normal patterns stand out as abnormal.

\textbf{Handling whole-slide images.} A whole-slide image contains thousands of patches but far fewer distinct tissue patterns. Region-level prototypes capture these patterns among the patches, so a slide can be represented by a handful of prototypes instead of by every patch. Meanwhile, global prototypes stand for an entire slide or patient, such as the typical case of a risk group, and are compared with a slide after its patches have been pooled. Either way, the model needs multiple prototypes to cover the variety of tissues or patients.

\textbf{Combining images with other modalities.} Because a shared dictionary is not tied to any class, its prototypes can stand for patterns that several modalities describe, such as a finding and the sentence that reports it. Image patches, genomic data, and report phrases can then be compared with the same prototypes and fused later, and a missing modality can even be estimated from the prototypes it would have matched.

\textbf{Rejecting inputs unlike the training data.} If multiple prototypes per class remain active at inference, how close an input lies to its nearest prototype tells us how familiar it is, and inputs that are far from all prototypes can be rejected. Having several prototypes per class also helps avoid rejecting valid cases from classes whose appearance varies.

\textbf{Training across hospitals without sharing data.} When it comes to federated learning, a global prototype can summarize many images in one compact vector. Hospitals can therefore exchange and aggregate prototypes instead of sharing patient images or full model weights.

\textbf{Building on a foundation model.} The foundation model usually stays frozen, and the prototypes are the light part that adapts it: they are either computed from its embeddings of a few labeled cases, such as class means, or learned as a few parameters. Because models such as CLIP describe a whole image or text prompt with one embedding, and SAM is guided by prompt embeddings, these prototypes are mostly global. They then serve as prompts for the foundation model, as a light classification head, or as references for pseudo-labeling.

\subsection{Evaluating Prototype Models}
\label{sec:evaluating_models}
Prototype models are hard to compare across papers. A new model usually differs from earlier ones not only in its prototype design but also in its backbone, pretraining, and training data, so a gain reported over another method cannot be attributed to the prototype design alone. We therefore suggest that each new model report its R-L-U coding (as in Table~\ref{tab:appendix_rlu_coding}) and be compared with the same backbone without the prototype layer, using the same pretraining, data splits, and training budget, so that what the prototypes add or cost is isolated. Design choices should be changed one component at a time (e.g., class-specific versus shared prototypes, or a fixed distance versus a learnable metric), and results should include the number of prototypes, the computation and memory cost, and the variation across random seeds or folds.

\begin{figure}[!t]
    \centering
    \includegraphics[width=\columnwidth]{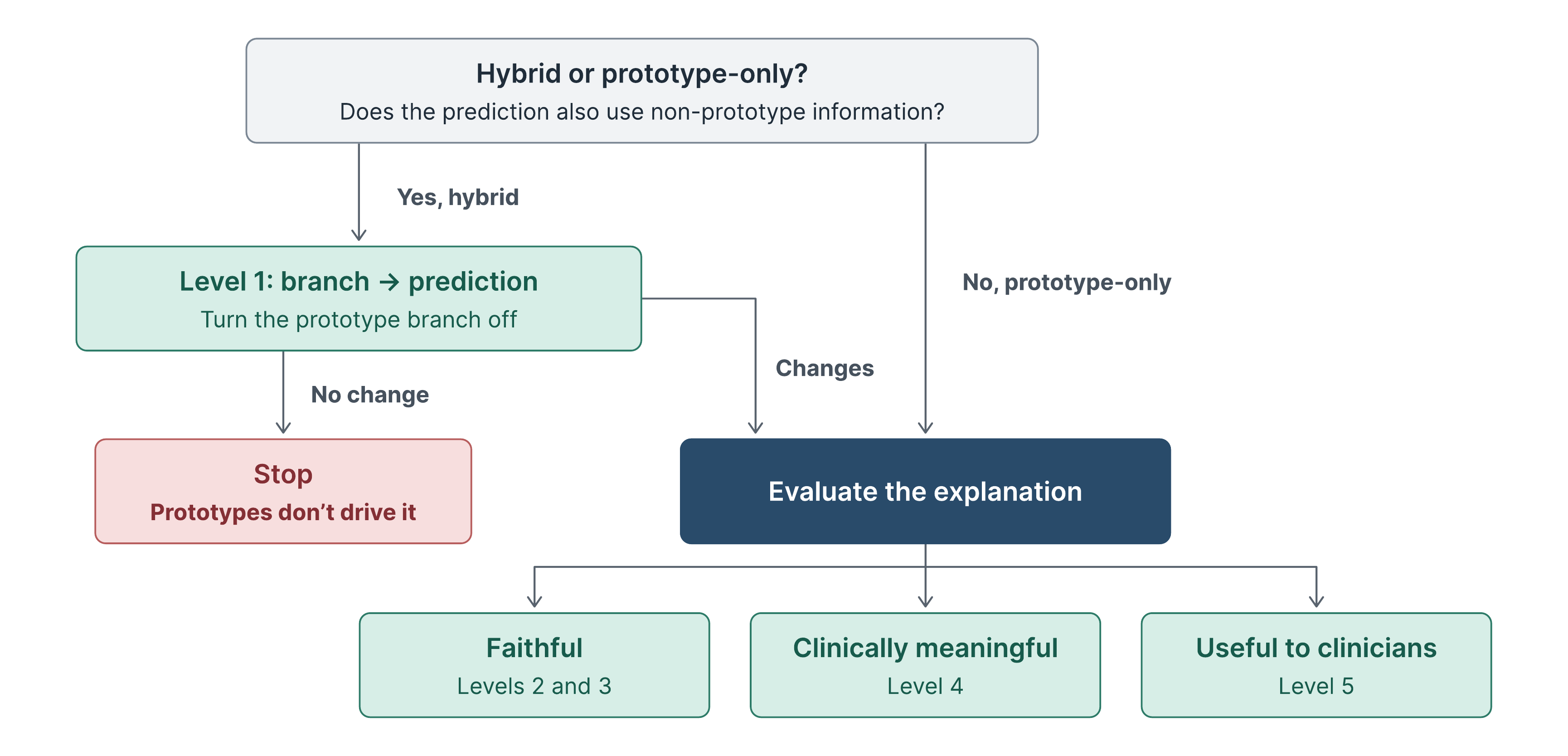}
    \caption{
    \textbf{Five-level evaluation of prototype explanations.}
    Level~1 applies only to hybrid models, whose prediction also uses non-prototype information: if turning the prototype branch off does not change the prediction, the prototypes do not drive it. Otherwise, and for prototype-only models, levels~2 and~3 test whether the explanation is faithful, level~4 whether it is clinically meaningful, and level~5 whether it is useful to clinicians.
    }
    \label{fig:prototype_evaluation}
\end{figure}

\textbf{Explanations also need their own evaluation.} As discussed in Section~\ref{sec:explainability_of_prototypes}, a prototype that can be displayed is not necessarily faithful, meaning that it shows what the model actually uses, and it may not be clinically meaningful or useful to clinicians either. Such evidence is still rare. Of the 26 core studies with part-based (20) or case-based (6) explanations (Section~\ref{sec:medical_image_applications}), only four studies \cite{wei2024mprotonet,djoumessi2024actually,gautam2022demonstrating,nauta2023interpreting} test whether their explanations are faithful, while the rest only show a few hand-picked images, check whether the heatmaps overlap lesions marked by experts (e.g., Barnett et al.~\cite{barnett2021iaia}), or report scores of the prototypes themselves. Faithfulness tests are also harder to read for hybrid models: when the prediction combines both prototype and non-prototype information, a change in the prediction may come from either branch.

Only two studies test the explanations with people. Huy et al. \cite{huy2025interactive} report a user study with one medical student and one non-expert, and Nicolson et al. \cite{nicolson2025human} run the only reader study with clinicians: ten clinicians estimated gestational age from fetal ultrasound on their own, then with the predictions of a ProtoPNet-based model, and then with its explanations as well. The predictions reduced their mean absolute error from 23.5 to 15.7 days, but the explanations gave no significant further gain. They raised the clinicians' confidence without significantly changing their trust in the model or their reliance on it, and some clinicians did worse with them than without. Such effects only show up when people use the model.

We therefore recommend evaluating prototype explanations at five levels (Fig.~\ref{fig:prototype_evaluation}), each answering one question:
\begin{enumerate}[leftmargin=*]
\item \textbf{Branch $\rightarrow$ prediction:} \emph{If the prediction is made by combining both prototype and non-prototype information (as in~\cite{wang2022knowledge}), how much does it change when the prototype branch is turned off?}
\item \textbf{Region $\rightarrow$ prototype:} \emph{When the highlighted region is removed, does the prototype's similarity decrease, showing that the prototype actually looks at that region?}
\item \textbf{Prototype $\rightarrow$ prediction:} \emph{When the prototype is removed, does the prediction change, showing that the prototype really drives the answer?}
\item \textbf{Prototype $\rightarrow$ clinical concept:} \emph{When experts check the prototype against their annotations or rate it, is it clinically meaningful or just a random pattern?}
\item \textbf{Explanation $\rightarrow$ clinician:} \emph{When clinicians use the explanations in a reader study, do they make better decisions than with the predictions alone?}
\end{enumerate}
Level~1 shows how much of the prediction comes from the prototypes rather than from other information, which matters only for hybrid models. Levels~2 and~3 test whether an explanation is faithful, level~4 whether it is clinically meaningful, and level~5 whether it is useful to clinicians. Passing one level does not imply the others: a prototype can match its region without driving the prediction, and a faithful prototype can still lack clinical meaning or fail to help clinicians. Of the four studies that test faithfulness, Gautam et al.~\cite{gautam2022demonstrating} replace the pixels a prototype's heatmap marks as unimportant and check its similarity (level~2), while Wei et al.~\cite{wei2024mprotonet} remove the most important voxels first, Djoumessi et al.~\cite{djoumessi2024actually} keep only the patches most similar to the prototypes, and Nauta et al.~\cite{nauta2023interpreting} add artifacts to images and turn off prototypes, each checking how the prediction changes (level~3). Barnett et al.~\cite{barnett2021iaia} check whether the heatmaps overlap lesions marked by experts (partly level~4), which shows that a prototype looks at the right place but not that it has the right clinical meaning. Only Nicolson et al.~\cite{nicolson2025human} compare clinicians' decisions with and without the explanations (level~5). \textbf{No study answers all five questions}, and none covers even levels~2 and~3 together. Within its resources, a study should cover as many levels as it can and claim only what those levels support.

\section{Open Challenges and Future Directions}
\label{sec:challenges_future_directions}

While prototype learning has demonstrated substantial potential in medical image analysis by offering interpretability, data efficiency, and robustness across a diverse spectrum of representational forms, learning strategies, and utilization mechanisms, the field remains in an active phase of evolution. Several fundamental challenges regarding computational efficiency, semantic validity, and clinical deployment remain unresolved. In this section, we discuss these core challenges and provide insights into future research directions.

\subsection{Aligning Visual Prototypes With Clinical Knowledge}
\label{subsec:aligning_clinical_knowledge}
A fundamental assumption of prototype learning is that mathematical proximity in a feature space corresponds to biological similarity. However, explicit spatial prototypes do not inherently guarantee clinical validity \cite{nauta2023anecdotal,nauta2023co12,hoffmann2021looks,pathak2024prototype}. Due to the phenomenon of shortcut learning, a network might construct a visual patch prototype that mathematically minimizes the training loss but inadvertently focuses on confounding features such as radiographic text markers, image edges, and colored skin patch artifacts rather than true pathology \cite{barnett2021iaia,gautam2023looks,nauta2023interpreting}. To resolve this problem, current architectures support direct human guidance by constraining prototypes with expert lesion masks \cite{barnett2021iaia}, manually inspecting and disabling spurious prototypes like PIP-Net \cite{nauta2023interpreting}, or explicitly steering prototypes toward verified medical concepts via interactive feedback \cite{huy2025interactive}.

These fixes, however, depend on fine-grained annotations or manual effort from experts, which is costly. Whether prototypes can be kept away from shortcuts with weaker supervision, such as report text or a few expert masks, is still an open question. It can be tested by training on data with a known confounder (e.g., an injected marker) and measuring how often prototypes fire on it, together with accuracy once the confounder is removed. Such a test reveals reliance on the confounder, but not whether the explanations are faithful to the model or useful to clinicians.

\subsection{Extending Interpretability Beyond Disease Classification}
\label{subsec:accuracy_interpretability_gap}

A persistent critique of inherently interpretable models is the perceived trade-off between transparency and predictive accuracy, which has been widely debated in different fields \cite{rudin2019stop,van2022explainable}. Several studies report that prototype models can match black-box performance on their benchmarks, but this comes at the cost of higher architectural and training complexity \cite{wang2022knowledge, wang2024interpretable}.

Furthermore, explicit explanations remain concentrated in classification: $17$ of the $22$ classification studies provide part- or case-based explanations, compared with only $9$ of the $52$ core studies in other tasks and none of the $27$ segmentation studies (Fig.~\ref{fig:quantitative_across_task}), so how to increase the interpretable aspect for those tasks is still an open space to explore. An open question is whether part- or case-based explanations can be added to segmentation, survival prediction, and anomaly detection without loss of task performance. Answering it requires comparing interpretable and implicit variants with the same backbone and supervision, and evaluating the explanations with faithfulness tests and reader studies.

\subsection{Scaling From 2D Patches to Volumetric and Gigapixel Scans}
\label{subsec:scaling_volumetric_gigapixel}

While prototype learning has been widely adopted for 3D volumetric segmentation, only a sparse subset of studies has successfully extended explicitly interpretable, part-based prototype architectures to 3D modalities and gigapixel WSIs \cite{wei2024mprotonet, rymarczyk2022protomil, yang2023tpmil}. The number of prototypes involves a trade-off between coverage, compactness, semantic consistency, and usability: more prototypes can cover more diverse disease patterns, but they also add redundancy and make explanations harder for clinicians to read (Section~\ref{sec:explainability_of_prototypes}). In 3D volumes and WSIs, the computational cost of comparing every prototype with every patch or voxel also grows with the number of prototypes; this cost is distinct from memory use, which depends on the implementation (e.g., batching or caching of similarity maps). To limit the cost, current methods restrict the prototype pool or subsample patches and voxels, at the risk of discarding clinically relevant features.

To reduce this cost without losing interpretability, future prototype models could decouple representation learning from dense pixel-to-prototype matching, for example through prototype-guided token pruning, which discards redundant background tokens before similarity computation. An open question is how few prototypes suffice for a given 3D or WSI task without losing coverage. This can be studied by varying the number of prototypes and reporting task performance, prototype redundancy, computation and memory cost, and clinician ratings of explanation usability.

\subsection{Reconciling Federated Patient Privacy With Visual Interpretability}
\label{subsec:preserving_patient_privacy}

A major challenge in medical imaging is that prototypes learned from a single hospital often fail to generalize to new hospitals due to variations in scanners, patient demographics, and imaging protocols \cite{yu2023source, wei2024prompting}. While strict medical privacy laws prohibit the pooling and sharing of raw patient images \cite{tan2022fedproto}, recent federated prototype learning works \cite{zhao2025medfedproto, yan2024buffalo, wang2025fedpmr} address this by exchanging learned prototypes instead of raw images or full model weights. Keeping images local, however, does not by itself guarantee privacy since shared prototypes summarize patient features and may still leak information, so the threat model and the protection mechanism (e.g., differential privacy or secure aggregation) need to be stated and tested.

Moreover, these federated prototype works all use global prototypes without prototype-based explanations \cite{zhao2025medfedproto, yan2024buffalo, wang2025fedpmr}. An open question is whether visually interpretable prototypes can be shared across sites without revealing patient-identifiable content. This can be evaluated with membership-inference or reconstruction attacks on the shared prototypes, reported together with explanation quality and accuracy at each site.



\subsection{Handling Out-of-Distribution Inputs}
\label{subsec:ood_handling}
Prototype learning assumes that a new test case will look like at least one of the stored prototypes. In practice this often fails: a rare pathology, an imaging artifact, or a different scanner can produce an image that resembles none of them, and a static model then either forces a match to an unrelated prototype or gives low similarity to every class. Some recent methods respond by building prototypes on-the-fly from the test image itself \cite{li2025diverse,wu2025anyad} or by conditioning learned prototypes on it \cite{wan2026beyond}.

Most core studies are not built for this. Almost all of them (98 of 101) use deterministic prototypes (Fig.~\ref{fig:quantitative_rlu}b), so the distance to a prototype says nothing about how much the class varies around it, and even the three probabilistic ones do not by themselves give calibrated uncertainty. Twenty-eight studies compute prototypes on-the-fly from the current support set or batch, but adapting them to the test image is mostly limited to implicit anomaly detection models such as those above, and it can backfire: when the references come from the test image itself, abnormal tissue can be absorbed into the normal prototypes and hide the anomaly. Only four core studies explicitly detect or reject unknown inputs \cite{liu2026openness, cai2026adaptive, sun2026sculpt, gu2025protoasnet}. Even testing on an external cohort is uncommon since among the 26 core studies with part- or case-based explanations, only Wang et al. \cite{wang2022knowledge}, PANTHER \cite{song2024morphological}, Lisazo et al. \cite{lisazo2026learning}, and Nicolson et al. \cite{nicolson2025human} report results on data from a source not used in training. It therefore remains open whether interpretable prototype models can recognize out-of-distribution inputs and abstain on them without losing the quality of their explanations. Answering this needs risk-coverage curves and calibration error under controlled scanner, site, and pathology shifts, together with a check of how stable the explanations stay on the shifted data.

\subsection{Prototypes and Foundation Models}
\label{subsec:foundation_models}
Several core studies build prototypes on top of foundation models such as SAM, CLIP, and large language models instead of training an encoder from scratch, using them to adapt or prompt these models or to pass knowledge between models and sites \cite{wei2024prompting,xu2025medical,wan2026beyond,lee2025fewshot,gou2025queryable,liu2026protosurv,yan2024buffalo,chen2025dia,pellegrini2026prototype}.

Here the prototype is a light, data-efficient interface to a model that is otherwise hard to adapt or inspect, yet none of these studies exposes part- or case-based explanations. These prototypes also inherit the foundation model's notion of similarity, which may differ from a clinician's \cite{hoffmann2021looks}, along with its biases. Future work could make prototypes the explanation layer of foundation models, test them with the faithfulness and reader studies discussed in Section~\ref{sec:evaluating_models}, and check what patient information shared prototypes reveal (Section~\ref{subsec:preserving_patient_privacy}).

\section{Conclusion}
\label{sec:conclusion}

Over the past decade, deep learning in medical imaging has largely prioritized accuracy at the cost of clinical transparency, making model decisions difficult for clinicians to audit or verify. This review shows that prototype learning offers one route toward reconciling high-capacity representation learning with inspectable reasoning by relating decisions to explicit reference representations that, when grounded in real cases or image regions, can be displayed and checked.

Deconstructing the field through our unified Representation, Learning Strategy, and Utilization (R-L-U) taxonomy reveals a decisive evolution: prototype learning is no longer merely a heuristic for scarce annotations but a foundational architectural principle. Its footprint now spans vital clinical challenges, mitigating prohibitive labeling costs in dense segmentation, bridging multi-modal report generation, and harmonizing cross-scanner and institutional domain shifts across federated networks. Across these settings, its enduring strength rests on explicit structural grounding: transforming the unconstrained latent space into either inspectable visual evidence for clinical diagnosis or stable semantic anchors for data-efficient and cross-institutional learning.

Ultimately, as medical AI transitions from closed benchmark evaluations to safety-critical deployment across heterogeneous patient populations, post hoc explanations and black-box predictions are no longer sufficient. Prototype learning can contribute to this transition by building explicit reference representations into the architecture. However, 75 of the 101 core studies expose no part- or case-based explanation, and displayed prototypes still need to be shown to be faithful and clinically useful. Architectural transparency should therefore not be equated with validated clinical interpretability. We anticipate that this survey will serve as a foundational blueprint, guiding the development of the next generation of steerable, data-efficient, and trustworthy medical foundation models.

\section*{Acknowledgments}
This study is partially supported by NIH U01-CA268808 (HITPIRADS), R01-HL171376 (PASC) and R21-HL186024 (SPARTACUS). The authors used Claude (Anthropic) to improve the grammar and wording of the manuscript and to write the code that plotted Figs.~1, 4, 11 and 12 from the authors' data. The authors take full responsibility for this article and declare no competing interests.


\bibliographystyle{IEEEtran}

\bibliography{bibliography/00_intro,bibliography/01_survey_methodology,bibliography/01_background, bibliography/02_prototype_foundation, bibliography/03_application_in_medical, bibliography/04_dataset_citation}

\onecolumn
\appendices
\section{Taxonomic Coding of the Medical Prototype Learning Corpus}
\label{app:rlu_coding}

\setcounter{table}{0}
\renewcommand{\thetable}{A\arabic{table}}

\noindent Following the taxonomic definitions established in Section~\ref{sec:core_components}, Table~\ref{tab:appendix_rlu_coding} categorizes all 101 core studies from Table~\ref{tab:medical_applications} according to their primary \textbf{Clinical Abstraction} (Section~\ref{sec:clinical_abstraction_of_prototypes}) and the three dimensions of our R-L-U framework: \textbf{Representation} (Section~\ref{sec:innovation_prototype_representation}), \textbf{Learning Strategy} (Section~\ref{sec:innovation_in_prototype_learning_strategies}), and \textbf{Utilization} (Section~\ref{sec:prototype_utilization}). For multi-stage or hybrid architectures, each dimension is coded according to how the prototype parameters are primarily structured, updated during the main training loop, and exposed at inference.

\begingroup
\scriptsize
\renewcommand{\arraystretch}{1.22}
\setlength{\aboverulesep}{0pt}
\setlength{\belowrulesep}{0pt}
\setlength{\LTcapwidth}{\textwidth}

\begin{xltabular}{\textwidth}{>{\raggedright\arraybackslash\hspace{0pt}}p{2.6cm} >{\raggedright\arraybackslash\hspace{0pt}}p{1.4cm} >{\raggedright\arraybackslash\hspace{0pt}}p{2.1cm} >{\raggedright\arraybackslash\hspace{0pt}}X >{\raggedright\arraybackslash\hspace{0pt}}X >{\raggedright\arraybackslash\hspace{0pt}}X}
\caption{\textbf{Complete R-L-U taxonomic coding of prototype learning methodologies in medical imaging.} Abbreviations: \textit{Deter.} = Deterministic; \textit{Prob.} = Probabilistic; \textit{Single} = Single Prototype; \textit{Multi} = Multiple Prototypes; \textit{CS} = Class-Specific; \textit{SD} = Shared Dictionary; \textit{No expl.} = No prototype-based explanation; \textit{No obj.} = No training objective (frozen prototypes).}
\label{tab:appendix_rlu_coding} \\
\toprule
\textbf{Ref} & \textbf{Task} & \textbf{Clinical Abstraction} & \textbf{Representation (R)} \newline \textit{(Form -- Granularity -- Card. -- Class assoc.)} & \textbf{Learning Strategy (L)} \newline \textit{(Objective -- Optimization)} & \textbf{Utilization (U)} \newline \textit{(Role -- Comparison -- Interpret.)} \\
\midrule
\endfirsthead

\toprule
\textbf{Ref} & \textbf{Task} & \textbf{Clinical Abstraction} & \textbf{Representation (R)} \newline \textit{(Form -- Granularity -- Card. -- Class assoc.)} & \textbf{Learning Strategy (L)} \newline \textit{(Objective -- Optimization)} & \textbf{Utilization (U)} \newline \textit{(Role -- Comparison -- Interpret.)} \\
\midrule
\endhead

\bottomrule
\endfoot
\bottomrule
\endlastfoot

\rowcolor{taskClassif} Yu et al. \cite{yu2025hierarchical} & Classif. & Class Identity & Deter. -- Global -- Single -- CS & Discriminative -- Gradient-based & Active -- Distance-based -- Case-based \\
\rowcolor{taskClassif} Barnett et al. \cite{barnett2021iaia} & Classif. & Appearance & Deter. -- Region -- Multi -- CS & Discriminative -- Gradient-based & Active -- Distance-based -- Part-based \\
\rowcolor{taskClassif} Singh and Yow \cite{singh2021interpretable} & Classif. & Appearance & Deter. -- Region -- Multi -- CS & Discriminative -- Gradient-based & Active -- Distance-based -- Part-based \\
\rowcolor{taskClassif} Kim et al. \cite{kim2021xprotonet} & Classif. & Appearance & Deter. -- Region -- Multi -- CS & Discriminative -- Gradient-based & Active -- Distance-based -- Part-based \\
\rowcolor{taskClassif} Djoumessi et al. \cite{djoumessi2024actually} & Classif. & Appearance & Deter. -- Region -- Multi -- CS & Discriminative -- Gradient-based & Active -- Distance-based -- Part-based \\
\rowcolor{taskClassif} Demir et al. \cite{demir2024explainable} & Classif. & Appearance & Deter. -- Region -- Multi -- SD & Discriminative -- Gradient-based & Active -- Learnable metric -- Part-based \\
\rowcolor{taskClassif} Wei et al. \cite{wei2024mprotonet} & Classif. & Appearance & Deter. -- Region -- Multi -- CS & Discriminative -- Gradient-based & Active -- Distance-based -- Part-based \\
\rowcolor{taskClassif} Gu et al. \cite{gu2025protoasnet} & Classif. & Function \& Dyn. & Deter. -- Region -- Multi -- CS & Discriminative -- Gradient-based & Active -- Distance-based -- Part-based \\
\rowcolor{taskClassif} Gall{\'e}e et al. \cite{gallee2023interpretable} & Classif. & Appearance & Deter. -- Global -- Multi -- CS & Discriminative -- Gradient-based & Active -- Distance-based -- Case-based \\
\rowcolor{taskClassif} Nauta et al. \cite{nauta2023interpreting} & Classif. & Appearance & Deter. -- Region -- Multi -- SD & Contrastive -- Gradient-based & Active -- Distance-based -- Part-based \\
\rowcolor{taskClassif} Huy et al. \cite{huy2025interactive} & Classif. & Semantics & Deter. -- Region -- Multi -- CS & Contrastive -- Gradient-based & Active -- Distance-based -- Part-based \\
\rowcolor{taskClassif} Gautam et al. \cite{gautam2022demonstrating} & Classif. & Appearance & Deter. -- Region -- Multi -- CS & Discriminative -- Gradient-based & Active -- Distance-based -- Part-based \\
\rowcolor{taskClassif} Wang et al. \cite{wang2022knowledge} & Classif. & Appearance & Deter. -- Region -- Multi -- CS & Discriminative -- Gradient-based & Active -- Distance-based -- Part-based \\
\rowcolor{taskClassif} Wang et al. \cite{wang2024interpretable} & Classif. & Appearance & Deter. -- Region -- Multi -- CS & Discriminative -- Gradient-based & Active -- Distance-based -- Part-based \\
\rowcolor{taskClassif} Qu et al. \cite{qu2024rethinking} & Classif. & Class Identity & Deter. -- Region -- Single -- CS & Discriminative -- Momentum-based & Discarded -- Distance-based -- No expl. \\
\rowcolor{taskClassif} Rymarczyk et al. \cite{rymarczyk2022protomil} & Classif. & Appearance & Deter. -- Region -- Multi -- CS & Discriminative -- Gradient-based & Active -- Distance-based -- Part-based \\
\rowcolor{taskClassif} Yang et al. \cite{yang2023tpmil} & Classif. & Appearance & Deter. -- Region -- Single -- CS & Discriminative -- Gradient-based & Active -- Distance-based -- Part-based \\
\rowcolor{taskClassif} Gou et al. \cite{gou2025queryable} & Classif. & Semantics & Deter. -- Global -- Multi -- SD & Discriminative -- Gradient-based & Active -- Distance-based -- Implicit \\
\rowcolor{taskClassif} Zhu et al. \cite{zhu2024sfpl} & Classif. & Class Identity & Deter. -- Global -- Multi -- CS & Contrastive -- Momentum-based & Active -- Distance-based -- Implicit \\
\rowcolor{taskClassif} Wang et al. \cite{wang2025cross} & Classif. & Appearance & Deter. -- Region -- Multi -- CS & Discriminative -- Gradient-based & Active -- Distance-based -- Part-based \\
\rowcolor{taskClassif} Zhao et al. \cite{zhao2025medfedproto} & Classif. & Population & Deter. -- Global -- Single -- CS & Contrastive -- Momentum-based & Active -- Distance-based -- Implicit \\
\rowcolor{taskClassif} Zhang et al. \cite{zhang2024disease} & Classif. & Semantics & Deter. -- Global -- Single -- CS & Contrastive -- Gradient-based & Discarded -- Distance-based -- No expl. \\
\midrule

\rowcolor{taskSeg} Ouyang et al. \cite{ouyang2020self} & Seg. & Shape \& Anatomy & Deter. -- Region -- Multi -- CS & Discriminative -- On-the-fly & Active -- Distance-based -- Implicit \\
\rowcolor{taskSeg} Yu et al. \cite{yu2021location} & Seg. & Shape \& Anatomy & Deter. -- Region -- Multi -- CS & Discriminative -- On-the-fly & Active -- Distance-based -- Implicit \\
\rowcolor{taskSeg} Zhu et al. \cite{zhu2023partition} & Seg. & Shape \& Anatomy & Deter. -- Region -- Multi -- CS & Discriminative -- On-the-fly & Active -- Distance-based -- Implicit \\
\rowcolor{taskSeg} Zhu et al. \cite{zhu2023few} & Seg. & Shape \& Anatomy & Deter. -- Region -- Multi -- CS & Discriminative -- On-the-fly & Active -- Distance-based -- Implicit \\
\rowcolor{taskSeg} Zhao et al. \cite{zhao2024dual} & Seg. & Shape \& Anatomy & Deter. -- Region -- Multi -- CS & Discriminative -- On-the-fly & Active -- Distance-based -- Implicit \\
\rowcolor{taskSeg} Tang et al. \cite{tang2025few} & Seg. & Shape \& Anatomy & Deter. -- Region -- Multi -- CS & Discriminative -- On-the-fly & Active -- Distance-based -- Implicit \\
\rowcolor{taskSeg} Yang et al. \cite{yang2026proto} & Seg. & Shape \& Anatomy & Deter. -- Region -- Multi -- CS & Contrastive -- Momentum-based & Active -- Distance-based -- Implicit \\
\rowcolor{taskSeg} Wang et al. \cite{wang2025edge} & Seg. & Shape \& Anatomy & Deter. -- Multi-scale -- Multi -- CS & Discriminative -- On-the-fly & Active -- Distance-based -- Implicit \\
\rowcolor{taskSeg} Jiang and Zhang \cite{jiang2025concentrate} & Seg. & Shape \& Anatomy & Deter. -- Region -- Multi -- CS & Discriminative -- On-the-fly & Active -- Distance-based -- Implicit \\
\rowcolor{taskSeg} Liang et al. \cite{liang2024advancing} & Seg. & Class Identity & Deter. -- Multi-scale -- Single -- CS & Discriminative -- Gradient-based & Active -- Distance-based -- Implicit\\
\rowcolor{taskSeg} Zhang et al. \cite{zhang2024prototype} & Seg. & Shape \& Anatomy & Deter. -- Region -- Multi -- CS & Discriminative -- On-the-fly & Active -- Learnable metric -- Implicit \\
\rowcolor{taskSeg} Huang et al. \cite{huang2024prototype} & Seg. & Shape \& Anatomy & Deter. -- Region -- Multi -- CS & Discriminative -- On-the-fly & Active -- Learnable metric -- Implicit \\
\rowcolor{taskSeg} Li et al. \cite{li2023prototypical} & Seg. & Shape \& Anatomy & Deter. -- Region -- Multi -- CS & Discriminative -- On-the-fly & Active -- Distance-based -- Implicit \\
\rowcolor{taskSeg} Wang and Zheng \cite{wang2024pfmnet} & Seg. & Class Identity & Deter. -- Region -- Single -- CS & Discriminative -- Gradient-based & Active -- Learnable metric -- Implicit \\

\rowcolor{taskSeg} Yu et al. \cite{yu2023source} & Seg. & Population & Deter. -- Region -- Single -- CS & No obj. -- Frozen & Active -- Distance-based -- Implicit \\

\rowcolor{taskSeg} Wei et al. \cite{wei2024prompting} & Seg. & Population & Deter. -- Global -- Multi -- SD & Discriminative -- Gradient-based & Active -- Distance-based -- Implicit \\

\rowcolor{taskSeg} Xu et al. \cite{wu2022all} & Seg. & Class Identity & Deter. -- Region -- Single -- CS & Discriminative -- On-the-fly & Discarded -- Distance-based -- No expl. \\
\rowcolor{taskSeg} Tian et al. \cite{tian2023implicit} & Seg. & Class Identity & Deter. -- Region -- Multi -- CS & Discriminative -- On-the-fly & Active -- Distance-based -- Implicit \\
\rowcolor{taskSeg} Liu et al. \cite{liu2023prototype} & Seg. & Shape \& Anatomy & Deter. -- Region -- Multi -- CS & Contrastive -- On-the-fly & Discarded -- Distance-based -- No expl. \\
\rowcolor{taskSeg} Liu et al. \cite{liu2026kp2l} & Seg. & Class Identity & Deter. -- Multi-scale -- Single -- CS & Discriminative -- On-the-fly & Discarded -- Distance-based -- No expl. \\
\rowcolor{taskSeg} Yuan et al. \cite{yuan2024effective} & Seg. & Statistics & Prob. -- Region -- Single -- CS & Discriminative -- Momentum-based & Active -- Distance-based -- Implicit \\
\rowcolor{taskSeg} He et al. \cite{he2025evidential} & Seg. & Class Identity & Deter. -- Region -- Single -- CS & Discriminative -- On-the-fly & Discarded -- Distance-based -- No expl. \\
\rowcolor{taskSeg} Liu et al. \cite{liu2026embracing} & Seg. & Appearance & Deter. -- Region -- Multi -- CS & Contrastive -- On-the-fly & Discarded -- Distance-based -- No expl. \\

\rowcolor{taskSeg} Liu et al. \cite{liu2025procns} & Seg. & Class Identity & Deter. -- Multi-scale -- Single -- CS & Discriminative -- On-the-fly & Discarded -- Distance-based -- No expl. \\
\rowcolor{taskSeg} Pan et al. \cite{pan2023human} & Seg. & Appearance & Deter. -- Region -- Multi -- CS & Contrastive -- Clustering-based & Discarded -- Distance-based -- No expl. \\

\rowcolor{taskSeg} Ge et al. \cite{ge2026progis} & Seg. & Class Identity & Deter. -- Region -- Single -- CS & Contrastive -- On-the-fly & Active -- Distance-based -- Implicit \\

\rowcolor{taskSeg} Zhu et al. \cite{zhu2025prime} & Seg. & Class Identity & Deter. -- Region -- Single -- CS & Contrastive -- Momentum-based & Active -- Distance-based -- Implicit \\
\midrule

\rowcolor{taskSurvival} Liu et al. \cite{liu2025prototype} & Survival Pred. & Semantics & Deter. -- Global -- Multi -- CS & Contrastive -- Gradient-based & Active -- Learnable metric -- Implicit \\
\rowcolor{taskSurvival} Raza et al. \cite{raza2025ps3} & Survival Pred. & Statistics & Prob. -- Region -- Multi -- SD & Generative -- On-the-fly & Active -- Learnable metric -- Implicit \\
\rowcolor{taskSurvival} Liu et al. \cite{liu2026together} & Survival Pred. & Semantics & Deter. -- Region -- Multi -- SD & Contrastive -- Gradient-based & Active -- Distance-based -- Implicit \\

\rowcolor{taskSurvival} Yu et al. \cite{yu2026prime} & Survival Pred. & Semantics & Deter. -- Global -- Multi -- SD & Contrastive -- Gradient-based & Active -- Distance-based -- Implicit \\

\rowcolor{taskSurvival} Wang et al. \cite{wang2026protosurv} & Survival Pred. & Appearance & Deter. -- Region -- Multi -- SD & Discriminative -- On-the-fly & Active -- Learnable metric -- Part-based \\

\rowcolor{taskSurvival} Jiang et al. \cite{jiang2026multimodal} & Survival Pred. & Semantics & Deter. -- Global -- Multi -- CS & Discriminative -- Momentum-based & Active -- Distance-based -- Implicit \\

\rowcolor{taskSurvival} Yang et al. \cite{yang2026protomm} & Survival Pred. & Semantics & Deter. -- Global -- Multi -- SD & Discriminative -- Gradient-based & Active -- Distance-based -- Case-based \\

\rowcolor{taskSurvival} Kumari et al. \cite{kumari2026protosurvx} & Survival Pred. & Semantics & Deter. -- Global -- Multi -- SD & Discriminative -- Gradient-based & Active -- Distance-based -- Implicit \\

\rowcolor{taskSurvival} Liu et al. \cite{liu2026protosurv} & Prognosis Pred. & Class Identity & Deter. -- Global -- Single -- CS & No obj. -- Frozen & Discarded -- Distance-based -- No expl. \\

\rowcolor{taskSurvival} Wang et al. \cite{wang2026impro} & Prognosis Pred. & Class Identity & Deter. -- Global -- Multi -- SD & Discriminative -- Gradient-based & Active -- Distance-based -- Implicit \\

\rowcolor{taskSurvival} Yu et al. \cite{yu2026learning} & Recurrence Pred. & Class Identity & Deter. -- Global -- Single -- CS & Contrastive -- Momentum-based & Discarded -- Distance-based -- No expl. \\

\midrule

\rowcolor{taskAnomaly} Xiang et al. \cite{xiang2024exploiting} & Anomaly Det. & Shape \& Anatomy & Deter. -- Multi-scale -- Multi -- SD & Generative -- Gradient-based & Active -- Distance-based -- Implicit \\
\rowcolor{taskAnomaly} Li et al. \cite{li2025diverse} & Anomaly Det. & Appearance & Deter. -- Region -- Multi -- SD & Generative -- On-the-fly & Active -- Learnable metric -- Implicit \\
\rowcolor{taskAnomaly} Wu et al. \cite{wu2025anyad} & Anomaly Det. & Appearance & Deter. -- Region -- Multi -- SD & Generative -- On-the-fly & Active -- Learnable metric -- Implicit \\
\rowcolor{taskAnomaly} Wan et al. \cite{wan2026beyond} & Anomaly Det. & Class Identity & Deter. -- Multi-scale -- Single -- CS & Discriminative -- Gradient-based & Active -- Distance-based -- Implicit \\
\rowcolor{taskAnomaly} Kobayashi and Yamaguchi \cite{kobayashi2024simple} & Anomaly Det. & Appearance & Deter. -- Region -- Multi -- CS & Discriminative -- Gradient-based & Active -- Distance-based -- Part-based \\
\rowcolor{taskAnomaly} Huang et al. \cite{huang2024uncertainty} & Anomaly Det. & Appearance & Deter. -- Region -- Multi -- SD & Contrastive -- Clustering-based & Active -- Learnable metric -- Implicit \\
\rowcolor{taskAnomaly} Lee et al. \cite{lee2025fewshot} & Anomaly Det. & Class Identity & Deter. -- Global -- Single -- CS & Discriminative -- On-the-fly & Active -- Distance-based -- Implicit \\

\rowcolor{taskAnomaly} Pan and Lu \cite{pan2025spmamba} & Anomaly Det. & Shape \& Anatomy & Deter. -- Region -- Multi -- SD & Generative -- Gradient-based & Active -- Distance-based -- Implicit \\

\rowcolor{taskAnomaly} She et al. \cite{she2024memory} & Anomaly Det. & Appearance & Deter. -- Region -- Multi -- CS & Contrastive -- Gradient-based & Active -- Distance-based -- Implicit \\

\midrule

\rowcolor{taskReportGen} Chen et al. \cite{chen2021cross} & Report Gen. & Semantics & Deter. -- Region -- Multi -- SD & Generative -- Gradient-based & Active -- Distance-based -- Implicit \\
\rowcolor{taskReportGen} Wang et al. \cite{wang2022cross} & Report Gen. & Semantics & Deter. -- Region -- Multi -- SD & Generative -- Gradient-based & Active -- Distance-based -- Implicit \\
\rowcolor{taskReportGen} Yan et al. \cite{yan2024buffalo} & Report Gen. & Population & Prob. -- Global -- Multi -- SD & Contrastive -- Clustering-based & Discarded -- Distance-based -- No expl. \\
\rowcolor{taskReportGen} Pellegrini et al. \cite{pellegrini2026prototype} & Structured Report Gen. & Semantics & Deter. -- Global -- Single -- CS & Discriminative -- Momentum-based & Active -- Distance-based -- Implicit \\

\rowcolor{taskReportGen} Yang et al. \cite{yang2026clinical} & Report Gen. & Semantics & Deter. -- Global -- Multi -- CS & Discriminative -- Gradient-based & Active -- Distance-based -- Implicit \\

\rowcolor{taskReportGen} Chen et al. \cite{chen2025dia} & Report Gen. & Semantics & Deter. -- Global -- Multi -- CS & Contrastive -- Gradient-based & Active -- Distance-based -- Implicit \\
\rowcolor{taskReportGen} Zhang et al. \cite{zhang2024attribute} & Report Gen. & Semantics & Deter. -- Region -- Single -- CS & Contrastive -- Gradient-based & Active -- Distance-based -- Implicit \\

\midrule

\rowcolor{taskEmerging} Song et al. \cite{song2024morphological} & Rep. Learning & Appearance & Deter. -- Region -- Multi -- SD & No obj. -- Frozen & Active -- Distance-based -- Part-based \\

\rowcolor{taskEmerging} Hesse et al. \cite{hesse2024prototype} & Age Pred. & Appearance & Deter. -- Global -- Multi -- CS & Discriminative -- Gradient-based & Active -- Distance-based -- Case-based \\

\rowcolor{taskEmerging} Lisazo et al. \cite{lisazo2026learning} & Age Pred. & Class Identity & Deter. -- Global -- Single -- CS & Generative -- Gradient-based & Active -- Distance-based -- Case-based \\

\rowcolor{taskEmerging} Bazargani et al. \cite{bazargani2025progrespine} & Age Pred. & Class Identity & Deter. -- Global -- Multi -- CS & Discriminative -- Gradient-based & Active -- Distance-based -- Case-based \\

\rowcolor{taskEmerging} Ghamary et al. \cite{ghamary2026protoefnet} & Ejection Fraction Pred. & Function \& Dyn. & Deter. -- Region -- Multi -- CS & Discriminative -- Gradient-based & Active -- Distance-based -- Part-based \\
\rowcolor{taskEmerging} Nicolson et al. \cite{nicolson2025human} & Gest. Age Pred. & Appearance & Deter. -- Region -- Multi -- SD & Discriminative -- Gradient-based & Active -- Distance-based -- Part-based \\

\rowcolor{taskEmerging} Cai et al. \cite{cai2026adaptive} & Open-set Recog. & Class Identity & Deter. -- Global -- Multi -- CS & Discriminative -- Gradient-based & Active -- Distance-based -- Implicit \\

\rowcolor{taskEmerging} Liu et al. \cite{liu2026openness} & Open-set Recog. & Class Identity & Deter. -- Global -- Multi -- CS & Discriminative -- Gradient-based & Active -- Distance-based -- Implicit \\

\rowcolor{taskEmerging} Sun et al. \cite{sun2026sculpt} & OOD Detection & Appearance & Deter. -- Region -- Multi -- CS & Contrastive -- Gradient-based & Active -- Learnable metric -- Implicit \\

\rowcolor{taskEmerging} Shandilya et al. \cite{shandilya2024improve} & Image Synthesis & Class Identity & Deter. -- Global -- Single -- CS & No obj. -- Frozen & Active -- Learnable metric -- Implicit \\

\rowcolor{taskEmerging} Chang et al. \cite{chang2025enhancing} & Image Synthesis & Class Identity & Deter. -- Global -- Single -- CS & Generative -- Momentum-based & Active -- Distance-based -- Implicit \\

\rowcolor{taskEmerging} Redekop et al. \cite{redekop2025prototype} & Image Synthesis & Appearance & Deter. -- Region -- Multi -- CS & No obj. -- Frozen & Active -- Learnable metric -- Implicit \\

\rowcolor{taskEmerging} Bhat et al. \cite{bhat2025exemplar} & Lesion Detection & Appearance & Deter. -- Region -- Single -- CS & Discriminative -- Momentum-based & Active -- Learnable metric -- Implicit \\
\rowcolor{taskEmerging} Xu et al. \cite{xu2026rethinking} & Landmark Loc. & Shape \& Anatomy & Deter. -- Region -- Single -- CS & Discriminative -- Gradient-based & Active -- Distance-based -- Implicit \\
\rowcolor{taskEmerging} Wang et al. \cite{wang2022multi} & Rep. Learning & Semantics & Deter. -- Global -- Multi -- SD & Contrastive -- Gradient-based & Discarded -- Distance-based -- No expl. \\
\rowcolor{taskEmerging} Cheng et al. \cite{cheng2023prior} & Rep. Learning & Semantics & Deter. -- Region -- Multi -- SD & Generative -- Gradient-based & Discarded -- Distance-based -- No expl. \\
\rowcolor{taskEmerging} Gowda et al. \cite{gowda2025prototype} & Image-Report Retr. & Semantics & Deter. -- Multi-scale -- Multi -- CS & Contrastive -- On-the-fly & Active -- Distance-based -- Implicit \\

\rowcolor{taskEmerging} Santos et al. \cite{santos2025toward} & Image Retrieval & Class Identity & Deter. -- Global -- Single -- CS & Discriminative -- On-the-fly & Discarded -- Distance-based -- No expl. \\

\rowcolor{taskEmerging} Yao et al. \cite{yao2024unsupervised} & Image Retrieval & Class Identity & Deter. -- Global -- Multi -- SD & Contrastive -- Clustering-based & Discarded -- Distance-based -- No expl. \\

\rowcolor{taskEmerging} Xu et al. \cite{xu2025medical} & Registration & Shape \& Anatomy & Deter. -- Region -- Single -- CS & Contrastive -- On-the-fly & Discarded -- Distance-based -- No expl. \\

\rowcolor{taskEmerging} Wang et al. \cite{wang2025fedpmr} & MR Reconstruction & Population & Deter. -- Global -- Single -- CS & Discriminative -- On-the-fly & Discarded -- Distance-based -- No expl. \\

\rowcolor{taskEmerging} Kong et al. \cite{kong2026concept} & Rep. Learning & Semantics & Deter. -- Multi-scale -- Multi -- SD & Contrastive -- Gradient-based & Discarded -- Distance-based -- No expl. \\

\rowcolor{taskEmerging} Chai et al. \cite{chai2022deep} & Cell Detection & Class Identity & Deter. -- Global -- Single -- CS & Contrastive -- Gradient-based & Discarded -- Distance-based -- No expl. \\

\rowcolor{taskEmerging} Dai et al. \cite{dai2026learning} & Landmark Loc. & Shape \& Anatomy & Deter. -- Region -- Single -- CS & Discriminative -- Gradient-based & Active -- Learnable metric -- Implicit \\

\rowcolor{taskEmerging} Wang et al. \cite{wang2024unsupervised} & Registration & Class Identity & Deter. -- Global -- Single -- CS & Generative -- Momentum-based & Active -- Learnable metric -- Implicit \\

\end{xltabular}
\endgroup
\twocolumn

\end{document}